\documentclass[10pt,twocolumn,letterpaper]{article}

\usepackage{cvpr}
\makeatletter
\@namedef{ver@everyshi.sty}{}
\makeatother
\usepackage{tikz}
\usepackage{adjustbox}
\usepackage{array}
\usepackage{pgfplots}
\usepackage{times}
\usepackage{epsfig}
\usepackage{graphicx}
\usepackage{amsmath}
\usepackage{amssymb}
\usepackage{booktabs}
\usepackage{multirow}
\usepackage{color}
\usepackage{pifont}
\usepackage{comment}
\usepackage{etoolbox}
\usepackage{overpic}
\usepackage{tabularx}
\usepackage{subcaption}

\usetikzlibrary{decorations.pathreplacing,calc,automata, shapes.geometric,shapes.arrows,decorations.pathmorphing,matrix,chains,scopes,positioning,arrows,fit}

\newcommand{\tikzmark}[2][-3pt]{\tikz[remember picture, overlay, baseline=-0.5ex]\node[#1](#2){};}

\newcounter{arrow}
\newcommand{\drawcurvedarrow}[3][]{%
 \refstepcounter{arrow}
 \tikz[remember picture, overlay]\draw (#2.center)edge[#1]node[coordinate,pos=0.5, name=arrow-\thearrow]{}(#3.center);
}

\usepackage{wrapfig}
\usepackage[theorems]{tcolorbox}

\usepackage{caption}
\usepackage[pagebackref=true,breaklinks=true,letterpaper=true,colorlinks,bookmarks=false]{hyperref}

\cvprfinalcopy 

\usepackage{balance}

\newcommand{\refsec}[1]{Section~\ref{sec:#1}}

\newcommand{\refeq}[1]{Equation~\ref{eq:#1}}
\newcommand{\reffig}[1]{Figure~\ref{fig:#1}}
\newcommand{\reftab}[1]{Table~\ref{tab:#1}}

\def\onedot{.\xspace}

\def\eg{\emph{e.g}\onedot} 
\def\ie{\emph{i.e}\onedot} 
\def\cf{\emph{c.f}\onedot}

\def\etal{\emph{et al}\onedot}

\newcommand{\euc}{Euclidean}
\newcommand{\geo}{geodesic}

\newcommand*\numcircledtikz[1]{\tikz[baseline=(char.base)]{
            \node[shape=circle,draw,inner sep=0.5pt] (char) {#1};}}

\newcommand{\namelong}{DualConvMesh-Nets}
\newcommand{\nameshort}{DCM-Net}

\newcommand{\dualarch}{DCM-Net}
\newcommand{\singlearch}{SCM-Net}
\newcommand{\singlearchshort}{Single}
\newcommand{\dualarchshort}{Dual}

\renewcommand{\vec}[1]{\ensuremath{\mathbf{#1}}}

\newcommand{\ColorMapCircle}{\ding{108}}

\newcommand{\PAR}[1]{\vskip4pt \noindent {\bf #1~}}
\newcommand{\PARbegin}[1]{\noindent {\bf #1~}}

\newcommand{\parag}[1]{\PAR{#1}}

\tcbset{
  defstyle/.style={
    fonttitle=\sffamily\bfseries,
    arc=0mm,
    colback=white,
    colframe=gray,
    left=0mm,
    separator sign none
  }
}
\newtcbtheorem[number within=subsection]{Def}{Main contributions of this work}{defstyle}{def}

\def\cvprPaperID{????} 

\ifcvprfinal\fi
\begin{document}

\title{DualConvMesh-Net:\\
Joint Geodesic and Euclidean Convolutions on 3D Meshes}

\newcommand{\footremember}[2]{%
   \thanks{#2}
    \newcounter{#1}
    \setcounter{#1}{\value{footnote}}%
}
\newcommand{\footrecall}[1]{%
    \footnotemark[\value{#1}]%
}

\author{
  \hspace{-1.3cm}
  \begin{tabular}[t]{c}
    Jonas Schult\footremember{cont}{Equal contribution.},~~Francis Engelmann\footrecall{cont}~,~~Theodora Kontogianni,~~Bastian Leibe\\
    RWTH Aachen University\\
    {\tt\small \{schult, engelmann, kontogianni, leibe\}@vision.rwth-aachen.de}
\end{tabular}
}

\maketitle
\thispagestyle{empty}

\begin{abstract}
\vspace{-4pt}

We propose \namelong{} (\nameshort{}) a family of deep hierarchical convolutional networks over 3D geometric data that \emph{combines two types} of convolutions.
The first type, \emph{\geo{} convolutions}, defines the kernel weights over mesh surfaces or graphs.
That is, the convolutional kernel weights are mapped to the local surface of a given mesh.
The second type, \emph{\euc{} convolutions}, is independent of any underlying mesh structure.
The convolutional kernel is applied on a neighborhood obtained from a local affinity representation based on the Euclidean distance between 3D points. 
Intuitively, geodesic convolutions can easily separate objects that are spatially close but have disconnected surfaces, while Euclidean convolutions can represent interactions between nearby objects better, as they are oblivious to object surfaces.
To realize a multi-resolution architecture, we borrow well-established mesh simplification methods from the geometry processing domain and adapt them to define mesh-preserving pooling and unpooling operations. 
We experimentally show that combining both types of convolutions in our architecture leads to significant performance gains for 3D semantic segmentation, and we report competitive results on three scene segmentation benchmarks.
Our models and code are publicly available\footnote{
\href{https://github.com/VisualComputingInstitute/dcm-net}{github.com/VisualComputingInstitute/dcm-net/}}.

\end{abstract}
\definecolor{center}{rgb}{0, 1, 0.141176471}

\vspace{-10pt}
\section{Introduction}
\vspace{-4pt}

\begin{figure}[t]
  \centering
  \begin{tikzpicture}[-,>=stealth',shorten >=0pt,auto,node distance=1.6cm,semithick, transform shape, scale=1.0]
\tikzstyle{every state}=[           rectangle,
           rounded corners,
           draw=black, thick,
           minimum height=1em,
           inner sep=2mm, align=center,
           text centered]

   \node (figure1) {\includegraphics[width=1.0\linewidth, trim={0 0 0 0cm},clip]{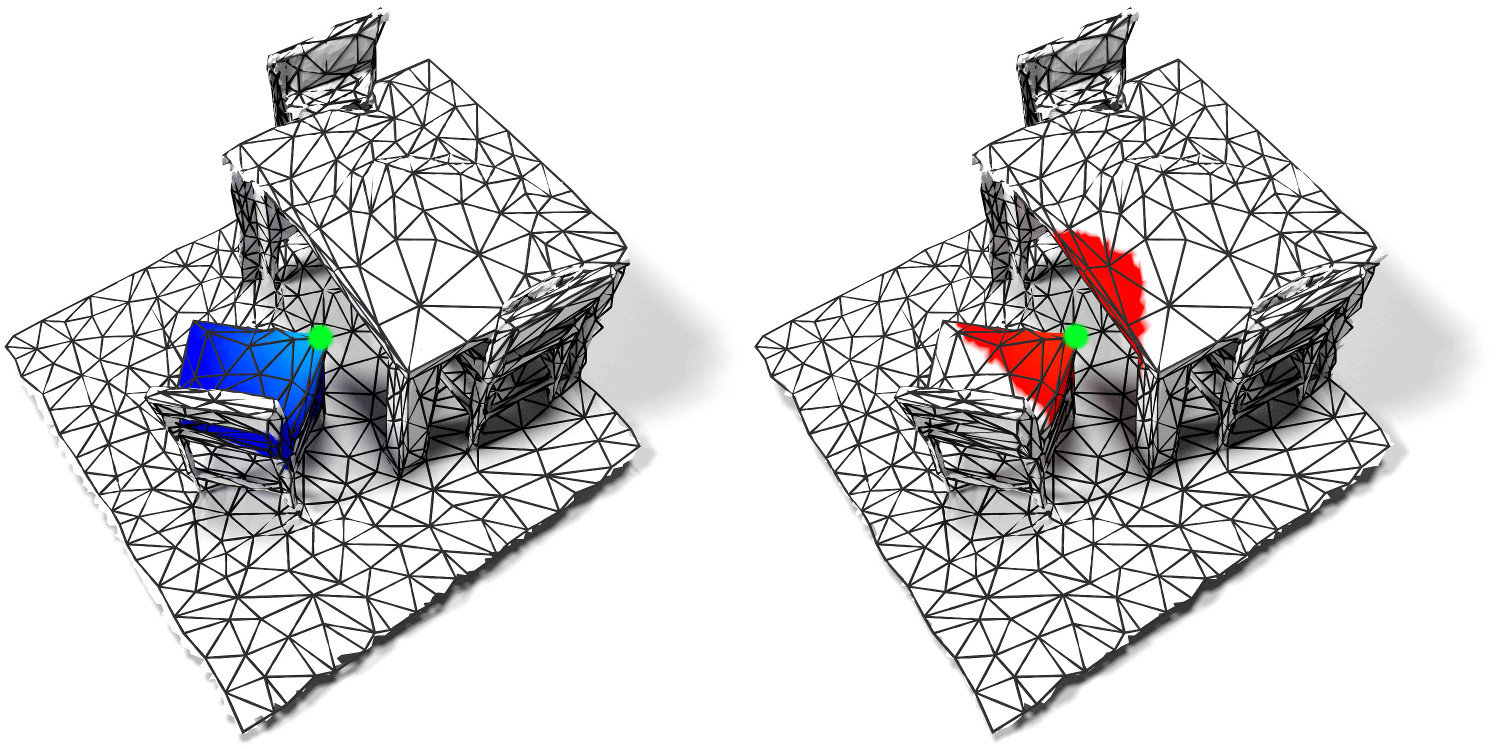}};

   \node        (geo) at (-2.5,2.3)                        {\emph{Geodesic Neighborhood}};
   
  \node        (euc) at (1.8,2.3)                        {\emph{Euclidean Neighborhood}};
\end{tikzpicture}
  \vspace{-20pt}
  \caption{\small \textbf{Comparison of geodesic (left) and Euclidean neighborhoods (right).}
  Our \nameshort{} combines geodesic and Euclidean convolutions.
Geodesic convolutions follow the surface of individual objects, which can be beneficial to learn specific object shapes.
Euclidean convolutions can bridge over small gaps, which can encourage the flow of relevant context information between spatially nearby but geodesically distant objects, while being able to connect disconnected parts due to scanning artifacts.
The color gradient shows the geodesic and Euclidean distances between the \textcolor{center}{\ColorMapCircle}~center point and its neighbors.
The scan section is taken from the ScanNet dataset\,\cite{Dai17CVPR}.}
\label{fig:motivation}
\end{figure}

\emph{Geometric deep learning}\,\cite{Boscaini2016NIPS, Bronstein2017SPM, Defferrard16NIPS, Henaff2015CoRR, Kipf17ICLR, Monti2017CVPR} aims at transferring the successes of CNNs from regular, discrete domains, \eg, 1D audio, 2D images or 3D voxel grids, onto irregular data representations such as graphs, point clouds or 3D meshes.
Currently, geometric deep learning is divided into two main areas relying on different data representations: \emph{3D scene understanding} and \emph{3D shape analysis}.

The former looks at tasks such as semantic segmentation\,\cite{Dai18ECCV, Engelmann18ECCVW, Graham18CVPR, Landrieu17CVPR, Qi17CVPR, Qi17ICCV, Wu18CVPR}, instance segmentation\,\cite{elich2019_bevis, Engelmann20CVPR, Narita15CoRR, Hou2018CoRR, Wang2018CVPR, Wang2019CoRR} and part segmentation\,\cite{Graham18CVPR, Qi17CVPR, Su18CVPR, Wu18CVPR}.
Here, the focus lies primarily on processing point cloud data.
One choice is to project raw point clouds into a discrete 3D grid representation, which enables standard 3D CNNs to be applied, \ie, by sliding kernels over neighboring voxels\,\cite{Dai18CVPR, Maturana2015IROS, Tchapmi173DV, Song15CVPR, Roynard2018CoRR}.
Alternative approaches operate directly on raw point clouds\,\cite{Matan2018Graph, Hua2019CVPR, Li18NIPS,Qi17CVPR,Qi17NIPS, Su18CVPR, XuSpider18Arxiv}.
In this case, the challenge consists in defining convolutional operators over point sets.
Commonly, the convolutional kernels are applied to local point neighborhoods obtained from spherical or $k$-nn neighborhoods defined over the Euclidean distance between pairs of points.
We refer to these convolutions as \emph{Euclidean convolutions} (see \reffig{motivation}, right).
Consequently, regardless of point cloud or voxel representations, these convolutions are agnostic to the surface information and, therefore, sensitive to surface deformations.

Unlike 3D scene understanding, \emph{3D shape analysis} is concerned with tasks such as shape correspondence\,\cite{Boscaini2016NIPS}, shape descriptors\,\cite{xie15cvpr}, and shape retrieval\,\cite{Monti2017CVPR}.
As opposed to the methods mentioned earlier, shape analysis focuses on the surface information encoded in meshes or graphs.
Here, convolutional kernels are defined over local patches or neighborhoods on the surface of a mesh or graph.
These neighborhoods are localized by the \emph{geodesic} distance between nodes on the surface mesh, \ie, points that are reachable by one edge connection along the surface mesh.
We therefore refer to them as \emph{geodesic convolutions} (see\,\reffig{motivation}, left).
A notable property of geodesic convolutions is their invariance to surface deformations, which is generally desired in tasks such as shape correspondence.

In this work, we investigate the role of \emph{geodesic} and \emph{Euclidean} convolutions in the task on 3D semantic segmentation of 3D meshes.
So far, few approaches have made use of explicit surface information and geodesic convolutions for semantic scene segmentation\,\cite{Pan2018CoRR, Huang2018CoRR, Tatarchenko18CVPR}, whereas Euclidean convolutions are very popular in the field, \eg, \cite{Matan2018Graph, Choy2019CVPR, Hua2019CVPR, thomas2019ICCV, XuSpider18Arxiv}.
As visualized in \reffig{motivation}, both approaches have their own characteristics.
While geodesic convolutions follow the surface to learn specific object shapes, Euclidean convolutions encourage the feature propagation over geodesically remote areas to accumulate contextual information.
It is therefore natural to ask how these advantages can be combined in a common architecture.
This is the question we address in this work.

We propose a novel deep hierarchical architecture, \emph{\namelong{}}, that starts from a mesh representation and combines both types of convolutions. 
In order to design such a hierarchical architecture that is capable of learning useful Euclidean and geodesic features at different scales, it is critical to define a mesh pooling algorithm which maintains a meaningful mesh structure throughout all mesh levels. We therefore adapt \emph{vertex clustering (VC)}\,\cite{Rossignac1993Multiresolution3A} and \emph{Quadric Error Metrics (QEM)}\,\cite{Garland1997CGIT}, two well-established mesh simplification approaches from the geometry processing domain, in order to define meaningful pooling and unpooling operations on meshes. We introduce \emph{Pooling Trace Maps} as an efficient way to keep track of vertex connectivity for pooling and unpooling. As a practical way of reducing the dependency to local vertex densities, we propose \emph{Random Edge Sampling (RES)} for radius neighborhoods\,\cite{Hermosilla2018Graph}.

Our proposed \nameshort{} architecture achieves competitive results on the popular ScanNet\,v2 benchmark\,\cite{Dai17CVPR}, as well as on Stanford 3D Indoor Scenes dataset\,\cite{Armeni16CVPR}. For graph convolutional approaches, we define a new state-of-the-art on both datasets.
Furthermore, we achieve state-of-the-art performance on the recent Matterport3D\,\cite{Matterport3D} benchmark.

In summary, the main contributions of this paper are: \numcircledtikz{1} We propose a novel family of deep convolutional networks, \emph{\nameshort{}s}, that operate in both the Euclidean and geodesic space.
\numcircledtikz{2} We adapt two theoretically well-founded mesh simplification algorithms as means of pooling and unpooling in order to create multi-scale architectures on meshes, and we experimentally compare their performance.
\numcircledtikz{3}~We~introduce a novel sampling method on graph neighborhoods, \emph{Random Edge Sampling}, which allows us to train networks with smaller sample sizes while evaluating them with better approximations.
\numcircledtikz{4} We present a thorough ablation study, which empirically proves that combining Euclidean and geodesic convolutions provides a consistent benefit using radius neighborhoods, regardless of the pooling method used in the architecture.
\vspace{-4pt}
\section{Related work}
\vspace{-4pt}

\begin{figure*}[t]
  \centering
    \begin{tikzpicture}[-,>=stealth',shorten >=0pt,auto,node distance=1.6cm,semithick, transform shape, scale=1.0]
\tikzstyle{every state}=[           rectangle,
           rounded corners,
           draw=black, thick,
           minimum height=1em,
           inner sep=2mm, align=center,
           text centered]

   \node (figure2) {    \includegraphics[width=1\linewidth, trim={0 0  0 0}]{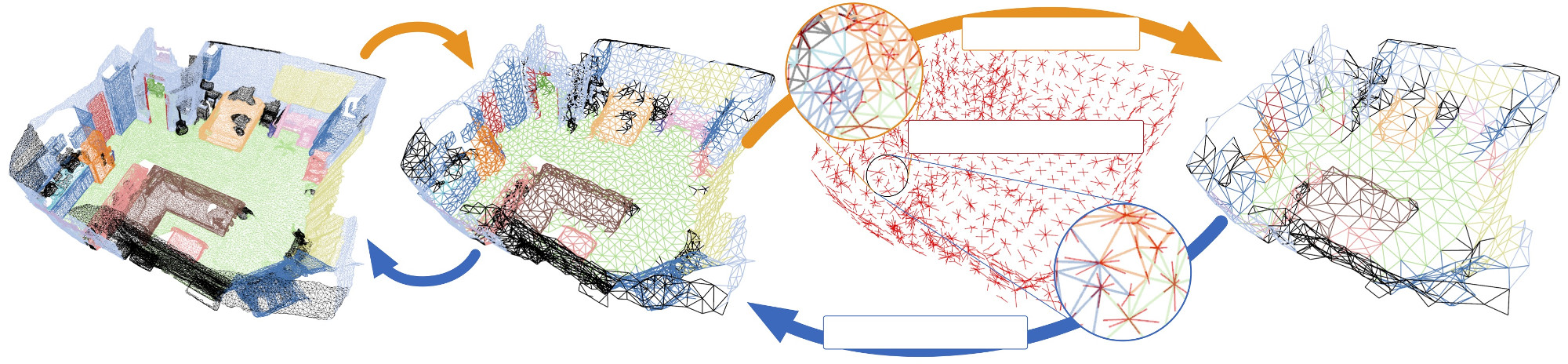}};

  \node        (pool) at (3.0,1.61)                        {\small \emph{Mesh Pooling}};
  \node        (unpool) at (1.58,-1.67)                        {\small \emph{Mesh Unpooling}};
  \node        (trace) at (2.65,0.5)                        {\small \emph{Pooling Trace Map}};
  
    \node        (mesh0) at (-6.5,-1.75)                        {\small \emph{Mesh Full Resolution $\mathcal{M}^0$}};
    
        \node        (mesh1) at (-2.2,-1.75)                        {\small \emph{Pooled Mesh $\mathcal{M}^1$}};
        
    \node        (mesh2) at (6.7,-1.75)                        {\small \emph{Pooled Mesh $\mathcal{M}^2$}};
\end{tikzpicture}
    \vspace{-22pt}
  \caption{
  \small \textbf{Pooling on meshes.} To perform geodesic convolutions on multiresolution representations, the geodesic mesh neighborhood needs to be preserved throughout the pooling operations.
  We leverage vertex clustering and Quadric Error Metrics which both preserve meaningful geodesic neighborhoods in all mesh levels.
  These pooling operations rely on a \emph{pooling trace map} (shown in red) that keeps track of vertex connectivity and is used for (un)pooling between adjacent mesh levels $\mathcal{M}^\ell$ with simple look-up operations.
  }
  \label{fig:pooling}
\end{figure*}
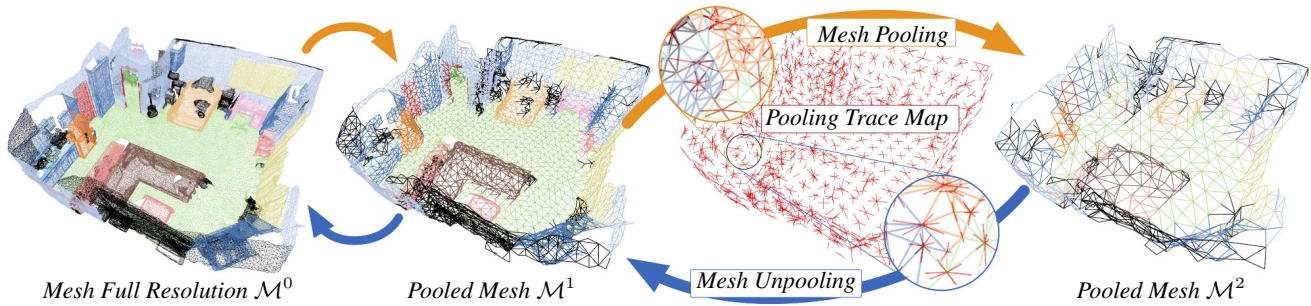

\PARbegin{Convolutions on point clouds.}
A simple way of handling point clouds is to transform them
into a voxel grid representation that enables standard CNNs to be applied \cite{Dai17CVPR, Dai18CVPR, Maturana2015IROS, Song15CVPR, Roynard2018CoRR}.
By construction, such approaches are limited to applying convolutional kernels on voxel neighborhoods, as
it is not trivial to define geodesic neighborhoods on regular grids.
Even recent methods focusing on efficient sparse voxel convolutions~\cite{Choy2019CVPR, Graham18CVPR} have similar limitations.
Numerous other approaches operate directly on raw point clouds using convolutional kernels that are applied to the local neighborhoods of points obtained using $k$-nn or spherical neighborhoods\,\cite{Matan2018Graph, LiCVPR18, Li18NIPS, Qi17NIPS, Wang18CVPRa}.
Alternative methods define the position of the kernel weights explicitly in the Euclidean space relative to point positions\,\cite{Matan2018Graph, Hua2019CVPR, thomas2019ICCV, XuSpider18Arxiv}. In both cases, the convolutional kernels are defined over the Euclidean space and are independent of the actual underlying object surface. In contrast, we additionally consider surface information using \geo{} convolutions in combination with the  \euc{} convolutions.

\PAR{Convolutions on meshes and graphs.}
Spectral filtering methods build on eigenvalue decomposition of the graph Laplacian\,\cite{Defferrard16NIPS, Henaff2015CoRR, Kipf17ICLR, Simonovsky17CVPR}.
While they work well on clean synthetic data, they are sensitive to reconstruction noise and do not generalize well across different graph structures.
Local filtering methods, such as geodesic CNN\,\cite{Masci2015ICCVWorkshops}, anisotropic CNN\,\cite{Boscaini2016NIPS} or the work of Monti \etal\cite{Monti2017CVPR} rely on handcrafted local coordinate systems defined over local patches on mesh surfaces. Verma \etal\cite{Verma2018CVPR} replace these hand-designed pseudo-coordinates with a learned mapping between filter weights and graph patches.
TextureNet\,\cite{Huang2018CoRR} applies traditional CNNs to high resolution textures originating from geodesic mesh surfaces.
TangentConvolutions\,\cite{Tatarchenko18CVPR} implicitly use surface information from estimated point normals by projecting point features on a local tangent plane and apply 2D CNNs.
Whereas all previously mentioned methods perform convolutions on vertices,
MeshCNN\,\cite{Hanocka2018arXiv} defines them over the edges of a mesh.

In summary, these methods use surface information from a mesh or graph to run geodesic convolutions.
Similarly, we consider \geo{} convolutions as graph convolutions defined over the mesh and take special provisions to enable pooling operations such that all simplifications of the original mesh still contain meaningful geodesic information.

\PAR{Pooling operations on point clouds and meshes.}
Hierarchical networks operate on multiple resolution levels of a 3D model (see~\reffig{pooling}), resulting in an increased receptive field of convolutions and robustness to small transformations.
To obtain fine-to-coarse representations, different pooling operations exist.
An important property of a pooling operation is whether it preserves the geometric and geodesic affinity information.
On point clouds, random sampling of points or Farthest Point Sampling (FPS) are popular and effective approaches\,\cite{Li18NIPS, Qi17NIPS, Wu18CVPR}.
They work well on point clouds; however, when applied to mesh vertices the interconnectivity of vertices is lost.
Hanocka \etal\cite{Hanocka2018arXiv} perform mesh pooling by \emph{learning} which edges to collapse.
Unlike previous works, Tatarchenko \etal\cite{Tatarchenko18CVPR} propose to pool on a regular 3D grid.
On meshes, Defferrard \etal\cite{Defferrard16NIPS} and Verma \etal\cite{Verma2018CVPR} use the Graclus algorithm\,\cite{Dhillon07PAMI}, while Ranjan \etal\cite{Ranjan2018ECCV} and Pan \etal\cite{Pan2018CoRR} rely on the mesh simplifying Quadric Error Metrics\,\cite{Garland1997CGIT}.
These mesh simplification approaches aim at reducing the number of vertices while introducing minimal geometric distortion by collapsing vertex pairs along the way.
However, this can lead to high-frequency signals in noisy areas.

In this work, we leverage two theoretically well-founded methods from the geometry processing domain: \emph{vertex clustering (VC)}\,\cite{Rossignac1993Multiresolution3A} and \emph{Quadric Error Metrics (QEM)}\,\cite{Garland1997CGIT}.
In order to allow multiresolution processing, we introduce \emph{Pooling Trace Maps} (see~\reffig{pooling}) to ensure well-defined pooling and unpooling operations on meshes.

\PAR{Sampling neighborhoods.}
Hermosilla \etal\cite{Hermosilla2018Graph} and Thomas \etal\cite{thomas2019ICCV} argue that $k$-nn graph approaches suffer from non-uniform point densities in point clouds.
Thus, they propose to use radius graphs to define the notion of neighborhoods for vertices in the Euclidean space.
However, very densely populated regions can lead to arbitrarily large neighborhoods, which introduces a computational burden for the algorithm.
Sampling the neighborhood space becomes inevitable.
Therefore, Hermosilla \etal\cite{Hermosilla2018Graph} use Poisson Disk Sampling, which preserves the relative density distribution of the point cloud but restricts the maximal density per cubic unit by the radius of the non-overlapping poisson disks. Then, the Kepler Conjecture\,\cite{hales-kepler} gives an upper bound for the neighborhood size.
Thomas \etal\cite{thomas2019ICCV} control the density of the point cloud by low-pass filtering it via grid subsampling.
In concurrent work, Lei \etal\cite{lei2019spherical} randomly sub-sample the neighborhood to obtain at most $K$ samples for approximating the neighborhood set.
In this work, we propose \emph{Random Edge Sampling} which is similar in spirit to\,\cite{lei2019spherical} but has a special appeal in its probabilistic interpretation of reducing the neighborhood size.
\begin{figure*}[t]
  \centering
  \begin{overpic}[unit=1mm,height=0.31\linewidth]{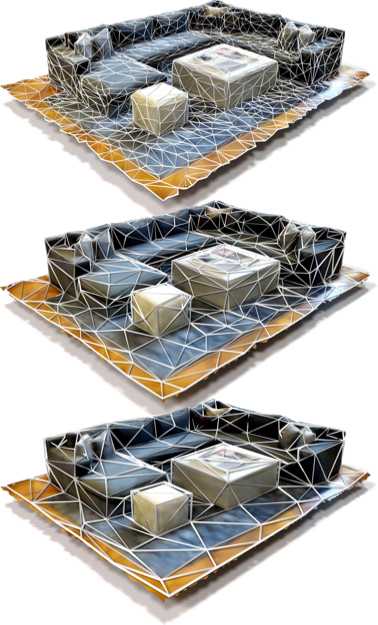}
\put(2,77){\rotatebox{-35}{\scriptsize \emph{Mesh} $\mathcal{M}^0$}}
\put(2,45){\rotatebox{-35}{\scriptsize \emph{Mesh} $\mathcal{M}^1$}}
\put(2,13){\rotatebox{-35}{\scriptsize \emph{Mesh} $\mathcal{M}^2$}}
\end{overpic}
  \includegraphics[height=0.31\linewidth, trim={0 0 0 0}]{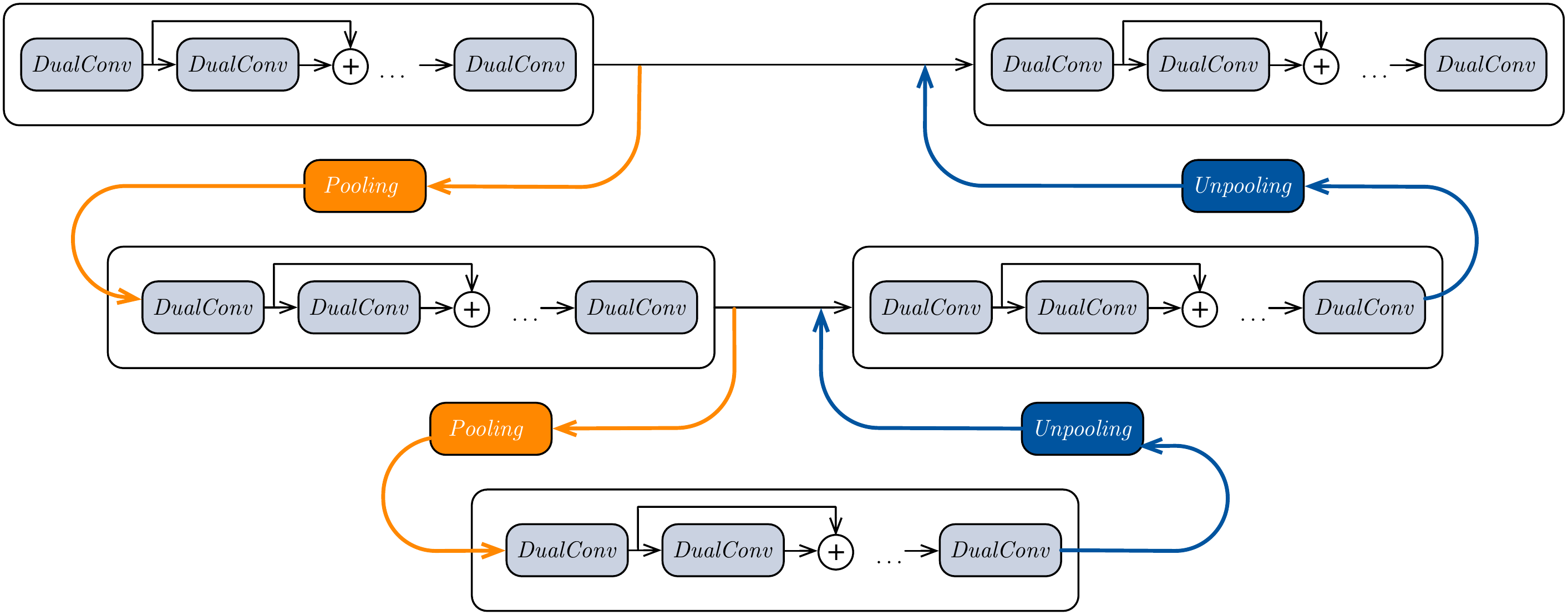}
  \caption{\textbf{Our deep hierarchical architecture} comprises several dual convolutions by-passed by skip connections in each mesh level and performs (un)pooling with pooling trace maps generated from mesh simplification algorithms.
  }
  \label{fig:architecture}
\end{figure*}

\section{Method}
\label{sec:method}
\vspace{-4pt}%
We propose a novel family of deep hierarchical network architectures.
\nameshort{}s combine the previously mentioned benefits of geodesic graph convolutions on 3D surface meshes and Euclidean graph convolutions on 3D vertices in the spatial domain.
An important feature of our proposed architecture family is its modularity, which allows us to measure the effects of all components individually.
To apply geodesic graph convolutions on multiple mesh levels, we describe the necessary mesh-centric pooling operations, \ie, our extensions to vertex clustering and Quadric Error Metrics.
The input to our method is a 3D mesh with vertex features, \eg, color and normals, and the outputs are learned features for each vertex of the input mesh, which are used for dense prediction tasks such as semantic segmentation.

\PAR{Network architecture.}
Inspired by U-Net\,\cite{Ronneberger15MICCAI}, our model is defined as an encoder-decoder architecture, where the encoder is symmetric to the decoder, including skip-connections between both.
Our deep hierarchical architecture is depicted in \reffig{architecture}.
At each mesh level $\mathcal{M}^{\ell}$, multiple \emph{dual convolutions} are applied. Dual convolutions perform \geo{} and \euc{} convolutions in parallel, and subsequently concatenate the resulting feature maps. 
As suggested by He \etal\cite{He2016CVPR}, we add residual connections such that gradients can by-pass the convolutions for better convergence.
For pooling, we leverage vertex clustering and Quadric Error Metrics.

\refsec{ablation_study} will present an ablation study, in which we disable each convolution type individually in order to measure its impact. 
We refer to an instantiation of our network that only operates in a single space as \textbf{Single}ConvMesh-Net (SCM-Net), whereas our full model operating in both spaces simultaneously is referred to as \textbf{Dual}ConvMesh-Net (DCM-Net). Note that \singlearch{}s are a subset of the family of \nameshort{}s  since they equal \dualarch{} if the number of filters is set to $0$ everywhere for one of the spaces.
An in-depth description including filter sizes and activation functions of both the \singlearch{} and \dualarch{} architecture is given in the supplementary material.

\PAR{Euclidean and geodesic graph convolutions.}
We perform graph convolutions on the graph 
$\mathcal{G}^{\ell}$\,=\,$\left(\mathcal{V}^{\ell}, E^{\ell}\right) $
induced by the underlying mesh $\mathcal{M}^{\ell}$ in hierarchy level~$\ell$.
The vertices of level $\ell$ are embedded in Euclidean 3D space, \ie, $\mathcal{V}^{\ell}$\,=\,$\mathbb{R}^3$.
The edge set $E^{\ell}$\,=\,$E_g^{\ell} \cup E_e^{\ell}$ is the union of the geodesic edge set $E_g^{\ell}$, induced by the faces of $\mathcal{M}^{\ell}$, and the Euclidean edge set $E_e^{\ell}$, obtained from the $k$-nn or radius graph neighborhood of each vertex $\vec{v}_i^\ell \in V^{\ell}$.
Note that we neglect the level superscript $\ell$ when it eases readability.
We implement convolutional layers over point features $\vec{x}_i$ associated with vertex $\vec{v}_i$ as for example in \emph{EdgeConv}\,\cite{Wang18CoRR} or \emph{FeaStNet}\,\cite{Verma2018CVPR}. Specifically, the output feature $\vec{y}_i \in \mathbb{R}^E$ of vertex $\vec{v}_i$ with input feature $\vec{x}_i \in \mathbb{R}^{F}$ is computed as
\vspace{-4pt}
\begin{equation}
\label{eq:edgeconv}
\vec{y}_i = \frac{1}{|\mathcal{N}_i|} \sum_{j \in \mathcal{N}_i} \varphi \left([\vec{x}_i, \vec{x}_j - \vec{x}_i]; \theta\right)
\end{equation}
where $\mathcal{N}_i$ is the geodesic/Euclidean neighborhood of the vertex $\vec{v}_i$, $| \mathcal{N}_i |$ its cardinality, $\varphi$ is a nonlinear function implemented as an MLP with trainable parameters $\theta$ and $[\,\cdot\, , \,\cdot\,]$ is the concatenation. Note that the number of kernel parameters $\theta$ is independent of the kernel size induced by the neighborhood $\mathcal{N}_i$. This is in contrast to 2D CNNs, where the number of parameters increases with the kernel size. By normalizing with $|\mathcal{N}_i|$, the convolutional layer is robust to variations in the number of neighbors.

On the very first convolution in the network, we define a translation invariant version of our convolutional layer which relies only on edge information.
Specifically, we apply $\varphi(\cdot)$ to $\vec{x}_j - \vec{x}_i$ and do not concatenate the initial features containing absolute positions, (\cf\,\refeq{edgeconv}). This makes it possible to train on scene crops, but evaluate on full rooms, which leads to broader context information for each vertex and decreases runtime during evaluation.

In contrast to DGCNN\,\cite{Wang18CoRR}, we do not recalculate the neighborhoods in the learned feature space but we reuse the initial neighborhoods in the Euclidean and geodesic spaces.
Skipping this dynamic recalculation of neighbors allows us to create deeper graph convolutional networks while enabling faster and more memory-efficient computations.

Alternative convolutional layers defined over relative vertex \emph{positions} may also be used, such as \emph{PointConv}\,\cite{Wu18CVPR} or \emph{DPCC}\,\cite{Wang18CVPRa}.
However, in this work we focus on the neighborhood $\mathcal{N}_i$ which differentiates \emph{\geo{}} convolutions from \emph{\euc{}} ones (see\,\reffig{motivation}):
\textit{Geodesic graph convolutions} define the \geo{} neighborhood $\mathcal{N}_i^G$ of a vertex $\vec{v}_i$ as the $1$-\,hop neighborhood, \ie, all points that are reachable from the center vertex by one edge connection along the surface mesh. As such, the geodesic neighborhood $\mathcal{N}_i^G$ contains only points in the localized geodesic proximity of vertex $\vec{v}_i$.
\textit{Euclidean graph convolutions} rely on the \euc{} neighborhood $\mathcal{N}_i^E$ of a vertex $\vec{v}_i$ that is only constrained by the Euclidean distance.
In this work, we obtain the Euclidean neighborhood $\mathcal{N}_i^E$ using a $k$-nn or radius graph. We compare both approaches in\,\refsec{ablation_study}.

\PAR{Random Edge Sampling (RES).}
Hermosilla \etal\cite{Hermosilla2018Graph} argue that radius neighborhoods increase the robustness to non-uniformly sampled point clouds in contrast to $k$-nn ones. 
Since the simplification with QEM does not guarantee uniformly sampled mesh simplification, we rely on radius neighborhoods.
However, radius neighborhoods may lead to arbitrarily many neighbors.
We thus resort to sampling methods for reducing the computational load.

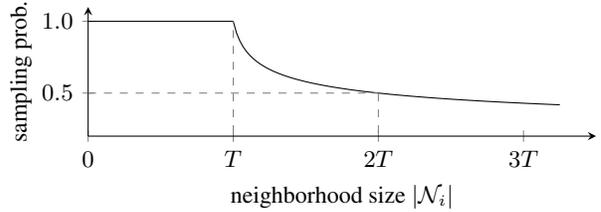
\begin{figure}
	\centering
	\begin{tikzpicture}%
    \begin{axis}[
        height=3.3cm,
        width=\linewidth,
        xmin=0, xmax=70,
        ymin=0.2, ymax=1.1,
        samples=100,
          axis lines=left,
        xlabel=neighborhood size $|\mathcal{N}_i|$,
        ylabel=sampling prob.,
        y label style={at={(axis description cs:0.05,0.5)}},
        xtick={0, 20, 40, 60},
        xticklabels={$0$, $T$, $2T$, $3T$},
        ytick={0.5, 1.0},
        yticklabels={$0.5$, $1.0$},
        label style={font=\small},
        tick label style={font=\small}  
    ]
        \addplot+[domain=20:65, mark=none, black] {x + 1 - 20)^(- 1 / (ln(20+1) / ln(2)))};
        \addplot+[domain=0:20, mark=none, black] {1};
        \addplot +[mark=none, dashed, gray] coordinates {(20, 0) (20, 1.)};
        \addplot +[mark=none, dashed, gray] coordinates {(0, 0.5) (40, 0.5)};
        \addplot +[mark=none, dashed, gray] coordinates {(40, 0) (40, 0.5)};
    \end{axis}%
    \end{tikzpicture}%
    \vspace{-10px}%
  \caption{\small \textbf{Sampling probabilities for Random Edge Sampling}.
  Using the function $D(\mathcal{N_i})$, RES only samples edges interconnecting the vertex $\vec{v}_i$ with neighboring vertices $\vec{v}_j \in \mathcal{N}_i$ if the neighborhood set $\mathcal{N}_i$ does not exceed the threshold $T$.
  }
  \label{fig:keep_probability}
  \vspace{-2px}%
\end{figure}

Motivated by Dropout\,\cite{srivastava2014dropout}, we define a novel sampling method on graph neighborhoods, called \emph{random edge sampling (RES)}.
RES randomly samples edges from the Euclidean edge set on all mesh levels.
We define a function $D: \mathcal{N}_i \rightarrow [0,1]$ which maps the vertex neighborhood $\mathcal{N}_i$ of a given vertex $\vec{v}_i$ to its corresponding sampling probability, which is subsequently applied to all edges between vertices $\vec{v}_i$ and $\vec{v}_j \in \mathcal{N}_i$.
$D$ is defined as follows:
\vspace{-6px}
\begin{equation}
\label{eq:red}
\hspace{-8px}D(\mathcal{N}_i) = \begin{cases}
1 &\text{if }|\mathcal{N}_i| \leq T \\
\left(|\mathcal{N}_i| - \left(T-    1\right)\right)^{-{\textrm{ld}\left({T+1}\right)}^{-1}} &\text{if }|\mathcal{N}_i| > T
\end{cases}
\end{equation}
Only edges $\left(\vec{v}_i, \vec{v}_j\right)$ connecting vertices $\vec{v}_j$ of the vertex neighborhood $\mathcal{N}_i$ whose size exceeds the threshold $T$ are subject to sampling. We argue that the approximation with a neighborhood of small size is already limited and therefore, the neighborhood should not be further decimated.
We visualize $D(\mathcal{N}_i)$ in\,\reffig{keep_probability}.
Varying the threshold $T$ equals to varying the expected number of vertices we draw from the vertex neighborhood distribution.
By doing so, we introduce a larger variety to the training data and thus increase the generalization capability of our approach, while simultaneously reducing the computational load.
We experience that decreasing the threshold for training still leads to a good approximation of the neighborhood.

\PAR{Mesh simplification as a means of pooling.}
\label{voxelhashing}
We interpret pooling operations as generating a hierarchy of mesh levels $(\mathcal{M}^0, ..., \mathcal{M}^\ell, ..., \mathcal{M}^\mathcal{L})$ of increasing simplicity interlinked by \emph{pooling trace maps} $(\mathcal{T}^0, ...,\mathcal{T}^\ell, ..., \mathcal{T}^{\mathcal{L}-1})$ (see\,\reffig{pooling}).
$\mathcal{M} ^ 0 $ is the mesh at its original resolution and $\mathcal{M} ^ \mathcal{L}$ is the coarsest representation after the final simplification operation.
A pooling trace map $\mathcal{T}^\ell$ maps the elements of a vertex partition $\{\vec{v}^{\ell}_i \}$\,$\subset$\,$\mathcal{V}^{\ell}$
bijectively to a single representative vertex $\vec{v}^{\ell+1}$\,$\in$\,$\mathcal{V}^{\ell+1}$ in the next mesh level $\ell$\,$+$\,$1$.
Vertices of the mesh level $\mathcal{M}^{\ell+1}$ are interconnected by the edge set $E_g^{\ell+1}$ obtained by the mesh simplification algorithm.
Similar to\,\cite{Pan2018CoRR}, on the features of $\{\vec{v}^{\ell}_i \}$, we propose to apply permutation invariant aggregation functions, \eg, $\textrm{sum}(\cdot)$, $\textrm{max}(\cdot)$ or $\textrm{mean}(\cdot)$.
To obtain pooled features for $\vec{v}^{\ell+1}$, we use \emph{mean} aggregation for our experiments, in accordance to the definition of graph convolutions (Eq.\,\ref{eq:edgeconv}).

As two well-approved methods from the geometry processing domain, we extend Vertex Clustering (VC)\,\cite{Rossignac1993Multiresolution3A} and Quadric Error Metrics (QEM)\,\cite{Garland1997CGIT} with pooling trace maps to achieve (un)pooling through simple look-up operations.

We modify the VC approach as follows:
We place a 3D uniform grid with cubical cells of a fixed side length $s$ over the input graph and group all vertices that fall into the same cell.
We define  $\vec{v}^{\ell+1}$ as the centroid $\vec{v}^{\ell+1} = |\{ \vec{v}^{\ell}_i \}|^{-1}  \sum \vec{v}^{\ell}_i $.
Moreover, we store the mapping between the representative vertex $\vec{v}^{\ell+1}$ and its corresponding vertices $\{ \vec{v}^{\ell}_i \}$ in the pooling trace map.
A similar approach is followed in\,\cite{Tatarchenko18CVPR}.
However, in order to perform \geo{} convolutions on pooled graphs, the surface information, \ie, the \emph{edges}, needs to be preserved as well.
To achieve this, we first delete all edges between vertices that fall into the same cell, then we connect the representative vertices of those cells that were previously connected with at least one edge.
Although the cell size~$s$ performs a low-pass filtering of the mesh vertex density and furthermore limits the introduced geometric error, this method is sensitive to the exact placement and orientation of the grid.

Alternatively, we consider QEM\,\cite{Garland1997CGIT}.
In contrast to VC, this approach incrementally contracts vertex pairs $(\vec{v}_1, \vec{v}_2)$ to a new representative $\bar{\vec{v}}$ according to an approximate error of the geometric distortion this contraction introduces.
We keep track of these contractions and thus are able to generate pooling trace maps.
Since QEM performs vertex contraction, we may contract vertices which are not adjacent in the mesh. This compensates for small scanning artifacts.

The same trace maps are used for unpooling a mesh from $\mathcal{M}^{\ell+1}$ to $\mathcal{M}^{\ell}$ by copying the features of $\vec{v}^{\ell+1}$ to its corresponding vertex $\vec{v}_i^{\ell}$. As VC aims for uniform vertex density and QEM for minimal geometric distortion, we compare both approaches in our ablation study in\,\refsec{ablation_study}.

\begin{table}[t!]
\centering
\resizebox{\columnwidth}{!}{%
\begin{tabular}{rcc|c}
\toprule 
\multirow{2}{*}{Method} & \multicolumn{2}{c|}{mIoU} & \multirow{2}{*}{\shortstack{Convolutional\\Category}}\\ 
&ScanNet&S3DIS\\
\toprule 
PointNet\,\cite{Qi17CVPR}				 	 & - & $41.1$ & \multirow{3}{*}{\shortstack{Permutation\\Invariant\\Networks}}\\
PointNet++\,\cite{Qi17NIPS} 				 	 & $33.9$ & - \\
FCPN\,\cite{Rethage18ECCV} 		 		 & $44.7$ & - \\
\hline
3DMV\,\cite{Dai18ECCV} & $48.3$ & -  & \multirow{3}{*}{2D-3D}\\	
JPBNet\,\cite{chiang2019unified} & $63.4$ & - \\
MVPNet\,\cite{Jaritz_2019_ICCV} & $64.1$ & $62.4$\\
\hline
TangentConv\,\cite{Tatarchenko18CVPR}& $43.8$ & $52.6$ & \multirow{3}{*}{SurfaceConv}\\
SurfaceConvPF*\,\cite{Pan2018CoRR} & $44.2$ & - \\
TextureNet\,\cite{Huang2018CoRR} & $56.6$ & - \\	
\hline
PointCNN\,\cite{Li18NIPS} &$45.8$ & $57.3$ & \multirow{5}{*}{PointConv}\\
ParamConv\,\cite{Wang18CVPRa} & - & $58.3$\\
DPC\,\cite{Engelmann20ICRA} & 59.2 & 61.3\\
MCCN\,\cite{Hermosilla2018Graph} & $63.3$ & -  \\
PointConv\,\cite{Wu18CVPR} & $66.6$ & - \\
KPConv\,\cite{thomas2019ICCV}& $68.4$ & $\mathbf{67.1}$ \\
\hline
SparseConvNet\,\cite{Graham18CVPR} & $72.5$ & - & \multirow{2}{*}{\shortstack{Voxelized\\SparseConv}}\\
MinkowskiNet\,\cite{Choy2019CVPR} & $\mathbf{73.4}$ & $65.3$  \\
 \hline \hline
DeepGCN\,\cite{li2019deepgcns} & - & $52.5$ & \multirow{5}{*}{GraphConv}\\
SPGraph\,\cite{Landrieu17CVPR}	        & - & $58.0$ \\
SPH3D-GCN*\,\cite{lei2019spherical} & $61.0$ & $59.5$ \\
HPEIN\,\cite{hpein_iccv19} & $61.8$ & $61.9$ \\
\nameshort{} \textbf{(Ours)} & $\mathbf{65.8}$ & $\mathbf{64.0}$ \\
\bottomrule
\end{tabular}
}
\vspace{-6px}
\caption{\small \textbf{Comparison to state-of-the-art.} Semantic segmentation mIoU scores on the offical ScanNet benchmark~\cite{Dai17CVPR} and S3DIS Area-5~\cite{Armeni16CVPR}.
We outperform other graph convolutional approaches on all benchmarks. * indicates concurrent work. Full network definitions in the supplementary.
ScanNet benchmark was accessed on 11/15/2019. S3DIS results as reported in original publications.}
\vspace{0pt}
\label{tab:scannet_test}
\end{table}
\definecolor{unlabeled}{rgb}{0., 0., 0.}
\definecolor{wall}{rgb}{0.68235294, 0.78039216, 0.90980392}
\definecolor{floor}{rgb}{0.59607843, 0.8745098 , 0.54117647}
\definecolor{cabinet}{rgb}{0.12156863, 0.46666667, 0.70588235}
\definecolor{bed}{rgb}{1.        , 0.73333333, 0.47058824}
\definecolor{chair}{rgb}{0.7372549 , 0.74117647, 0.13333333}
\definecolor{sofa}{rgb}{0.54901961, 0.3372549 , 0.29411765}
\definecolor{table}{rgb}{1.        , 0.59607843, 0.58823529}
\definecolor{door}{rgb}{0.83921569, 0.15294118, 0.15686275}
\definecolor{window}{rgb}{0.77254902, 0.69019608, 0.83529412}
\definecolor{bookshelf}{rgb}{0.58039216, 0.40392157, 0.74117647}
\definecolor{picture}{rgb}{0.76862745, 0.61176471, 0.58039216}
\definecolor{counter}{rgb}{0.09019608, 0.74509804, 0.81176471}
\definecolor{desk}{rgb}{0.96862745, 0.71372549, 0.82352941}
\definecolor{curtain}{rgb}{0.85882353, 0.85882353, 0.55294118}
\definecolor{refrigerator}{rgb}{1.        , 0.49803922, 0.05490196}
\definecolor{showercurtain}{rgb}{0.61960784, 0.85490196, 0.89803922}
\definecolor{toilet}{rgb}{0.17254902, 0.62745098, 0.17254902}
\definecolor{sink}{rgb}{0.43921569, 0.50196078, 0.56470588}
\definecolor{bathtub}{rgb}{0.89019608, 0.46666667, 0.76078431}
\definecolor{otherfurn}{rgb}{0.32156863, 0.32941176, 0.63921569}

\begin{figure}[t]
\centering
\includegraphics[width=0.33\linewidth, trim={0 0  0 0}, clip]{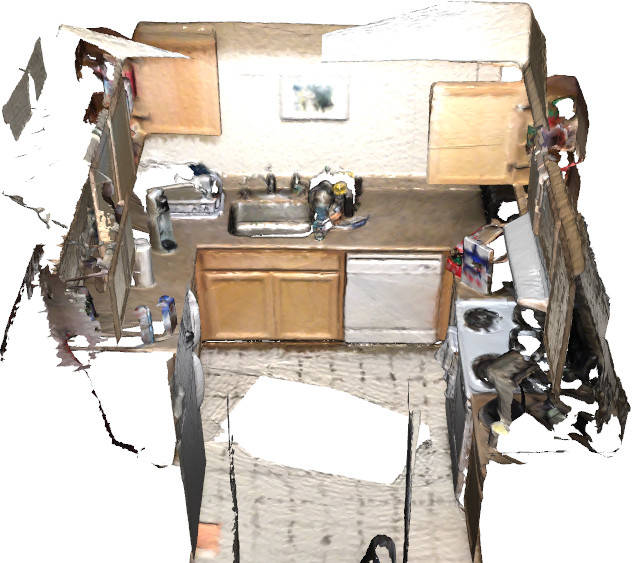}\hfill%
\includegraphics[width=0.33\linewidth, trim={0 0  0 0}, clip]{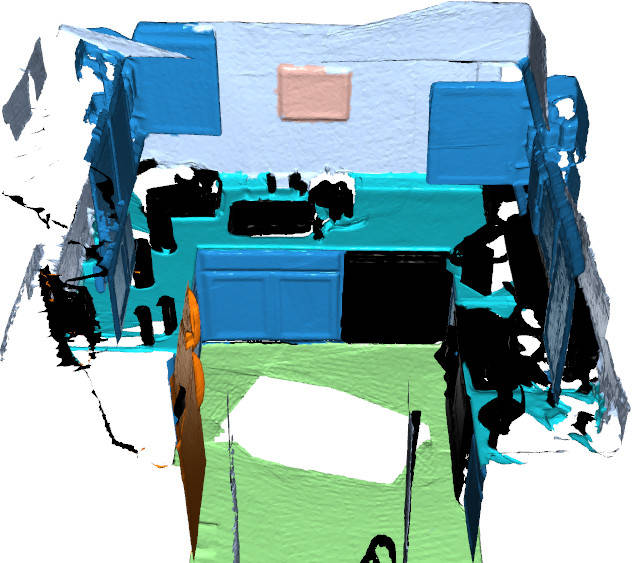}\hfill%
\includegraphics[width=0.33\linewidth, trim={0 0 0 0}, clip]{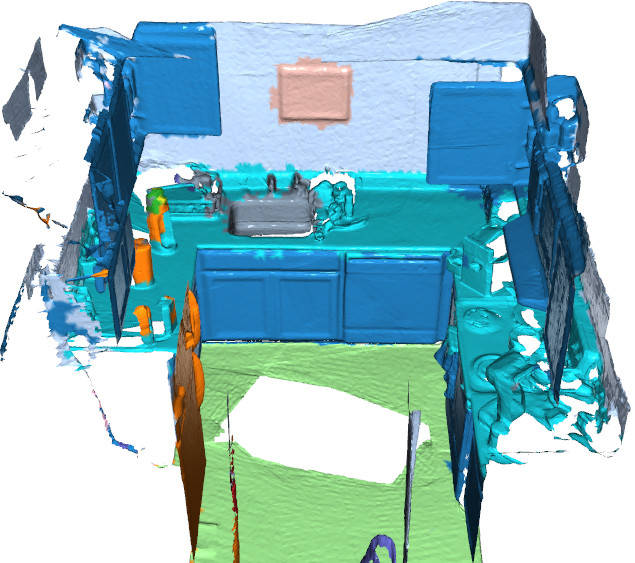}

\includegraphics[width=0.25\linewidth, trim={0 0  0 0}, clip]{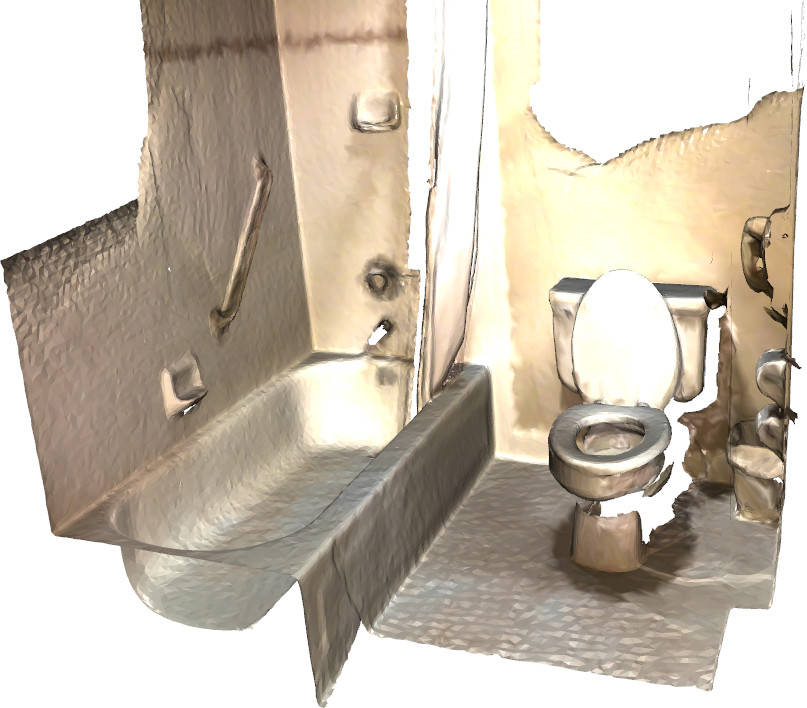}\hfill%
\includegraphics[width=0.25\linewidth, trim={0 0  0 0}, clip]{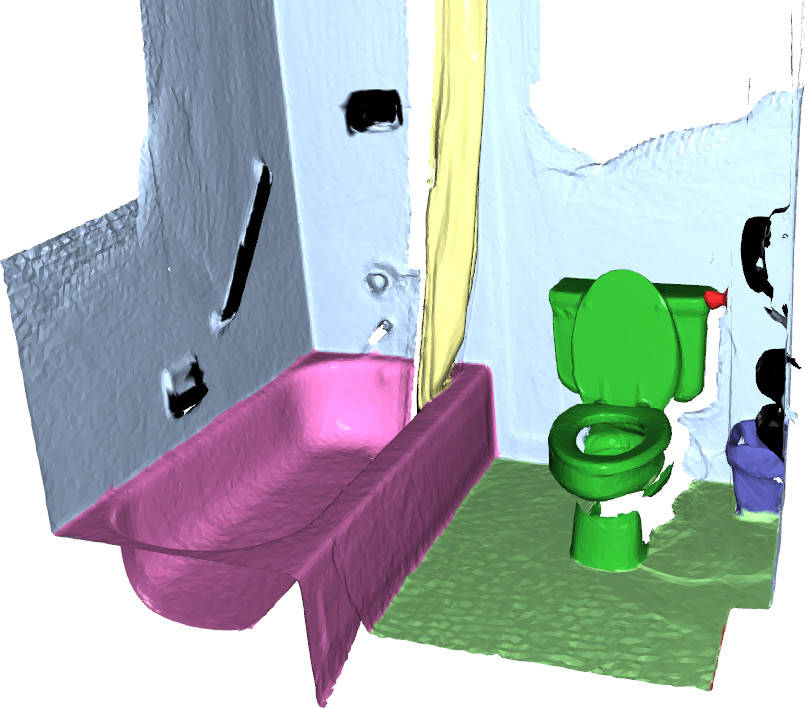}\hfill%
\includegraphics[width=0.25\linewidth, trim={0 0  0 0}, clip]{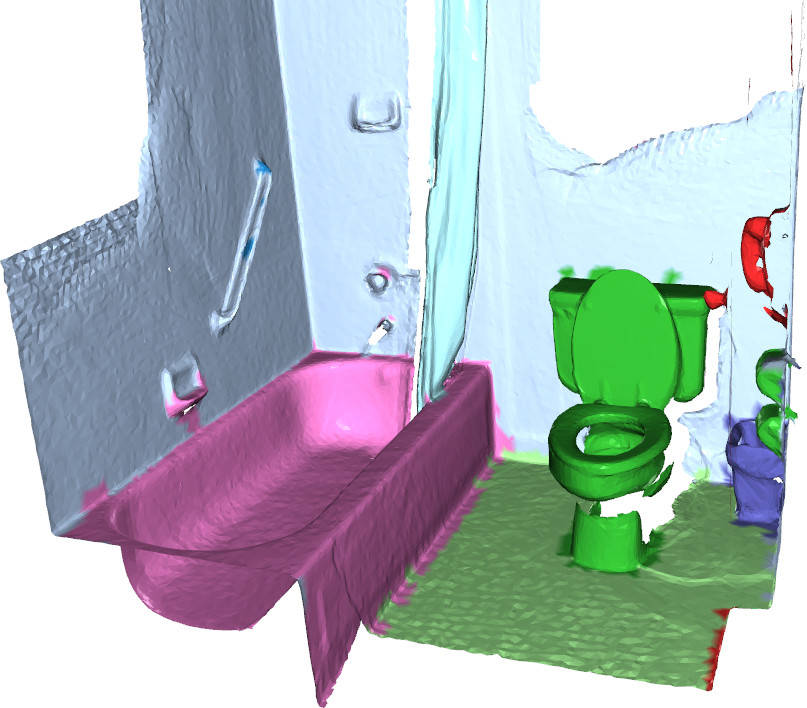}

\includegraphics[width=0.33\linewidth, trim={0 0  0 0}, clip]{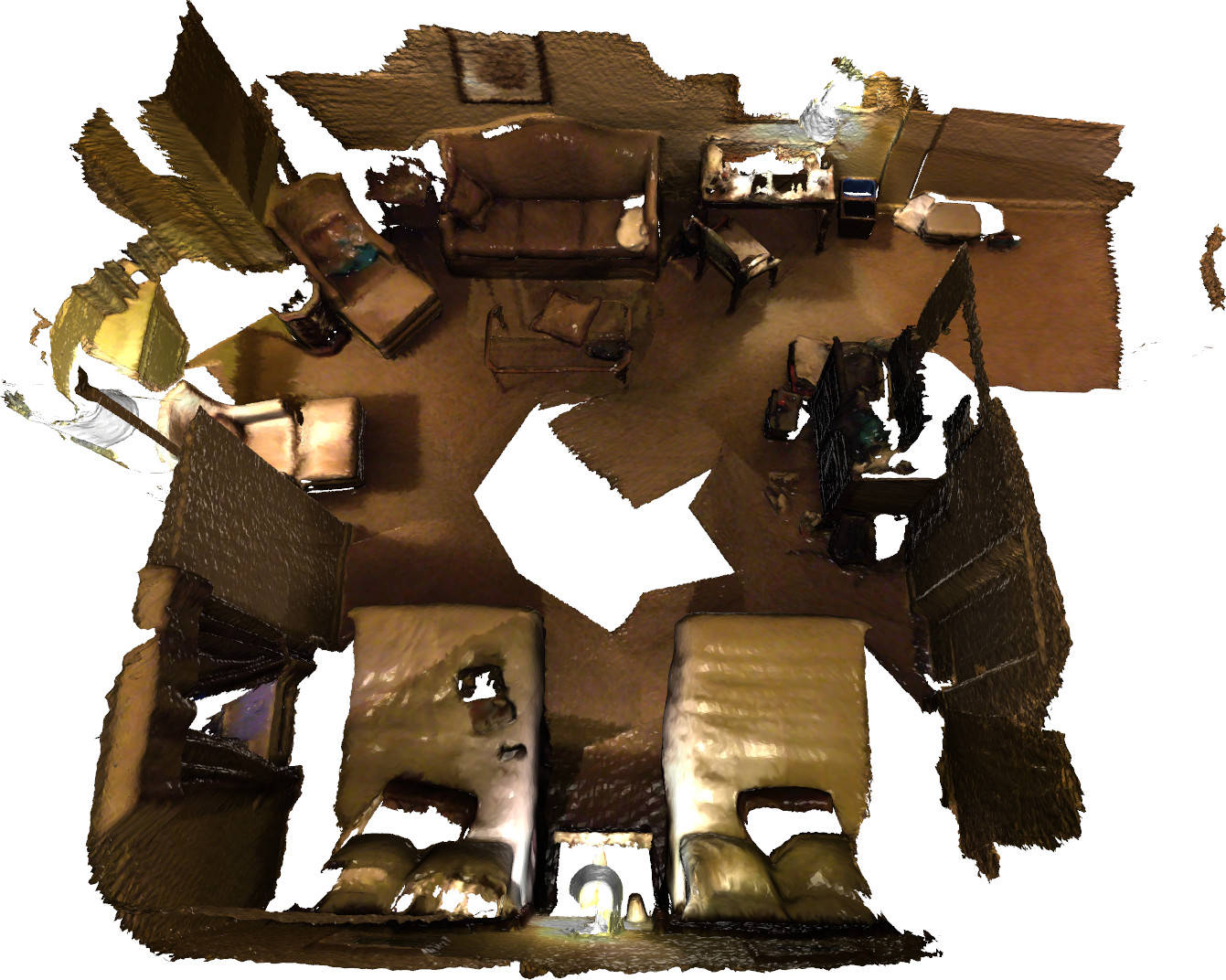}\hfill%
\includegraphics[width=0.33\linewidth, trim={0 0  0 0}, clip]{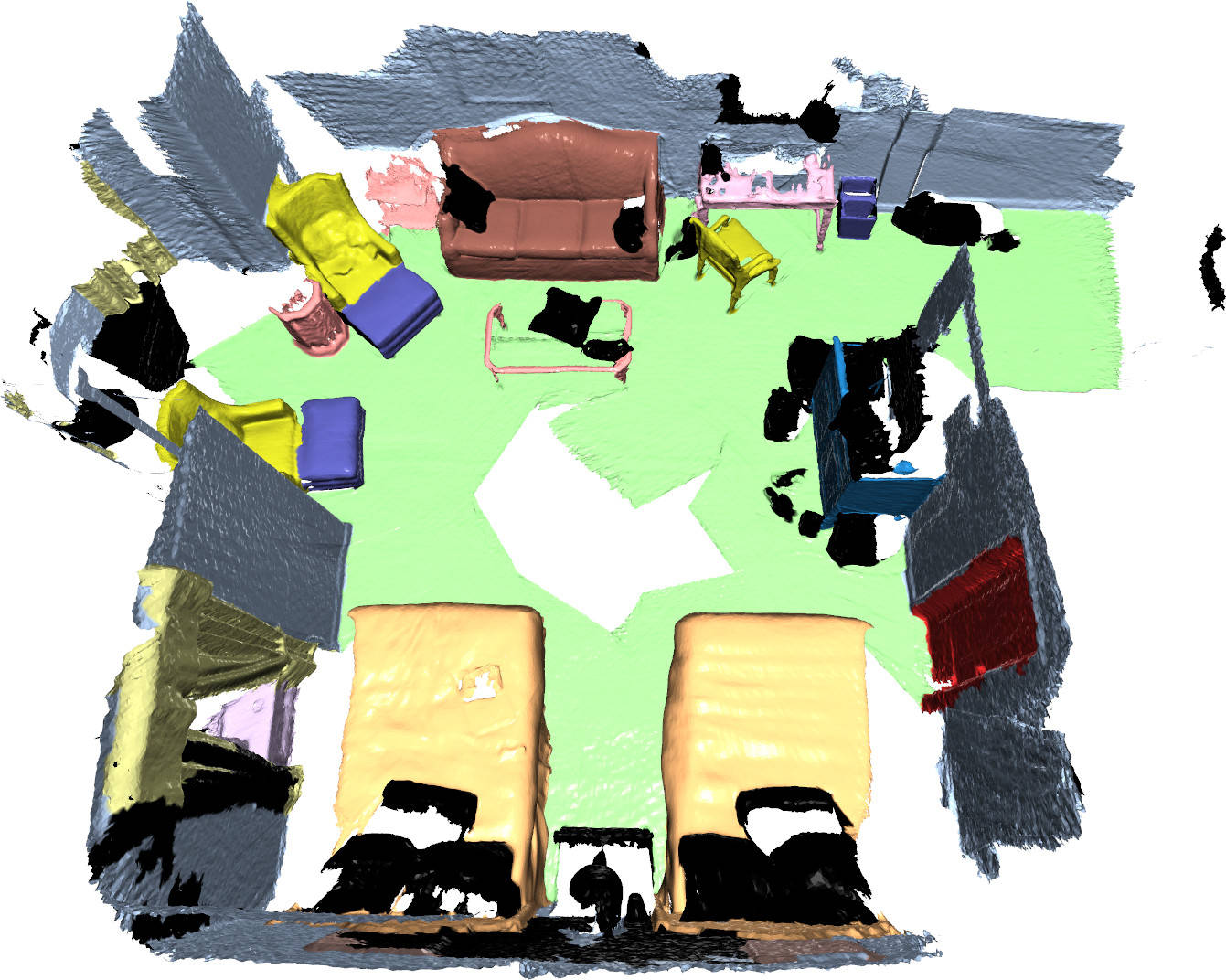}\hfill%
\includegraphics[width=0.33\linewidth, trim={0 0  0 0}, clip]{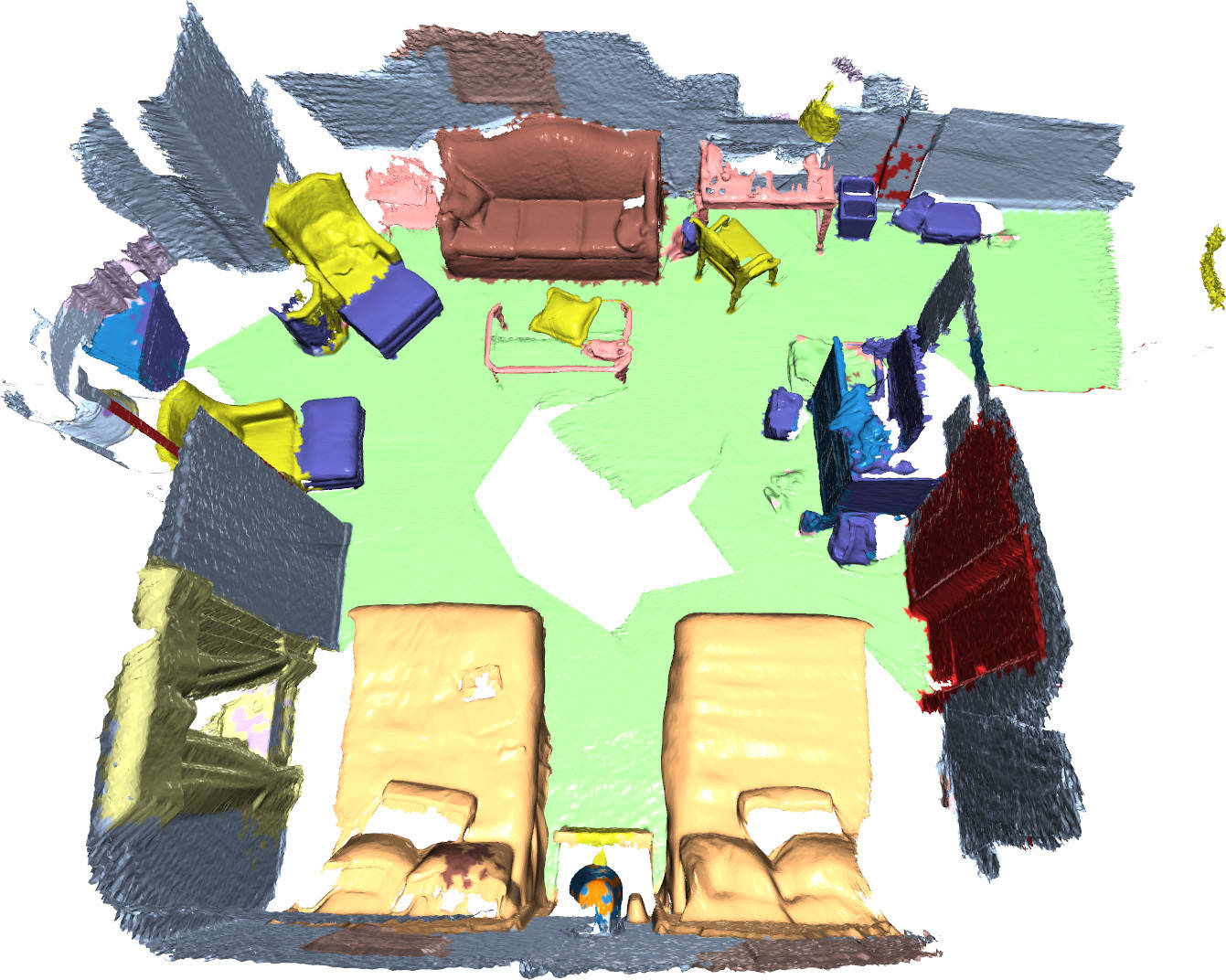}

\includegraphics[width=0.33\linewidth, trim={0 20  0 0}, clip]{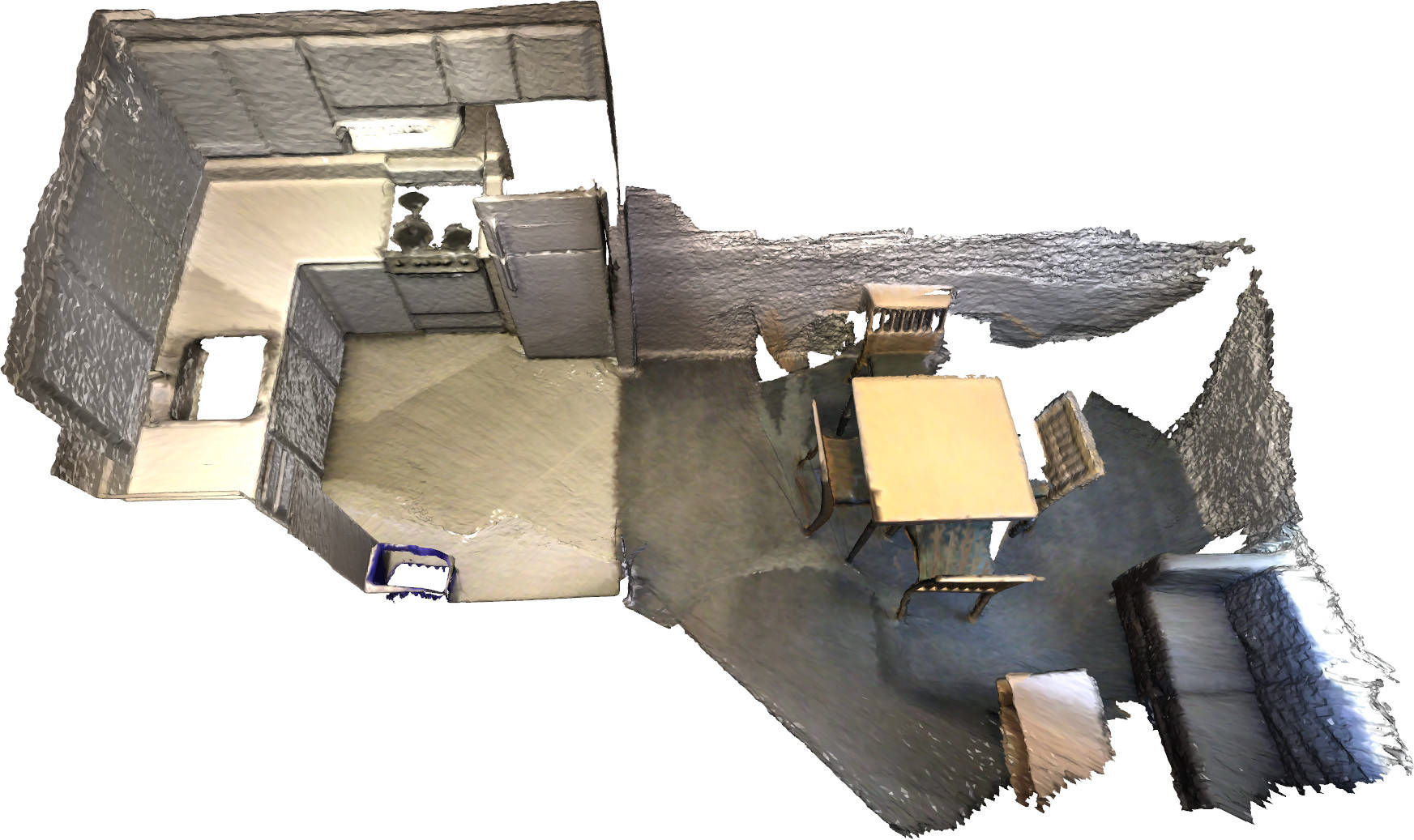}\hfill%
\includegraphics[width=0.33\linewidth, trim={0 20  0 0}, clip]{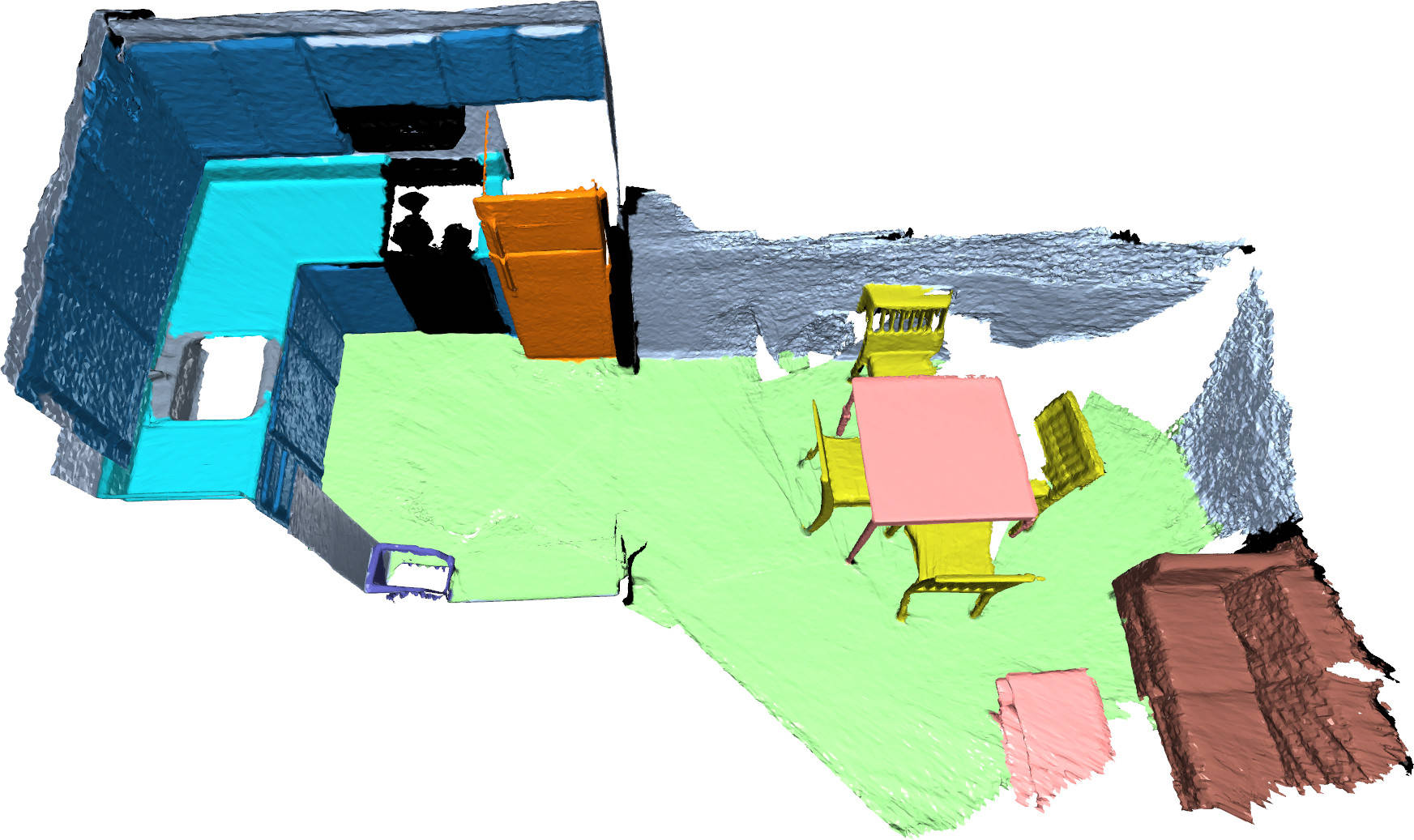}\hfill%
\includegraphics[width=0.33\linewidth, trim={0 20  0 0}, clip]{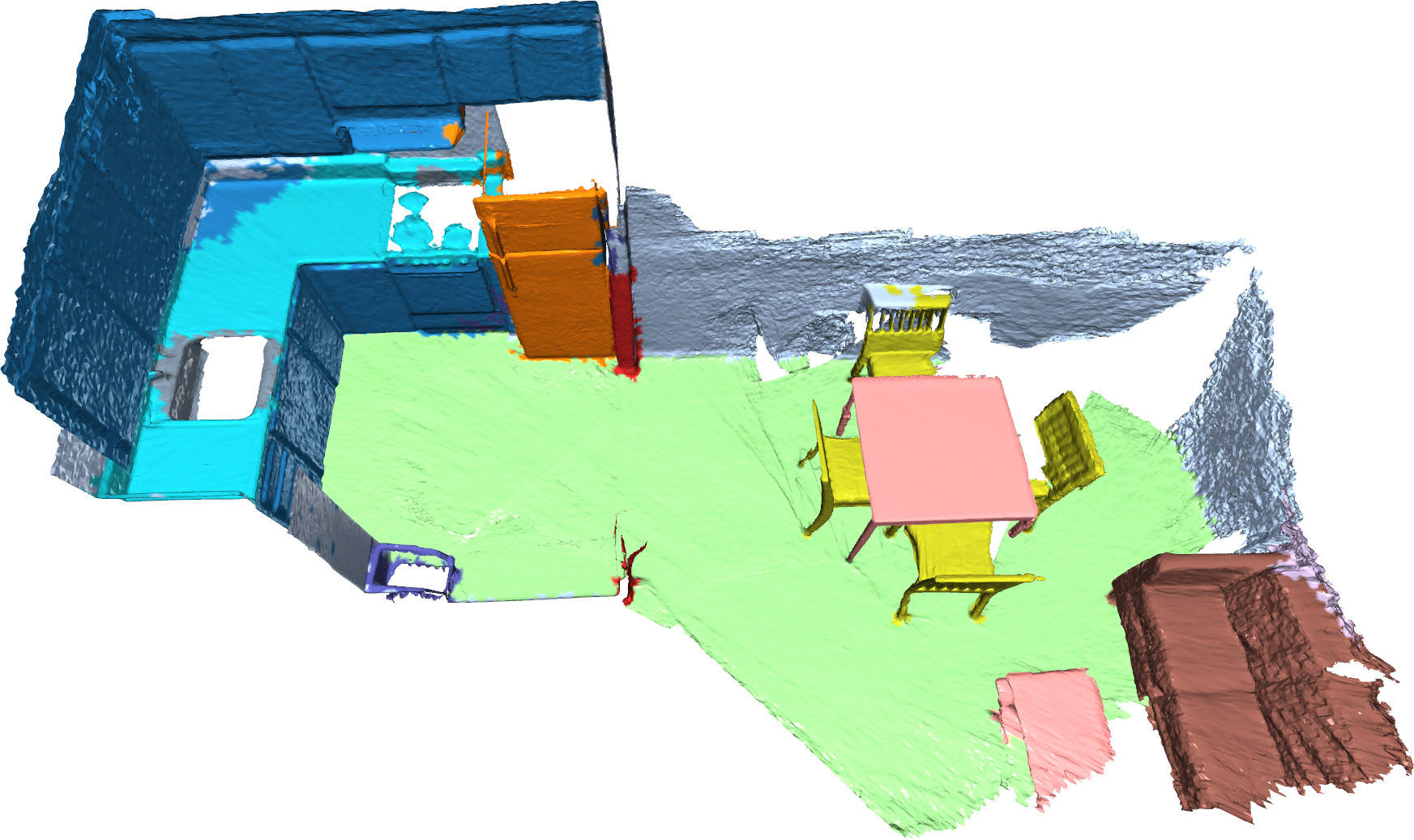}
\begin{small}
\begin{tabular}{ccc}
\textbf{Input Mesh} & \textbf{Ground Truth} & \textbf{Prediction} \\
\hspace{0.30\linewidth} & \hspace{0.25\linewidth} & \hspace{0.25\linewidth} \\
\end{tabular}
\\
\vspace{-10pt}
\textcolor{unlabeled}{\ColorMapCircle}~unlabeled
\textcolor{wall}{\ColorMapCircle}~wall
\textcolor{floor}{\ColorMapCircle}~floor
\textcolor{cabinet}{\ColorMapCircle}~cabinet
\textcolor{bed}{\ColorMapCircle}~bed
\textcolor{chair}{\ColorMapCircle}~chair
\textcolor{sofa}{\ColorMapCircle}~sofa
\textcolor{table}{\ColorMapCircle}~table
\textcolor{door}{\ColorMapCircle}~door
\textcolor{picture}{\ColorMapCircle}~picture
\textcolor{counter}{\ColorMapCircle}~counter
\textcolor{desk}{\ColorMapCircle}~desk
\textcolor{curtain}{\ColorMapCircle}~curtain
\textcolor{refrigerator}{\ColorMapCircle}~fridge
\textcolor{showercurtain}{\ColorMapCircle}~shower curtain
\textcolor{toilet}{\ColorMapCircle}~toilet
\textcolor{sink}{\ColorMapCircle}~sink
\textcolor{bathtub}{\ColorMapCircle}~bathtub
\textcolor{otherfurn}{\ColorMapCircle}~otherfurn
\end{small}
\\
\vspace{-3px}
\caption{\textbf{Results on ScanNet\,v2 validation~\cite{Dai17CVPR}.}
Our method correctly predicts challenging classes such as images and curtains, while maintaining clear boundaries.
In the second row, our method correctly predicts shower\,\textcolor{showercurtain}{\ColorMapCircle} curtain even though the ground truth is falsely labeled as regular curtain\,\textcolor{curtain}{\ColorMapCircle}.
Similarly in row three the partial ground truth label of the bottom right corner is properly predicted fully.
There are some reasonable mistakes like the desk\,\textcolor{desk}{\ColorMapCircle} in row three labeled as table\,\textcolor{table}{\ColorMapCircle}. }
\label{fig:scannet_quali}


\end{figure}

\begin{table*}[t]
\resizebox{\textwidth}{!}{%
\setlength{\tabcolsep}{1pt}
\begin{tabular}{rc|cccccccccccccccccccccc}
        \toprule
        Method & mAcc & wall & floor & cab & bed & chair & sofa & table & door & wind & shf & pic & cntr & desk & curt & ceil & fridg & show & toil & sink & bath & other\\
        \midrule
        PointNet++\,\cite{Qi17NIPS} & $43.8$ & $80.1$ & $81.3$ & $34.1$ & $71.8$ & $59.7$ & $63.5$ & $\mathbf{58.1}$ & $49.6$ & $28.7$ & $1.1$ & $34.3$ & $10.1$ & $0.0$ & $68.8$ & $79.3$ & $0.0$ & $29.0$ & $70.4$ & $29.4$ & $62.1$ & $8.5$ \\
        SplatNet\,\cite{Su18CVPR} & $26.7$ & $\mathbf{90.8}$ & $\mathbf{95.7}$ & $30.3$ & $19.9$ & $\mathbf{77.6}$ & $36.9$ & $19.8$ & $33.6$ & $15.8$ & $15.7$ & $0.0$ & $0.0$ & $0.0$ & $12.3$ & $75.7$ & $0.0$ & $0.0$ & $10.6$ & $4.1$ & $20.3$ & $1.7$ \\
        TangentConv\,\cite{Tatarchenko18CVPR} & $46.8$ & $56.0$ & $87.7$ & $41.5$ & $73.6$ & $60.7$ & $69.3$ & $38.1$ & $55.0$ & $30.7$ & $33.9$ & $50.6$ & $38.5$ & $19.7$ & $48.0$ & $45.1$ & $22.6$ & $35.9$ & $50.7$ & $49.3$ & $56.4$ & $16.6$ \\
        3DMV\,\cite{Dai18ECCV} & $56.1$ & $79.6$ & $95.5$ & $59.7$ & $82.3$ & $70.5$ & $73.3$ & $48.5$ & $64.3$ & $55.7$ & $8.3$ & $55.4$ & $34.8$ & $2.4$ & $\mathbf{80.1}$ & $\mathbf{94.8}$ & $4.7$ & $54.0$ & $71.1$ & $47.5$ & $76.7$ & $19.9$  \\
        TextureNet\,\cite{Huang2018CoRR}& $63.0$ & $63.6$ & $91.3$ & $47.6$ & $82.4$ & $66.5$ & $64.5$ & $45.5$ & $69.4$ & $60.9$ & $30.5$ & $\mathbf{77.0}$ & $\mathbf{42.3}$ & $44.3$ & $75.2$ & $92.3$ & $49.1$ & $\mathbf{66.0}$ & $80.1$ & $\mathbf{60.6}$ & $86.4$ & $27.5$ \\
        \midrule
        \nameshort{} (\textbf{Ours}) & $\mathbf{66.2}$ & $78.4$ & $93.6$ & $\mathbf{64.5}$ & $\mathbf{89.5}$ & $70.0$ & $\mathbf{85.3}$ & $46.1$ & $\mathbf{81.3}$ & $\mathbf{63.4}$ & $\mathbf{43.7}$ & $73.2$ & $39.9$ & $\mathbf{47.9}$ & $60.3$ & $89.3$ & $\mathbf{65.8}$ & $43.7$ & $\mathbf{86.0}$ & $49.6$ & $\mathbf{87.5}$ & $\mathbf{31.1}$\\
        \bottomrule
\end{tabular}
}
\vspace{-6pt}
\caption{\small \label{tab:matterport3d_test_results} \textbf{Mean class accuracy scores on Matterport3D Test~\cite{Matterport3D}.} We outperform other approaches in $11$ out of $21$ classes. We use the same network definition as for the ScanNet v2 benchmark. Scores from~\cite{Huang2018CoRR}.}
\end{table*}

\vspace{-2pt}
\section{Experiments}
\vspace{-3pt}
We evaluate our method on three large scale 3D scene segmentation datasets, which contain meshed point clouds of various indoor scenes.
\label{sec:experiments}

\PAR{Stanford Large-Scale 3D Indoor Spaces (S3DIS)\cite{Armeni16CVPR}} contains dense 3D point clouds from 6 large-scale indoor areas,
consisting of 271 rooms from 3 different buildings.
The points are annotated with 13 semantic classes.
It also includes 3D meshes, which are not semantically annotated.
On average, each mesh contains $2\cdot10^5$ triangular faces\,\cite{Armeni16CVPR}.
As the resolution of these meshes is low compared to ScanNet\,v2 or Matterport3D, we oversample all faces and interpolate the color and ground truth information from the semantically annotated points. Our final predictions are then interpolated to the original point cloud to generate comparable results on the benchmark.
We follow the common train/test split\,\cite{Armeni16CVPR, Wang18CVPRa, Tchapmi173DV} and train on all areas except Area~5, which we keep for testing. we provide cross-validation mean IoU scores in the supplementary material.

\PAR{ScanNet\,v2 Benchmark\,\cite{Dai17CVPR}.}
We furthermore evaluate our architecture on the ScanNet\,v2 benchmark dataset.
ScanNet contains 3D meshed point clouds of a wide variety of indoor scenes with reconstructed surfaces, textured meshes, and semantic ground truth annotations.
The dataset contains 20 valid semantic classes.
We perform all our experiments using the public training, validation, and test split of 1201, 312 and 100 scans, respectively.
To validate our proposed components, the ablation study is conducted on the ScanNet validation set, where we report mean IoU scores.

\parag{Matterport3D\,\cite{Matterport3D}.}
Similar to ScanNet\,v2, Matterport3D contains meshed reconstructions of 90 building-scale RGB-D scans. We use the same evaluation protocol as introduced in 3DMV\,\cite{Dai18ECCV} and TextureNet\,\cite{Huang2018CoRR} and report mean class accuracy scores on 21 classes on the test set.

\PAR{Implementation and Training Details.}
We use VC and QEM to precompute the hierarchical mesh levels $\mathcal{M}^{\ell}$ interlinked with pooling trace maps $\mathcal{T}^\ell$.
For VC, we set the cubical cell lengths to $4\,\textrm{cm}$, $8\,\textrm{cm}$, $16\,\textrm{cm}$, and $32\,\textrm{cm}$, respectively, for each mesh level.
We experience that directly applying QEM on the full-resolution mesh results in high-frequency signals in noisy areas. Before applying QEM, we therefore first apply VC on the original mesh with a cubical cell length of $4\,\textrm{cm}$.
For each mesh level, QEM simplifies the mesh until the vertex number is reduced to $30\%$ of its preceding mesh level.
As input features, we use the position, color, and normal of each vertex in the mesh. At each mesh level, we perform three dual convolutions (see~\reffig{architecture}).
We train the network end-to-end by minimizing the cross entropy loss using the Adam optimizer \cite{Kingma15ICLR} with an initial learning rate of $10^{-3}$ and exponential learning rate decay of $0.5$ after every $40$ epochs and a batch size of $4$.

It is common practice among recent approaches to discard training samples of low quality. Methods only differ in the used criteria:
Qi \etal\,\cite{Qi17CVPR} reject training examples if the number of points in a training crop falls below a certain threshold. Analogously to our method, \cite{Qi17NIPS} rejects crops when the number of unlabeled points exceeds a threshold of $70\,\%$.
We reject training crops, which have more than $80\,\%$ unlabeled vertices, which corresponds to $0.8\,\%$ of the $18,530$ cropped training samples of the ScanNet\,v2 train set.
We do not apply this filtering during inference.

We conduct our experiments with random edge sampling with threshold $T$\,=\,$15$ while training and $T$\,=\,$25$ while testing, as we observed that a lower threshold for training reduces the computational load of the algorithm while learning useful features.
Since we use a random sampling method for neighborhoods, the predictions vary in each run. We therefore run each evaluation $10$ times and provide mean and standard deviations in our ablation study.
Our models are implemented in PyTorch (Geometric)~\cite{Fey/Lenssen/2019, Paszke2017NIPSW} and trained on a Tesla V100 16GB.

\PAR{Data Augmentation.}
From each mesh in ScanNet\,v2, we obtain $3$\,$\textrm{m}$\,$\times$\,$3$\,$\textrm{m}$ crops with a stride of $1.5$\,$\textrm{m}$ from the ground plane.
Since S3DIS and Matterport3D provide denser meshes, we reduce the crop size to $2$\,$\textrm{m}$\,$\times$\,$2$\,$\textrm{m}$ with a stride of $1$\,$\textrm{m}$.
Each cropped mesh is transformed by a random affine transformation, colors and positions are normalized to the range $[0,1]$.
Despite training on cropped meshes, we can perform inference on full meshes as the model is invariant to absolute vertex positions.

\PAR{Results.} 
\reftab{scannet_test} shows the performance of our approach compared to recent competing approaches on the ScanNet benchmark test dataset as well as S3DIS Area~5, grouped by the approaches' inherent categories.
We are able to report state-of-the-art results for graph convolutional approaches by a significant margin of $4$\,$\%$ mIoU for the ScanNet benchmark, as well as $2.1$\,$\%$ mIoU for S3DIS Area~5.
Only $4$ approaches report better results on ScanNet.
SparseConvNet\,\cite{Graham18CVPR} and MinkowskiNet\,\cite{Choy2019CVPR} use Voxelized Sparse Convolutions, which currently perform best on ScanNet, but which are inherently limited for other tasks in that they cannot make use of detailed surface information.
We also evaluated our algorithm on the novel Matterport3D dataset\,\cite{Matterport3D} and report overall state-of-the-art results on that benchmark in~\reftab{matterport3d_test_results}.
\reffig{scannet_quali} shows our qualitative results on the ScanNet validation set.
In the supplementary material, we provide our results on S3DIS in the $k$-fold test setting,
as well as detailed descriptions of our models.
\begin{figure}
	\centering
\begin{tikzpicture}

\definecolor{color0}{rgb}{0.917647058823529,0.917647058823529,0.949019607843137}
\definecolor{color1}{rgb}{0.298039215686275,0.447058823529412,0.690196078431373}

\begin{axis}[
height=3.1cm,
width=\linewidth,
xlabel={threshold $T$ during inference},
x label style={at={(axis description cs:0.5,0.03)}},
axis lines=left,
xmin=8, xmax=45,
ylabel={mIoU},
y label style={at={(axis description cs:0.03,0.5)}},
ymin=61.5, ymax=64.1,
ytick={62.0, 62.715, 63.811},
yticklabels={$62.0$, $62.7$, $63.8$},
xtick={10, 15, 25, 35},
xticklabels={$10$, $15$, $25$, $35$},
label style={font=\small},
tick label style={font=\small}  
]
\path [draw=color1, fill=color1, opacity=0.2]
(axis cs:10,61.8178333333333)
--(axis cs:10,61.5833055555556)
--(axis cs:15,62.659)
--(axis cs:20,63.341975)
--(axis cs:25,63.63395)
--(axis cs:30,63.625)
--(axis cs:35,63.755)
--(axis cs:40,63.727975)
--(axis cs:45,63.67278)
--(axis cs:50,63.689)
--(axis cs:50,63.803)
--(axis cs:50,63.803)
--(axis cs:45,63.800605)
--(axis cs:40,63.874)
--(axis cs:35,63.868)
--(axis cs:30,63.8)
--(axis cs:25,63.759025)
--(axis cs:20,63.538025)
--(axis cs:15,62.771)
--(axis cs:10,61.8178333333333)
--cycle;
\addplot +[mark=none, dashed, gray] coordinates {(0, 62.715) (15., 62.715)};
\addplot +[mark=none, dashed, gray] coordinates {(15, 62.715) (15., 0)};
\addplot +[mark=none, dashed, gray] coordinates {(0, 63.811) (35., 63.811)};
\addplot +[mark=none, dashed, gray] coordinates {(35, 63.811) (35., 0)};
\addplot [semithick, color1]
table {%
10 61.6966666666667
15 62.715
20 63.444
25 63.703
30 63.71
35 63.811
40 63.8
45 63.7358
50 63.742
};
\end{axis}

\end{tikzpicture}
\vspace{-10px}%
  \caption{\small \textbf{Varying the threshold $T$ during inference}.
  We observe that a smaller number of samples during test is sufficient for learning useful neighborhood features ($T$\,=\,$15$). During test, we gain $1.1$\,$\%$ mIoU by increasing the threshold to $T$\,=\,$35$. (Experiments conducted on S3DIS Area $5$ with $10$ runs for each threshold).
  }
  \label{fig:varying_threshold}
\end{figure}
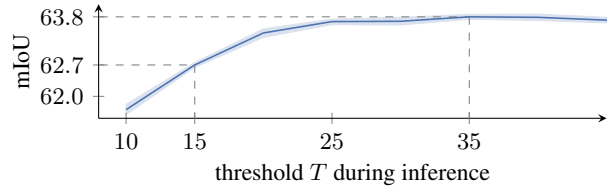
\begin{table}[t]
\centering
\begin{tabular}{ccclr}
\toprule 
pool & arch & neighb & mIoU {\footnotesize($\pm$\,stdev)} & $\Delta$\tabularnewline
\midrule
VC  & \singlearchshort{} & geo     & $57.1$ & \tikzmark[xshift=2.2em]{r}$-0.3$ \tabularnewline
QEM  & \singlearchshort{} & geo     & $56.8$ & \tikzmark[xshift=0em]{s} \tabularnewline
\midrule
VC  & \singlearchshort{} & knn     & $60.1$ & \tikzmark[xshift=2.2em]{x}$+0.8$ \tabularnewline
QEM  & \singlearchshort{} & knn     & $60.9$ & \tikzmark[xshift=0em]{y} \tabularnewline
\midrule
VC  & \singlearchshort{} & rad     & $61.9$ {\footnotesize($\pm$\,$0.20$)} & \tikzmark[xshift=2.2em]{c}$+2.0$ \tabularnewline
FPS & \singlearchshort{} & rad     & $63.5$ {\footnotesize($\pm$\,$0.13$)} & \tikzmark[xshift=2.2em]{a}$+0.4$ \tabularnewline
QEM  & \singlearchshort{} & rad     & $63.9$ {\footnotesize($\pm$\,$0.20$)} &  \tikzmark[xshift=0em]{b}\tabularnewline
\midrule\midrule
VC  & \dualarchshort{} & knn/geo & $59.7$ & \tikzmark[xshift=2.2em]{g}$+3.2$ \tabularnewline
QEM  & \dualarchshort{} & knn/geo & $62.9$ & \tikzmark[xshift=0em]{h} \tabularnewline
\midrule
VC  & \dualarchshort{} & rad/geo & $62.8$ {\footnotesize($\pm$\,$0.12$)} & \tikzmark[xshift=2.2em]{o}$+4.5$ \tabularnewline
QEM  & \dualarchshort{} & rad/geo & $67.3$ {\footnotesize($\pm$\,$0.22$)} & \tikzmark[xshift=0em]{p} \tabularnewline
\bottomrule
\end{tabular}
\drawcurvedarrow[bend right=60,-stealth]{y}{x}
\drawcurvedarrow[bend right=60,-stealth]{p}{o}
\drawcurvedarrow[bend right=60,-stealth]{h}{g}
\drawcurvedarrow[bend right=60,-stealth]{s}{r}
\drawcurvedarrow[bend right=80,-stealth]{b}{a}
\drawcurvedarrow[bend right=60,-stealth]{b}{c}
\vspace{-4px}
\caption{\small \label{tab:comparison_pooling_methods} \textbf{Comparison of pooling methods.} We compare Vertex Clustering (VC), Farthest Point Sampling (FPS), and Quadric Error Metrics (QEM) as pooling methods.}
\vspace{-6px}
\end{table}

\subsection{Ablation study.}
\label{sec:ablation_study}
We conduct an ablation study to support our claims that \numcircledtikz{1} a mesh-centric pooling method in the shape of Quadric Error Metrics, and \numcircledtikz{2} the combination of geodesic and Euclidean graph convolutions,
and \numcircledtikz{3} random edge sampling for effectively sampling the neighborhood space
independently contribute to an overall improved performance.
\vfill
In our study, we compare vertex clustering (VC), Quadric Error Metrics (QEM) and Farthest Point Sampling (FPS) as means of pooling in our \nameshort{} architecture. We conduct experiments with the \dualarch{} and \singlearch{} instantiations of our architecture with different notions of neighborhoods for the Euclidean ($k$-nn and radius (rad)) as well as the geodesic domain (geo).
For each EdgeConv in the \singlearch{} architecture, we set the hidden feature size to $128$ and the output size to $64$. To enable a fair comparison, we halve the hidden and output feature size of \dualarch{} architectures, such that the total number of feature channels is equal between the two versions. Note that this results in more than $15$\,$\%$ less parameters for \dualarch{} architectures while performing better.

\PAR{Varying the expected sample size during test.}
In Equation\,\ref{eq:red}, we introduce RES for reducing the expected size of the neighborhood set and therefore reducing the computational load.
In \reffig{varying_threshold}, we show the relationship between training a network with a relatively small sampled neighborhood and evaluating the algorithm with other set sizes.
We experience that a small neighborhood size, \eg, $T$\,=\,$15$, during training is still sufficient to learn useful features.
During test, we obtain better approximations of the neighborhood with larger thresholds, \eg, $T$\,=\,$35$, and report significantly better segmentation performances of +$1$\,$\%$ mIoU.
By decoupling the neighborhood size of the train and test times, we can adapt the expected size of the neighborhood to the computational resources given in the respective setting.
\vspace{-2pt}
\PAR{Comparison of pooling methods.}
In Section\,\ref{sec:method}, we motivate to adapt the mesh simplification algorithms Vertex Clustering and Quadric Error Metrics as means of pooling using pooling trace maps.
In \reftab{comparison_pooling_methods}, we evaluate the influence of different pooling methods in our architecture.
As an additional experiment, we perform pooling using Farthest Point Sampling\,\cite{Qi17NIPS} on the underlying point cloud since it neglects the mesh structure. Therefore, we can only perform Euclidean graph convolutions in this setting.
QEM performs significantly better than other pooling methods when using radius neighborhoods or the \dualarch{}, while being on par with VC when considering $k$-nn or geodesic neighborhoods for \singlearch{}s. We assume that the interplay between radius neighborhoods and QEM leads to this result. In contrast to VC, QEM does not aim for uniform vertex density. $k$-nn neighborhoods are sensible to varying vertex densities, as their spatial size is not limited\,\cite{Hermosilla2018Graph}.
\vspace{-2pt}
\PAR{Comparison of geodesic and Euclidean convolutions.}
\begin{table}[t]
\centering
\begin{tabular}{ccclr}
\toprule 
pool & arch & neighb & mIoU {\footnotesize($\pm$\,stdev)} & $\Delta$\tabularnewline
\midrule
VC & \singlearchshort{} & geo     & $57.1$ & \tikzmark[xshift=2.2em]{x}$+2.6$ \tabularnewline
VC & \singlearchshort{} & knn     & $60.1$ & \tikzmark[xshift=2.2em]{y}$-0.4$ \tabularnewline
VC & \dualarchshort{} & knn/geo & $59.7$ & \tikzmark[xshift=0.0em]{w}\tabularnewline
\midrule
QEM & \singlearchshort{} & geo     & $56.8$ & \tikzmark[xshift=2.2em]{z}$+6.1$ \tabularnewline
QEM & \singlearchshort{} & knn    & $60.9$ & \tikzmark[xshift=2.2em]{r}$+2.0$ \tabularnewline
QEM & \dualarchshort{} & knn/geo & $62.9$ &  \tikzmark[xshift=0.0em]{v}\tabularnewline
 \midrule\midrule
VC & \singlearchshort{} & geo     & $57.1$ & \tikzmark[xshift=2.2em]{a}$+5.7$ \tabularnewline
VC & \singlearchshort{} & rad     & $61.9$ {\footnotesize($\pm$\,$0.20$)} & \tikzmark[xshift=2.2em]{c}$+0.9$ \tabularnewline
VC & \dualarchshort{} & rad/geo & $62.8$ {\footnotesize($\pm$\,$0.12$)} & \tikzmark[xshift=0.0em]{b}\tabularnewline
\midrule
QEM & \singlearchshort{} & geo     & $56.8$ & \tikzmark[xshift=2.6em]{d}$+10.5$ \tabularnewline
QEM & \singlearchshort{} & rad     & $63.9$ {\footnotesize($\pm$\,$0.20$)} & \tikzmark[xshift=2.2em]{e}$+3.4$ \tabularnewline
QEM & \dualarchshort{} & rad/geo & $67.3$ {\footnotesize($\pm$\,$0.22$)} &  \tikzmark[xshift=0em]{f}\tabularnewline
\bottomrule
\end{tabular}
\drawcurvedarrow[bend right=80,-stealth]{w}{x}
\drawcurvedarrow[bend right=60,-stealth]{w}{y}
\drawcurvedarrow[bend right=80,-stealth]{v}{z}
\drawcurvedarrow[bend right=60,-stealth]{v}{r}
\drawcurvedarrow[bend right=80,-stealth]{b}{a}
\drawcurvedarrow[bend right=60,-stealth]{b}{c}
\drawcurvedarrow[bend right=80,-stealth]{f}{d}
\drawcurvedarrow[bend right=60,-stealth]{f}{e}
\vspace{-10px}
\caption{\small \label{tab:combination_geo_euc} \textbf{Combining geodesic and Euclidean convolutions} in our \nameshort{} brings significant performance improvements, especially compared to solely geodesic convolutions.}
\vspace{-6pt}
\end{table}
In Section\,\ref{sec:method}, we prompt the question whether a combination of geodesic and Euclidean convolutions leads to performance gains.
In \reftab{combination_geo_euc}, we compare models using only geodesic convolutions, only Euclidean convolutions, and both combined in dual convolution modules, while keeping the pooling method fixed.
We experience a clear trend that geodesic \singlearch{} architectures fall behind their Euclidean counterparts, whereas the effect for radius neigbhorhoods is stronger than for $k$-nn ones.
While the \dualarch{} combining VC with $k$-nn falls behind its \singlearch{} counterpart, the combination of geodesic and Euclidean neighborhoods in a \dualarch{} architecture in all other settings outperforms the corresponding \singlearch{} architectures.
To evaluate the confounding factor of storage limitation for \singlearch{}s with radius neighborhoods and QEM pooling, we test our model on a Titan RTX with $24$\,GB. We experience a performance of $65.9$\,$\%$, \ie, +$2$\,$\%$ to the model trained on a $16$\,GB V100.
To additionally prove that these performance gains do not just originate from the change to a \dualarch{} architecture, we conduct further experiments in \reftab{influence_of_architecture}.
Introducing our \dualarch{} architecture leads to worse results in direct comparison with the \singlearch{} architecture when using the same notion of neighborhood twice.
We thus conclude that the improvements brought by the combination of neighborhoods is based on the design decision of combining geodesic and Euclidean neighborhoods and is not just due to architectural artifacts.
\begin{table}[t!]
\centering
\begin{tabular}{ccclr}
\toprule
pool & arch & neighb & mIoU {\footnotesize($\pm$\,stdev)} & $\Delta$ \tabularnewline
\midrule
QEM & \dualarchshort{} & geo/geo & $56.3$ & \tikzmark[xshift=2.6em]{a}$+11.0$ \tabularnewline
QEM & \dualarchshort{} & rad/rad & $62.6$ {\footnotesize($\pm$\,$0.21$)} &  \tikzmark[xshift=2.2em]{c}$+4.7$ \tabularnewline
QEM & \dualarchshort{} & rad/geo & $67.3$ {\footnotesize($\pm$\,$0.22$)} &  \tikzmark[xshift=0em]{b}\tabularnewline
\midrule\midrule
QEM & \singlearchshort{} & rad     & $63.9$ {\footnotesize($\pm$\,$0.20$)} &  \tikzmark[xshift=2.2em]{r}$-1.3$ \tabularnewline
QEM & \dualarchshort{} & rad/rad & $62.6$ {\footnotesize($\pm$\,$0.26$)} &  \tikzmark[xshift=0em]{s} \tabularnewline
\midrule
QEM & \singlearchshort{} & geo     & $56.8$ &  \tikzmark[xshift=2.2em]{o}$-0.5$ \tabularnewline
QEM & \dualarchshort{} & geo/geo & $56.3$ &  \tikzmark[xshift=0em]{p}\tabularnewline
\bottomrule
\end{tabular}
\drawcurvedarrow[bend right=80,-stealth]{b}{a}
\drawcurvedarrow[bend right=60,-stealth]{b}{c}
\drawcurvedarrow[bend right=60,-stealth]{s}{r}
\drawcurvedarrow[bend right=60,-stealth]{p}{o}
\vspace{-8px}
\caption{
\small \label{tab:influence_of_architecture}
\textbf{Architectural influence.}
For the \dualarch{}, we see improvements when using geodesic and Euclidean neighborhoods in parallel, in contrast to only using the same neighborhood notion.}
\vspace{-4px}
\end{table}
\vspace{-8pt}
\section{Conclusion}
\vspace{-5pt}
In this paper, we have motivated a mesh-centric view on 3D scene segmentation and we have proposed \nameshort{}s to take advantage of the geometric surface information available in meshes. 
We hope that our work encourages fellow researchers to perform convolutions in both the geodesic and Euclidean domain, as we have empirically shown that this combination brings significant improvements independent to the architecture used.
Future work might include incorporating geodesic convolutions for better separating instances in the task of 3D instance segmentation, as well as extending our work for leveraging point convolutions.
\vspace{-14px}
\small{\paragraph{Acknowledgements.}
\footnotesize{
We thank Ali Athar, Markus Knoche, Tobias Fischer and Mark Weber for helpful discussions.
This work was supported by the ERC Consolidator Grant DeeViSe(ERC-2017-COG-773161).
The experiments were performed with computing resources granted by RWTH Aachen University under project rwth0470 and thes0617.}}

{
\small
\bibliographystyle{ieee_fullname}
\bibliography{egbib}
}

\clearpage

\twocolumn[{%
\renewcommand\twocolumn[1][]{#1}
\vspace{0.1cm}
\begin{center}
\textbf{\Large DualConvMesh-Net:\\[0.2cm]
Joint Geodesic and Euclidean Convolutions on 3D Meshes\\[0.2cm]
\textit{Supplementary Material}}
\end{center}
\vspace{1.7cm}
}]

\renewcommand{\thesection}{\Alph{section}}
\setcounter{section}{0}

\normalsize{
\section{Architectural design choices}
In this section, we give more details about our architectural design choices.
\numcircledtikz{1} By altering the filter ratio between geodesic and Euclidean convolutions for each mesh level, we further motivate the assumptions about the characteristics of Euclidean and geodesic convolutions and back them up with empirical evidence.
\numcircledtikz{2} We show the impact of the number of mesh levels for the \dualarch{} architecture.
\numcircledtikz{3}~We compare activation functions in our architecture. 
\vspace{8px}
\parag{Ratio between geodesic and Euclidean filters.}
\begin{table}[t]
\centering
\setlength{\tabcolsep}{1pt}
\begin{tabular}{cccr}
\toprule 
\multicolumn{2}{c}{\definecolor{geo_blue}{RGB}{40, 169, 250}
\definecolor{euc_red}{RGB}{255, 75, 51}

\tikzset{
horizontal fill/.style 2 args={fill=#2, path picture={
\fill[#1, sharp corners] (path picture bounding box.south west) -|
(path picture bounding box.north west) -|
(path picture bounding box.north) -|
                         (path picture bounding box.south) -- cycle;}}
}

\adjustbox{valign=c}{
\begin{tikzpicture}[shorten >=0pt,auto,node distance=0.0cm,thin, transform shape, scale=1.0]
\tikzstyle{every state}=[           rectangle,
           rounded corners,
           draw=black, thin,
           minimum height=0em,
           inner sep=0pt,
           text centered]

  \node (geo) [align=center, inner sep=2pt] at (-1,0) {Geodesic};
  \node (euc) [align=center, inner sep=2pt] at (0.65,0) {Euclidean};
  
  \node (bb) [state, minimum height=0cm, horizontal fill={geo_blue}{euc_red}, inner sep=0pt, fit={(geo) (euc)}] {};
  
\node (geo) [align=center] at (-1,0) {Geodesic};
\node (euc) [align=center] at (0.65,0) {Euclidean};

  \draw ($(bb.north) + (0.0, 0)$) --
($(bb.south) + (0.0, 0)$);

\node [align=center] at (2.0,0) {Ratio};
\end{tikzpicture}
}} \\
 \cmidrule(r){1-2}
level 1-2 & level 3-4 & mIoU {\footnotesize($\pm$\,stdev)} & $\Delta$\tabularnewline
\midrule
\definecolor{geo_blue}{RGB}{40, 169, 250}
\definecolor{euc_red}{RGB}{255, 75, 51}

\tikzset{
horizontal fill/.style 2 args={fill=#2, path picture={
\fill[#1, sharp corners] (path picture bounding box.south west) -|
(path picture bounding box.north west) --
($(path picture bounding box.north) + (-0.3, 0)$) --
($(path picture bounding box.south) + (-0.3, 0)$) -- cycle;}}
}
\adjustbox{valign=c}{
\centering
\begin{tikzpicture}[shorten >=0pt,auto,node distance=0.0cm,thin, transform shape, scale=1.0]
\tikzstyle{every state}=[           rectangle,
           rounded corners,
           draw=black, thin,
           minimum height=0em,
           inner sep=0pt,
           text centered]

  \node (geo) [align=center, inner sep=2pt] at (-0.85,0) {\small $25\%$};
  \node (euc) [align=center, inner sep=2pt] at (0.45,0) {\small $75\%$};
  
  \node (bb) [state, minimum height=0cm, horizontal fill={geo_blue}{euc_red}, inner sep=0pt, fit={(geo) (euc)}] {};
  
      \draw ($(bb.north) + (-0.3, 0)$) --
($(bb.south) + (-0.3, 0)$);
  
  \node (geo) [align=center] at (-0.85,0) {\small $25\%$};
  \node (euc) [align=center] at (0.45,0) {\small $75\%$};
\end{tikzpicture}
} & \definecolor{geo_blue}{RGB}{40, 169, 250}
\definecolor{euc_red}{RGB}{255, 75, 51}

\tikzset{
horizontal fill/.style 2 args={fill=#2, path picture={
\fill[#1, sharp corners] (path picture bounding box.south west) -|
(path picture bounding box.north west) --
($(path picture bounding box.north) + (0.3, 0)$) --
($(path picture bounding box.south) + (0.3, 0)$) -- cycle;}}
}

\adjustbox{valign=c}{
\begin{tikzpicture}[shorten >=0pt,auto,node distance=0.0cm,thin, transform shape, scale=1.0]
\tikzstyle{every state}=[           rectangle,
           rounded corners,
           draw=black, thin,
           minimum height=0em,
           inner sep=0pt,
           text centered]

  \node (geo) [align=center, inner sep=2pt] at (-0.85,0) {\small $75\%$};
  \node (euc) [align=center, inner sep=2pt] at (0.45,0) {\small $25\%$};
  
  \node (bb) [state, minimum height=0cm, horizontal fill={geo_blue}{euc_red}, inner sep=0pt, fit={(geo) (euc)}] {};
  
  \draw ($(bb.north) + (0.3, 0)$) --
($(bb.south) + (0.3, 0)$);
  
  \node (geo) [align=center] at (-0.85,0) {\small $75\%$};
  \node (euc) [align=center] at (0.45,0) {\small $25\%$};
\end{tikzpicture}
} & $66.0$ {\footnotesize($\pm$\,$0.14$)} & \tikzmark[xshift=2.2em]{b}$+2.3$\tabularnewline
\definecolor{geo_blue}{RGB}{40, 169, 250}
\definecolor{euc_red}{RGB}{255, 75, 51}

\tikzset{
horizontal fill/.style 2 args={fill=#2, path picture={
\fill[#1, sharp corners] (path picture bounding box.south west) -|
(path picture bounding box.north west) --
($(path picture bounding box.north) + (0.3, 0)$) --
($(path picture bounding box.south) + (0.3, 0)$) -- cycle;}}
}

\adjustbox{valign=c}{
\begin{tikzpicture}[shorten >=0pt,auto,node distance=0.0cm,thin, transform shape, scale=1.0]
\tikzstyle{every state}=[           rectangle,
           rounded corners,
           draw=black, thin,
           minimum height=0em,
           inner sep=0pt,
           text centered]

  \node (geo) [align=center, inner sep=2pt] at (-0.85,0) {\small $75\%$};
  \node (euc) [align=center, inner sep=2pt] at (0.45,0) {\small $25\%$};
  
  \node (bb) [state, minimum height=0cm, horizontal fill={geo_blue}{euc_red}, inner sep=0pt, fit={(geo) (euc)}] {};
  
  \draw ($(bb.north) + (0.3, 0)$) --
($(bb.south) + (0.3, 0)$);
  
  \node (geo) [align=center] at (-0.85,0) {\small $75\%$};
  \node (euc) [align=center] at (0.45,0) {\small $25\%$};
\end{tikzpicture}
} & \definecolor{geo_blue}{RGB}{40, 169, 250}
\definecolor{euc_red}{RGB}{255, 75, 51}

\tikzset{
horizontal fill/.style 2 args={fill=#2, path picture={
\fill[#1, sharp corners] (path picture bounding box.south west) -|
(path picture bounding box.north west) --
($(path picture bounding box.north) + (0.3, 0)$) --
($(path picture bounding box.south) + (0.3, 0)$) -- cycle;}}
}

\adjustbox{valign=c}{
\begin{tikzpicture}[shorten >=0pt,auto,node distance=0.0cm,thin, transform shape, scale=1.0]
\tikzstyle{every state}=[           rectangle,
           rounded corners,
           draw=black, thin,
           minimum height=0em,
           inner sep=0pt,
           text centered]

  \node (geo) [align=center, inner sep=2pt] at (-0.85,0) {\small $75\%$};
  \node (euc) [align=center, inner sep=2pt] at (0.45,0) {\small $25\%$};
  
  \node (bb) [state, minimum height=0cm, horizontal fill={geo_blue}{euc_red}, inner sep=0pt, fit={(geo) (euc)}] {};
  
  \draw ($(bb.north) + (0.3, 0)$) --
($(bb.south) + (0.3, 0)$);
  
  \node (geo) [align=center] at (-0.85,0) {\small $75\%$};
  \node (euc) [align=center] at (0.45,0) {\small $25\%$};
\end{tikzpicture}
} & $66.1$ {\footnotesize($\pm$\,$0.19$)} & \tikzmark[xshift=2.2em]{a}$+2.2$\tabularnewline
\definecolor{geo_blue}{RGB}{40, 169, 250}
\definecolor{euc_red}{RGB}{255, 75, 51}

\tikzset{
horizontal fill/.style 2 args={fill=#2, path picture={
\fill[#1, sharp corners] (path picture bounding box.south west) -|
(path picture bounding box.north west) --
($(path picture bounding box.north) + (-0.3, 0)$) --
($(path picture bounding box.south) + (-0.3, 0)$) -- cycle;}}
}
\adjustbox{valign=c}{
\centering
\begin{tikzpicture}[shorten >=0pt,auto,node distance=0.0cm,thin, transform shape, scale=1.0]
\tikzstyle{every state}=[           rectangle,
           rounded corners,
           draw=black, thin,
           minimum height=0em,
           inner sep=0pt,
           text centered]

  \node (geo) [align=center, inner sep=2pt] at (-0.85,0) {\small $25\%$};
  \node (euc) [align=center, inner sep=2pt] at (0.45,0) {\small $75\%$};
  
  \node (bb) [state, minimum height=0cm, horizontal fill={geo_blue}{euc_red}, inner sep=0pt, fit={(geo) (euc)}] {};
  
      \draw ($(bb.north) + (-0.3, 0)$) --
($(bb.south) + (-0.3, 0)$);
  
  \node (geo) [align=center] at (-0.85,0) {\small $25\%$};
  \node (euc) [align=center] at (0.45,0) {\small $75\%$};
\end{tikzpicture}
} & \definecolor{geo_blue}{RGB}{40, 169, 250}
\definecolor{euc_red}{RGB}{255, 75, 51}

\tikzset{
horizontal fill/.style 2 args={fill=#2, path picture={
\fill[#1, sharp corners] (path picture bounding box.south west) -|
(path picture bounding box.north west) --
($(path picture bounding box.north) + (-0.3, 0)$) --
($(path picture bounding box.south) + (-0.3, 0)$) -- cycle;}}
}
\adjustbox{valign=c}{
\centering
\begin{tikzpicture}[shorten >=0pt,auto,node distance=0.0cm,thin, transform shape, scale=1.0]
\tikzstyle{every state}=[           rectangle,
           rounded corners,
           draw=black, thin,
           minimum height=0em,
           inner sep=0pt,
           text centered]

  \node (geo) [align=center, inner sep=2pt] at (-0.85,0) {\small $25\%$};
  \node (euc) [align=center, inner sep=2pt] at (0.45,0) {\small $75\%$};
  
  \node (bb) [state, minimum height=0cm, horizontal fill={geo_blue}{euc_red}, inner sep=0pt, fit={(geo) (euc)}] {};
  
      \draw ($(bb.north) + (-0.3, 0)$) --
($(bb.south) + (-0.3, 0)$);
  
  \node (geo) [align=center] at (-0.85,0) {\small $25\%$};
  \node (euc) [align=center] at (0.45,0) {\small $75\%$};
\end{tikzpicture}
} & $66.9$ {\footnotesize($\pm$\,$0.20$)} & \tikzmark[xshift=2.2em]{c}$+1.4$\tabularnewline
\definecolor{geo_blue}{RGB}{40, 169, 250}
\definecolor{euc_red}{RGB}{255, 75, 51}

\tikzset{
horizontal fill/.style 2 args={fill=#2, path picture={
\fill[#1, sharp corners] (path picture bounding box.south west) -|
(path picture bounding box.north west) --
($(path picture bounding box.north) + (-0, 0)$) --
($(path picture bounding box.south) + (-0, 0)$) -- cycle;}}
}

\adjustbox{valign=c}{
\begin{tikzpicture}[shorten >=0pt,auto,node distance=0.0cm,thin, transform shape, scale=1.0]
\tikzstyle{every state}=[           rectangle,
           rounded corners,
           draw=black, thin,
           minimum height=0em,
           inner sep=0pt,
           text centered]

  \node (geo) [align=center, inner sep=2pt] at (-0.85,0) {\small $50\%$};
  \node (euc) [align=center, inner sep=2pt] at (0.45,0) {\small $50\%$};
  
  \node (bb) [state, minimum height=0cm, horizontal fill={geo_blue}{euc_red}, inner sep=0pt, fit={(geo) (euc)}] {};
  
    \draw ($(bb.north) + (0.0, 0)$) --
($(bb.south) + (0.0, 0)$);
  
  \node (geo) [align=center] at (-0.85,0) {\small $50\%$};
  \node (euc) [align=center] at (0.45,0) {\small $50\%$};
\end{tikzpicture}
} & \definecolor{geo_blue}{RGB}{40, 169, 250}
\definecolor{euc_red}{RGB}{255, 75, 51}

\tikzset{
horizontal fill/.style 2 args={fill=#2, path picture={
\fill[#1, sharp corners] (path picture bounding box.south west) -|
(path picture bounding box.north west) --
($(path picture bounding box.north) + (-0, 0)$) --
($(path picture bounding box.south) + (-0, 0)$) -- cycle;}}
}

\adjustbox{valign=c}{
\begin{tikzpicture}[shorten >=0pt,auto,node distance=0.0cm,thin, transform shape, scale=1.0]
\tikzstyle{every state}=[           rectangle,
           rounded corners,
           draw=black, thin,
           minimum height=0em,
           inner sep=0pt,
           text centered]

  \node (geo) [align=center, inner sep=2pt] at (-0.85,0) {\small $50\%$};
  \node (euc) [align=center, inner sep=2pt] at (0.45,0) {\small $50\%$};
  
  \node (bb) [state, minimum height=0cm, horizontal fill={geo_blue}{euc_red}, inner sep=0pt, fit={(geo) (euc)}] {};
  
    \draw ($(bb.north) + (0.0, 0)$) --
($(bb.south) + (0.0, 0)$);
  
  \node (geo) [align=center] at (-0.85,0) {\small $50\%$};
  \node (euc) [align=center] at (0.45,0) {\small $50\%$};
\end{tikzpicture}
} & $67.5$ {\footnotesize($\pm$\,$0.13$)} & \tikzmark[xshift=2.2em]{d}$+0.8$\tabularnewline
\definecolor{geo_blue}{RGB}{40, 169, 250}
\definecolor{euc_red}{RGB}{255, 75, 51}

\tikzset{
horizontal fill/.style 2 args={fill=#2, path picture={
\fill[#1, sharp corners] (path picture bounding box.south west) -|
(path picture bounding box.north west) --
($(path picture bounding box.north) + (0.3, 0)$) --
($(path picture bounding box.south) + (0.3, 0)$) -- cycle;}}
}

\adjustbox{valign=c}{
\begin{tikzpicture}[shorten >=0pt,auto,node distance=0.0cm,thin, transform shape, scale=1.0]
\tikzstyle{every state}=[           rectangle,
           rounded corners,
           draw=black, thin,
           minimum height=0em,
           inner sep=0pt,
           text centered]

  \node (geo) [align=center, inner sep=2pt] at (-0.85,0) {\small $75\%$};
  \node (euc) [align=center, inner sep=2pt] at (0.45,0) {\small $25\%$};
  
  \node (bb) [state, minimum height=0cm, horizontal fill={geo_blue}{euc_red}, inner sep=0pt, fit={(geo) (euc)}] {};
  
  \draw ($(bb.north) + (0.3, 0)$) --
($(bb.south) + (0.3, 0)$);
  
  \node (geo) [align=center] at (-0.85,0) {\small $75\%$};
  \node (euc) [align=center] at (0.45,0) {\small $25\%$};
\end{tikzpicture}
} & \definecolor{geo_blue}{RGB}{40, 169, 250}
\definecolor{euc_red}{RGB}{255, 75, 51}

\tikzset{
horizontal fill/.style 2 args={fill=#2, path picture={
\fill[#1, sharp corners] (path picture bounding box.south west) -|
(path picture bounding box.north west) --
($(path picture bounding box.north) + (-0.3, 0)$) --
($(path picture bounding box.south) + (-0.3, 0)$) -- cycle;}}
}
\adjustbox{valign=c}{
\centering
\begin{tikzpicture}[shorten >=0pt,auto,node distance=0.0cm,thin, transform shape, scale=1.0]
\tikzstyle{every state}=[           rectangle,
           rounded corners,
           draw=black, thin,
           minimum height=0em,
           inner sep=0pt,
           text centered]

  \node (geo) [align=center, inner sep=2pt] at (-0.85,0) {\small $25\%$};
  \node (euc) [align=center, inner sep=2pt] at (0.45,0) {\small $75\%$};
  
  \node (bb) [state, minimum height=0cm, horizontal fill={geo_blue}{euc_red}, inner sep=0pt, fit={(geo) (euc)}] {};
  
      \draw ($(bb.north) + (-0.3, 0)$) --
($(bb.south) + (-0.3, 0)$);
  
  \node (geo) [align=center] at (-0.85,0) {\small $25\%$};
  \node (euc) [align=center] at (0.45,0) {\small $75\%$};
\end{tikzpicture}
} & $\mathbf{68.3}${\footnotesize($\pm$\,$0.12$)}&\tikzmark[xshift=0.0em]{e}\tabularnewline
\bottomrule
\end{tabular}
\vspace{-8px}
\caption{\small \label{tab:scannet_different_ratios} \textbf{Geodesic/Euclidean filter ratio per mesh level.}
Geodesic convolutions are particularly useful in early mesh levels, when high frequency signals of the mesh are still preserved.
In later levels, we benefit from Euclidean convolutions for localizing objects better.
To this end, we use a larger ratio of geodesic filters in early levels, whereas we use more Euclidean ones in later levels.
(Level 1-2 use $64$ and level 3-4 use $96$ filters in total.)}
\end{table}
Following the intuition that geodesic convolutions mainly benefit from high-frequency mesh information in order to learn the inherent shape of objects, we want to learn more geodesic than Euclidean features in high resolution mesh levels. Contrastingly, Euclidean features are beneficial for localizing objects in the scene which is better performed in lower resolutions. In order to verify this intuition, we present the results of an experiment in which we systemically modified the ratio of geodesic and Euclidean filters per mesh level.

In~\reftab{scannet_different_ratios}, more geodesic filters in the first two levels and more Euclidean filters in later two levels bring significant performance gains over other ratio settings.
We see this as a clear indicator that our assumption about the inherent properties about Euclidean and geodesic convolutions hold.
\parag{Number of mesh levels.}
\begin{table}[t]
\centering
\begin{tabular}{ccr}
\toprule 
\#level & mIoU {\footnotesize($\pm$\,stdev)} & $\Delta$\tabularnewline
\midrule
$2$ & $54.4$ {\footnotesize($\pm$\,$0.07$)} & \tikzmark[xshift=2.8em]{a}$+12.9$ \tabularnewline
$3$ & $64.0$ {\footnotesize($\pm$\,$0.14$)} & \tikzmark[xshift=2.2em]{b}$+3.3$ \tabularnewline
$4$ & $\mathbf{67.3}${\footnotesize($\pm$\,$0.22$)} & \tikzmark[xshift=0.0em]{c} \tabularnewline
\bottomrule
\end{tabular}
\vspace{-8px}
\caption{\small \label{tab:influence_of_graph_levels} \textbf{Influence of the number of mesh levels.} We observe that the multi-scale architecture has a strong impact on the performance. With decreasing effect, more mesh levels bring performance gains.
(Experiments where conducted with QEM pooling and geodesic/radius neighborhoods in our \dualarch{}.)}
\end{table}
In \reftab{influence_of_graph_levels}, we experimentally show the importance of multi-scale hierarchies for semantic segmentation for meshed point clouds. We see a clear trend that an increased number of mesh levels with different resolutions bring a significant performance gain.
\parag{Activation functions.}
\begin{table}[t]
\centering
\begin{tabular}{rcr}
\toprule 
activation function & mIoU {\footnotesize($\pm$\,stdev)} & $\Delta$\tabularnewline
\midrule
Leaky ReLU & $65.7$ {\footnotesize($\pm$\,$0.14$)} & \tikzmark[xshift=2.2em]{a}$+1.4$ \tabularnewline
ReLU       & $\mathbf{67.3}${\footnotesize($\pm$\,$0.22$)} & \tikzmark[xshift=0.0em]{b} \tabularnewline
\bottomrule
\end{tabular}
\vspace{-8px}
\caption{\small \label{tab:comparison_activation_functions} \textbf{Comparison of activation functions.} As Leaky ReLU gains popularity, we compare it with standard ReLU activation functions. We conclude that default ReLU units work significantly better for our architecture. (Experiments are conducted with QEM pooling and geodesic/radius neighborhoods in a \dualarch{}.)}
\end{table}
Recent publications on 3D scene segmentation rely on Leaky ReLU activation functions~\cite{thomas2019ICCV}. In~\reftab{comparison_activation_functions}, we compare standard ReLU with LeakyReLU activation functions. We conclude that for our architecture LeakyReLU activations do not bring any benefits and decrease the performance by $1.6\%$\,mIoU.
\vspace{-8px}
\section{Detailed network descriptions}
\vspace{-4px}
\input{supplementary/tikz_figures/runtime.tex}
In the ablation study of the main paper, we focus particularly on the comparability of our proposed networks.
We compare basic instantiations of \singlearch{}s with its \dualarch{} equivalents in the ablation study (see Table~\ref{tab:architectures_ablation_study}).
Note that we ensure the same size of hidden and output channels for each edge convolution and dual convolution. That is, the $128$ hidden and $64$ output channels of single edge convolutions of the \singlearch{}s are halved resulting in $64$ hidden and $32$ output features for geodesic and Euclidean filters of the dual convolutions. Thus, \dualarch{}s have $15\%$ less parameters than their \singlearch{} equivalents.

However, we use extended networks for obtaining final scores on the benchmarks. Motivated by Table~\ref{tab:scannet_different_ratios}, we additionally vary the ratio of geodesic and Euclidean filters and changed the number of features in each mesh level.
In the following paragraphs, we give detailed network descriptions for each benchmark.

\parag{Network architectures for ScanNet / Matterport3D.}
We use the \dualarch{} with $75\%$ geodesic out of $48$ features in the first two mesh levels and $25\%$ geodesic out of $96$ features in the last two mesh levels.
We use batch normalization and ReLU activations for the edge convolutions.
In Table~\ref{tab:architecture_2_ratio}, we show the detailed network architecture for the ScanNet and Matterport3D benchmark.
\parag{Special provisions for S3DIS.}
In contrast to ScanNet and Matterport3D, S3DIS is characterized by the comparably lower resolution of its underlying mesh structure.
In order to use the ground truth information of the official point clouds sampled from these meshes,
we artificially increase the resolution of the mesh by splitting edges exactly in the middle if the edge length does not fall under $2$\,cm. We create new triangles by connecting the old vertices with their adjacent vertices at the midst of the edges. Thus, we obtain $4$ smaller triangles from the original triangle. We subsequently interpolate the ground truth information on this newly created mesh. In Figure~\ref{fig:s3dis_preprocess}, we provide an illustration of the preprocessing pipeline for S3DIS.
\begin{figure*}[t]
\centering
\begin{tikzpicture}[-,>=stealth',shorten >=0pt,auto,node distance=1.6cm,semithick, transform shape, scale=1.0]
\tikzstyle{every state}=[           rectangle,
           rounded corners,
           draw=black, thick,
           minimum height=1em,
           inner sep=2mm, align=center,
           text centered]
  \def\filterSS{\node{};  
     \draw[line width=1pt] (-2mm,-4mm) to[in=200,out=20] (-2mm, 4mm) 
                           (0mm,-4mm) to[in=200,out=20] (0mm, 4mm) 
                           (2mm,-4mm) to[in=200,out=20] (2mm, 4mm); 
     }

    \tikzset{trianglenode/.style={fill=blue!20, regular polygon, regular polygon sides=3, rotate=90, shape border rotate=90}};
    
    \tikzset{circlenode/.style={circle,
           draw=black, thick,
           minimum height=0em,
           inner sep=0mm, align=center,fill,circle,minimum size=0.1cm,
           text centered}};
     
    \tikzset{blue dotted/.style={draw=blue!50!white, line width=1pt,
                               dotted, inner sep=4mm, rectangle, rounded corners}};
                               
    \tikzset{orange dotted/.style={draw=orange!50!white, line width=1pt,
                               dotted, inner sep=4mm, rectangle, rounded corners}};
                               
    \tikzset{green dotted/.style={draw=green!50!white, line width=1pt,
                               dotted, inner sep=4mm, rectangle, rounded corners}};
                               
        \tikzset{skip/.style={draw=black, line width=0.5pt,
                              inner sep=1mm, rectangle, rounded corners}};

  \def\MOD#1{\node{#1}; 
    \draw[line width=1pt,sharp corners](-0.75cm,0cm)--(-0.35cm,0.25cm)--
         (0.35cm, 0.25cm)--(0.75cm, 0cm)--(0.35cm, -0.25cm)--(-0.35cm, -0.25cm) -- cycle; 
    }
    
           \node[circlenode] (A) at (4.0,0.0) {};  
    \node[circlenode] (B) at (3.0,0.0) {}; 
    \node[circlenode] (C) at (3.5,1.0) {}; 
    
   \fill[fill=gray!20] (A.center)--(B.center)--(C.center);
   
       \node[circlenode] (A2) at (5.5,0.0) {};  
    \node[circlenode] (B2) at (4.5,0.0) {}; 
    \node[circlenode] (C2) at (5,1.0) {}; 
        \node[circlenode] (D2) at (5,0.0) {}; 
    \node[circlenode] (E2) at (4.75,0.5) {}; 
        \node[circlenode] (F2) at (5.25,0.5) {}; 
        
       \fill[fill=purple!20] (A2.center)--(D2.center)--(F2.center);
  \fill[fill=green!20] (B2.center)--(D2.center)--(E2.center);
  \fill[fill=blue!20] (D2.center)--(F2.center)--(E2.center);
  \fill[fill=orange!20] (E2.center)--(C2.center)--(F2.center);
  
      \node[circlenode] (A) at (4.0,0.0) {};  
    \node[circlenode] (B) at (3.0,0.0) {}; 
    \node[circlenode] (C) at (3.5,1.0) {}; 
           \node[circlenode] (A2) at (5.5,0.0) {};  
    \node[circlenode] (B2) at (4.5,0.0) {}; 
    \node[circlenode] (C2) at (5,1.0) {}; 
        \node[circlenode] (D2) at (5,0.0) {}; 
    \node[circlenode] (E2) at (4.75,0.5) {}; 
        \node[circlenode] (F2) at (5.25,0.5) {}; 
  
\draw [->] (4.0,0.5) -- (4.5,0.5);
       \node[state, inner sep=1mm, fit={(B) (C) ($(F2.east)+(7pt,30pt)$)}] (midpoint) [] {};
   \node[state] (low_res) at ($(midpoint) + (-100pt, 0pt)$) {Unlabeled\\Low-Res Mesh};
  
  \node[] (descr) at ($(midpoint) + (0pt, 20pt)$) {Subdivision};
   \node[state] (hi_res) [right=2em of midpoint] {Unlabeled\\Hi-Res Mesh};
      \node[state] (interpolate) [right=2em of hi_res] {Interpolation};
    \node[state] (labeled) [right=2em of interpolate] {Labeled\\Hi-Res Mesh};
   \node[state] (gt) [above=1em of interpolate] {Ground Truth\\Point Cloud
   };

    \path (low_res) edge[->, shorten >=2pt]          node[sloped, anchor=center, above] {} (midpoint);
    
    \path (gt) edge[->, shorten >=2pt]          node[sloped, anchor=center, above] {} (interpolate);
    
    \path (hi_res) edge[->, shorten >=2pt]          node[sloped, anchor=center, above] {} (interpolate);
    
    \path (interpolate) edge[->, shorten >=2pt]          node[sloped, anchor=center, above] {} (labeled);
    
    \path (midpoint) edge[->, shorten >=2pt]          node[sloped, anchor=center, above] {} (hi_res);

\end{tikzpicture}
  \caption{\textbf{Preprocessing Pipeline for S3DIS.}
  Our approach requires meshes as input for which the S3DIS data set does not provide an RGB + Label format. Therefore, we establish a preprocessing pipeline in order to leverage low-resolution meshes given by the dataset. Here, we perform midpoint subdivision to artificially enhance the resolution of the mesh, before interpolating RGB colors as well as labels from the ground truth point cloud onto the mesh.}
  \label{fig:s3dis_preprocess}
  \vspace{10px}
\end{figure*}
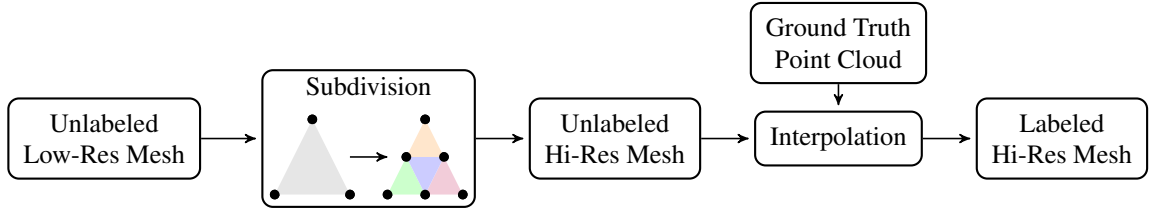

Since the original resolution of the mesh is low, we do not benefit from increasing the number of geodesic filters in the early levels, as we motivate for ScanNet in~\reftab{scannet_different_ratios}. Thus, we set the ratio of geodesic convolutions in each level to $50\%$, similarly to the ablation study in the main paper. In Table~\ref{tab:architecture_2_ratio_s3dis}, we provide the adapted network structure.
\vfill
\section{Runtime}
In Figure~\ref{fig:runtime}, we provide forward pass times for our ScanNet benchmark model with respect to the input size. We see a linear relationship between the number of input vertices and the runtime which is always well under $0.7$ seconds for all scans. Overall, the mean runtime for the ScanNet validation set is $211$ms with an average input size of $39161$ vertices.
We perform this experiment with a Tesla V100.
\section{Quantitative and qualitative results}
We provide additional segmentation results on Stanford Large-Scale 3D Indoor Spaces (S3DIS) to allow an in-depth comparison with competitive approaches. In Table~\ref{tab:Table_S3DIS_area5} and~\ref{tab:TableS3DISkfold}, we report class-wise segmentation results on S3DIS $k$-fold and Area~$5$.
We show further qualitative results on S3DIS~\cite{Armeni16CVPR} and Matterport3D~\cite{Matterport3D} in Figures \ref{fig:s3dis_quali} and \ref{fig:matterport_quali}.
\parag{Majority Voting.}
\begin{table}[t]
\centering
\begin{tabular}{rccc}
\toprule 
dataset & single run & majority & $\Delta$\tabularnewline
\midrule
ScanNet~\cite{Dai17CVPR} (\emph{test}) & $65.3$ & $65.8$ & $0.5$ \tabularnewline
S3DIS~\cite{Armeni16CVPR} (Area-$5$)   & $63.8$ & $64.0$ & $0.2$ \tabularnewline
S3DIS~\cite{Armeni16CVPR}~~($k$-fold)   & $69.4$ & $69.7$ & $0.3$ \tabularnewline
\midrule
Matterport3D~\cite{Matterport3D} & $65.5$ & $66.2$ & $0.7$ \tabularnewline
\bottomrule
\end{tabular}
\caption{\small \label{tab:majority_voting} \textbf{Majority voting.}
By using majority voting with $100$ runs on augmented scenes, we experience performance gains up to $0.5$\,$\%$ mIoU on ScanNet and S3DIS.
Our scores on Matterport3D increase by $0.7$\,$\%$ mAcc compared to the single run variant with no test time augmentations.}
\end{table}
To obtain the final scores for the benchmarks, we leverage \emph{majority voting} with $100$ runs of the best performing model on augmented test scenes. In Table~\ref{tab:majority_voting}, we compare single runs of the models on non-augmented scenes against the majority voting method explained before.

\begin{table*}[t!]
\begin{subtable}[t!]{0.48\textwidth}
\vspace{-35px}
\begin{tabular}{ccrr}
\toprule 
\#level & level type & module type& filters\tabularnewline
\midrule
$1$ & encoder & edge+BN+ReLU & $(9,128,64)$ \tabularnewline
$1$ & encoder & edge+BN+ReLU & $(128,128,64)$ \tabularnewline
$1$ & encoder & edge+BN+ReLU & $(128,128,64)$ \tabularnewline
\midrule
$2$ & encoder & edge+BN+ReLU & $(128,128,64)$ \tabularnewline
$2$ & encoder & edge+BN+ReLU & $(128,128,64)$ \tabularnewline
$2$ & encoder & edge+BN+ReLU & $(128,128,64)$ \tabularnewline
\midrule
$3$ & encoder & edge+BN+ReLU & $(128,128,64)$ \tabularnewline
$3$ & encoder & edge+BN+ReLU & $(128,128,64)$ \tabularnewline
$3$ & encoder & edge+BN+ReLU & $(128,128,64)$ \tabularnewline
\midrule
$4$ & encoder & edge+BN+ReLU & $(128,128,64)$ \tabularnewline
$4$ & encoder & edge+BN+ReLU & $(128,128,64)$ \tabularnewline
$4$ & encoder & edge+BN+ReLU & $(128,128,64)$ \tabularnewline
\specialrule{.2em}{.1em}{.1em} 
$3$ & decoder & edge+BN+ReLU & $(256,128,64)$ \tabularnewline
$3$ & decoder & edge+BN+ReLU & $(128,128,64)$ \tabularnewline
$3$ & decoder & edge+BN+ReLU & $(128,128,64)$ \tabularnewline
\midrule
$2$ & decoder & edge+BN+ReLU & $(256,128,64)$ \tabularnewline
$2$ & decoder & edge+BN+ReLU & $(128,128,64)$ \tabularnewline
$2$ & decoder & edge+BN+ReLU & $(128,128,64)$ \tabularnewline
\midrule
$1$ & decoder & edge+BN+ReLU & $(256,128,64)$ \tabularnewline
$1$ & decoder & edge+BN+ReLU & $(128,128,64)$ \tabularnewline
$1$ & decoder & edge+BN+ReLU & $(128,128,64)$ \tabularnewline
\midrule\midrule
$1$ & final & Lin+BN+ReLU & $(64, 32)$ \tabularnewline
$1$ & final & Lin & $(32, 21)$ \tabularnewline
\bottomrule
\tabularnewline
&&\# parameters: & $\mathbf{564,949}$
\end{tabular}
\caption{\small \label{tab:architecture_1_branch} \textbf{\singlearch{} architecture.} We use \singlearch{}s for the ablation study. Here, we only consider either geodesic or Euclidean neighborhood information and do not fuse this information.}
\end{subtable}
\hspace{\fill}
\begin{subtable}[t!]{0.48\textwidth}
\flushright
\vspace{-5px}
\begin{tabular}{ccrr}
\toprule 
\#level & level type & module type& filters\tabularnewline
\midrule
$1$ & encoder & edge+BN+ReLU & $2*(9,64,32)$ \tabularnewline
$1$ & encoder & edge+BN+ReLU & $2*(128,64,32)$ \tabularnewline
$1$ & encoder & edge+BN+ReLU & $2*(128,64,32)$\tabularnewline
\midrule
$2$ & encoder & edge+BN+ReLU & $2*(128,64,32)$ \tabularnewline
$2$ & encoder & edge+BN+ReLU & $2*(128,64,32)$ \tabularnewline
$2$ & encoder & edge+BN+ReLU & $2*(128,64,32)$ \tabularnewline
\midrule
$3$ & encoder & edge+BN+ReLU & $2*(128,64,32)$ \tabularnewline
$3$ & encoder & edge+BN+ReLU & $2*(128,64,32)$ \tabularnewline
$3$ & encoder & edge+BN+ReLU & $2*(128,64,32)$ \tabularnewline
\midrule
$4$ & encoder & edge+BN+ReLU & $2*(128,64,32)$ \tabularnewline
$4$ & encoder & edge+BN+ReLU & $2*(128,64,32)$ \tabularnewline
$4$ & encoder & edge+BN+ReLU & $2*(128,64,32)$ \tabularnewline
\specialrule{.2em}{.1em}{.1em} 
$3$ & decoder & edge+BN+ReLU & $2*(256,64,32)$ \tabularnewline
$3$ & decoder & edge+BN+ReLU & $2*(128,64,32)$ \tabularnewline
$3$ & decoder & edge+BN+ReLU & $2*(128,64,32)$ \tabularnewline
\midrule
$2$ & decoder & edge+BN+ReLU & $2*(256,64,32)$ \tabularnewline
$2$ & decoder & edge+BN+ReLU & $2*(128,64,32)$ \tabularnewline
$2$ & decoder & edge+BN+ReLU & $2*(128,64,32)$ \tabularnewline
\midrule
$1$ & decoder & edge+BN+ReLU & $2*(256,64,32)$ \tabularnewline
$1$ & decoder & edge+BN+ReLU & $2*(128,64,32)$ \tabularnewline
$1$ & decoder & edge+BN+ReLU & $2*(128,64,32)$ \tabularnewline
\midrule\midrule
$1$ & final & Lin+BN+ReLU & $(64, 32)$ \tabularnewline
$1$ & final & Lin & $(32, 21)$ \tabularnewline
\bottomrule
\tabularnewline
&&\# parameters: & $\mathbf{478,933}$
\end{tabular}
\caption{\small \label{tab:architecture_2_branch} \textbf{\dualarch{} architecture.}
We perform convolutions in the geodesic and Euclidean space simultaneously and subsequently concatenate the features. Note that the total size of hidden and output features for each dual convolution equals its \singlearch{} edge convolution equivalent. We are therefore able to perform fair comparisons between these two types. 
}
\end{subtable}
\caption{\textbf{Architectures for the ablation study.}
In our ablation study, we experimentally prove the effectiveness of combining geodesic and Euclidean convolutions. We propose \singlearch{}s for applying convolutions either in the geodesic or Euclidean space and \dualarch{}s which jointly perform convolutions in the geodesic and Euclidean space.}
\label{tab:architectures_ablation_study}
\end{table*}
\begin{table*}[t!]
\begin{subtable}[t!]{0.48\textwidth}
\begin{tabular}{ccrr}
\toprule 
&&\multicolumn{2}{c}{filters} \\
 \cmidrule(r){3-4}
\#level & level type & geodesic & Euclidean\tabularnewline
\midrule
$1$ & encoder &  $(9,96,48)$ & $(9, 32, 16)$ \tabularnewline
$1$ & encoder &  $(128,96,48)$ & $(128, 32, 16)$ \tabularnewline
$1$ & encoder &  $(128,96,48)$ & $(128, 32, 16)$\tabularnewline
\midrule
$2$ & encoder &  $(128,96,48)$ & $(128, 32, 16)$ \tabularnewline
$2$ & encoder &  $(128,96,48)$ & $(128, 32, 16)$ \tabularnewline
$2$ & encoder &  $(128,96,48)$ & $(128, 32, 16)$ \tabularnewline
\midrule
$3$ & encoder &  $(128,48,24)$ & $(128, 144, 72)$ \tabularnewline
$3$ & encoder &  $(192,48,24)$ & $(192, 144, 72)$ \tabularnewline
$3$ & encoder &  $(192,48,24)$ & $(192, 144, 72)$ \tabularnewline
\midrule
$4$ & encoder &  $(192,48,24)$ & $(192, 144, 72)$\tabularnewline
$4$ & encoder &  $(192,48,24)$ & $(192, 144, 72)$ \tabularnewline
$4$ & encoder &  $(192,48,24)$ & $(192, 144, 72)$ \tabularnewline
\specialrule{.2em}{.1em}{.1em} 
$3$ & decoder &  $(384,48,24)$ & $(384, 144, 72)$\tabularnewline
$3$ & decoder &  $(192,48,24)$ & $(192, 144, 72)$\tabularnewline
$3$ & decoder &  $(192,48,24)$ & $(192, 144, 72)$\tabularnewline
\midrule
$2$ & decoder &  $(320,96,48)$ & $(320, 32, 16)$ \tabularnewline
$2$ & decoder &  $(128,96,48)$ & $(128, 32, 16)$\tabularnewline
$2$ & decoder &  $(128,96,48)$ & $(128, 32, 16)$\tabularnewline
\midrule
$1$ & decoder &  $(256,96,48)$ & $(256, 32, 16)$\tabularnewline
$1$ & decoder &  $(128,96,48)$ & $(128, 32, 16)$\tabularnewline
$1$ & decoder &  $(128,96,48)$ & $(128, 32, 16)$\tabularnewline
\midrule\midrule
$1$ & final & \multicolumn{2}{c}{$(64, 32)$} \tabularnewline
$1$ & final & \multicolumn{2}{c}{$(32, C)$} \tabularnewline
\bottomrule
\tabularnewline
&ScanNet&\# parameters: & $\mathbf{761,333}$ \tabularnewline
&Matterport3D&\# parameters: & $\mathbf{761,366}$
\end{tabular}
\caption{\label{tab:architecture_2_ratio} \textbf{ScanNet/Matterport architecture.}
We use more filters in the later two mesh levels and the best performing filter ratio from~\reftab{scannet_different_ratios}.
We obtain different numbers of parameters for ScanNet and Matterport3D since they differ in their number of semantic classes ($C_\text{scannet}=21$ and $C_\text{matterport}=22$).}
\end{subtable}
\hspace{\fill}
\begin{subtable}[t!]{0.48\textwidth}
\flushright
\vspace{-15px}
\begin{tabular}{ccrr}
\toprule 
\#level & level type & module type & filters \tabularnewline
\midrule
$1$ & encoder & edge+BN+ReLU & $2*(9,64,32)$ \tabularnewline
$1$ & encoder & edge+BN+ReLU & $2*(128,64,32)$ \tabularnewline
$1$ & encoder & edge+BN+ReLU & $2*(128,64,32)$\tabularnewline
\midrule
$2$ & encoder & edge+BN+ReLU & $2*(128,64,32)$ \tabularnewline
$2$ & encoder & edge+BN+ReLU & $2*(128,64,32)$ \tabularnewline
$2$ & encoder & edge+BN+ReLU & $2*(128,64,32)$ \tabularnewline
\midrule
$3$ & encoder & edge+BN+ReLU & $2*(128,96,48)$ \tabularnewline
$3$ & encoder & edge+BN+ReLU & $2*(192,96,48)$ \tabularnewline
$3$ & encoder & edge+BN+ReLU & $2*(192,96,48)$ \tabularnewline
\midrule
$4$ & encoder & edge+BN+ReLU & $2*(192,96,48)$ \tabularnewline
$4$ & encoder & edge+BN+ReLU & $2*(192,96,48)$ \tabularnewline
$4$ & encoder & edge+BN+ReLU & $2*(192,96,48)$ \tabularnewline
\specialrule{.2em}{.1em}{.1em} 
$3$ & decoder & edge+BN+ReLU & $2*(384,96,48)$ \tabularnewline
$3$ & decoder & edge+BN+ReLU & $2*(192,96,48)$ \tabularnewline
$3$ & decoder & edge+BN+ReLU & $2*(192,96,48)$ \tabularnewline
\midrule
$2$ & decoder & edge+BN+ReLU & $2*(320,64,32)$ \tabularnewline
$2$ & decoder & edge+BN+ReLU & $2*(128,64,32)$ \tabularnewline
$2$ & decoder & edge+BN+ReLU & $2*(128,64,32)$ \tabularnewline
\midrule
$1$ & decoder & edge+BN+ReLU & $2*(256,64,32)$ \tabularnewline
$1$ & decoder & edge+BN+ReLU & $2*(128,64,32)$ \tabularnewline
$1$ & decoder & edge+BN+ReLU & $2*(128,64,32)$ \tabularnewline
\midrule\midrule
$1$ & final & Lin+BN+ReLU & $(64, 32)$ \tabularnewline
$1$ & final & Lin & $(32, 13)$ \tabularnewline
\bottomrule
\tabularnewline
&&\# parameters: & $\mathbf{728,045}$ \tabularnewline
\end{tabular}
\vspace{10px}
\caption{\small \textbf{S3DIS architecture.} Unlike the ablation study, we use more filters in the final two mesh levels.}
\label{tab:architecture_2_ratio_s3dis}
\end{subtable}
\caption{\textbf{Architectures for benchmarks.} We present two slightly different architectures for S3DIS and ScanNet/Matterport, respectively. This is due to the comparably lower mesh quality of S3DIS.}
\label{tab:architectures_benchmarks}
\vspace{5px}
\end{table*}
\begin{table*}[t]
\setlength\tabcolsep{2.8pt}
\begin{center}
\begin{tabular}{rcc|ccccccccccccc}
\toprule
Method	 & mIoU & mAcc & ceil.	 & floor	 & wall	 & beam	 & col.	 & wind.	 & door	 & chair	 & table	 & book.	 & sofa	 & board & clut. \\
\midrule
Pointnet\,\cite{Qi17CVPR}	& $41.1$	& $49.0$	& $88.8$	& $97.3$	& $69.8$	& $0.1$	& $3.9$	& $46.3$	& $10.8$	& $52.6$	& $58.9$	& $40.3$	& $5.9$	& $26.4$	& $33.2$	 \\
SegCloud\,\cite{Tchapmi173DV}	& $48.9$	& $57.4$	& $90.1$	& $96.1$	& $69.9$	& $0.0$	& $18.4$	& $38.4$	& $23.1$	& $75.9$	& $70.4$	& $58.4$	& $40.9$	& $13.0$	& $41.6$	\\
Eff 3D Conv\,\cite{ZhangLU18}	& $51.8$	& $68.3$	& $79.8$	& $93.9$	& $69.0$	& $0.2$	& $28.3$	& $38.5$	& $48.3$	& $71.1$	& $73.6$	& $48.7$	& $59.2$	& $29.3$	& $33.1$	\\
RSNet\,\cite{Huang18CVPR} & $51.9$ & $59.4$ & $93.3$ & $98.4$ & $79.2$ & $0.0$ & $15.8$ & $45.4$ & $50.1$ & $65.5$ & $67.9$ & $22.5$ & $52.5$ & $41.0$ & $43.6$ \\
TangentConv\,\cite{Tatarchenko18CVPR}	& $52.6$	& $62.2$	& $90.5$	& $97.7$	& $74.0$	& $0.0$	& $20.7$	& $39.0$	& $31.3$	& $69.4$	& $77.5$	& $38.5$	& $57.3$	& $48.8$	& $39.8$	\\
PointCNN\,\cite{Li18NIPS}	& $57.3$	& $63.9$	& $92.3$	& $98.2$	& $79.4$	& $0.0$	& $17.6$	& $22.8$	& $62.1$	& $80.6$	& $74.4$	& $66.7$	& $31.7$	& $62.1$	& $56.7$	\\
RNN Fusion\,\cite{Ye18ECCV}	& $57.3$	& $63.9$	& $92.3$	& $\mathbf{98.2}$	& $79.4$	& $0.0$	& $17.6$	& $22.8$	& $62.1$	& $74.4$	& $80.6$	& $31.7$	& $66.7$	& $62.1$	& $56.7$	\\
ParamConv\,\cite{Wang18CVPRa}	& $58.3$	& $67.1$	& $92.3$	& $96.2$	& $75.9$	& $\mathbf{0.3}$	& $6.0$	& $\mathbf{69.5}$	& $63.5$	& $66.9$	& $65.6$	& $47.3$	& $68.9$	& $59.1$	& $46.2$ \\
MinkowskiNet\,\cite{Choy2019CVPR}& $65.4$ & $71.7$ & $91.8$ & $98.7$ & $86.2$ & $0.0$ & $34.1$ & $48.9$ & $62.4$ & $89.8$ & $81.6$ & $74.9$ & $47.2$ & $74.4$ & $58.6$ \\
KPConv\,\cite{thomas2019ICCV}	& $\mathbf{67.1}$	& $\mathbf{72.8}$	& $\mathbf{92.8}$	& $97.3$	& $\mathbf{82.4}$	& $0.0$	& $23.9$	& $58.0$	& $69.0$	& $\mathbf{91.0}$	& $81.5$	& $\mathbf{75.3}$	& $\mathbf{75.4}$	& $\mathbf{66.7}$	& $\mathbf{58.9}$ \\
\hline \hline
SPGraph\,\cite{Landrieu17CVPR}	& $58.0$	& $66.5$	& $89.4$	& $96.9$	& $78.1$	& $0.0$	& $\mathbf{42.8}$	& $48.9$	& $61.6$	& $84.7$	& $75.4$	& $69.8$	& $52.6$	& $2.1$	& $52.2$	\\
SPH3D-GCN*\,\cite{lei2019spherical} & $59.5$ & $65.9$ & $93.3$ & $97.1$ & $81.1$ & $0.0$ & $33.2$ & $45.8$ & $43.8$ & $79.7$ & $86.9$ & $33.2$ & $71.5$ & $54.1$ & $53.7$ \\
HPEIN\,\cite{hpein_iccv19} & $61.9$ & $68.3$ & $91.5$ & $98.2$ & $81.4$ & $0.0$ & $23.3$ & $65.3$ & $40.0$ & $75.5$ & $87.7$ & $58.5$ & $67.8$ & $65.6$ & $49.4$ \\
DCM Net (\textbf{Ours}) & $64.0$ & $71.2$ & $92.1$ & $96.8$ & $78.6$ & $0.0$ & $21.6$ & $61.7$ & $54.6$ & $78.9$ & $\mathbf{88.7}$ & $68.1$ & $72.3$ & $66.5$ & $52.4$ \\
\bottomrule
\end{tabular}
\end{center}
\vspace{-15px}
\caption{\textbf{Semantic segmentation IoU scores on S3DIS Area $\mathbf{5}$.}
We furthermore provide mean class accuracy scores. Among all approaches, we perform third best only outperformed by KPConv~\cite{thomas2019ICCV} and MinkowskiNet~\cite{Choy2019CVPR}. Among graph convolutional approaches, we clearly report state-of-the-art with a gap of $2.1\%$ to HPEIN~\cite{hpein_iccv19}.
}
\label{tab:Table_S3DIS_area5} 
\vspace{20px}
\end{table*}
\begin{table*}[t]
\setlength\tabcolsep{3pt}
\begin{center}
\begin{tabular}{rcc|ccccccccccccc}
\toprule
Method	 & mIoU & mAcc & ceil.	 & floor	 & wall	 & beam	 & col.	 & wind.	 & door	 & chair	 & table	 & book.	 & sofa	 & board & clut. \\
\midrule
Pointnet\,\cite{Qi17CVPR}	& $47.6$	& $66.2$	& $88.0$	& $88.7$	& $69.3$	& $42.4$	& $23.1$	& $47.5$	& $51.6$	& $42.0$	& $54.1$	& $38.2$	& $9.6$	& $29.4$	& $35.2$	\\
RSNet\,\cite{Huang18CVPR}	& $56.5$	& $66.5$	& $92.5$	& $92.8$	& $78.6$	& $32.8$	& $34.4$	& $51.6$	& $68.1$	& $60.1$	& $59.7$	& $50.2$	& $16.4$	& $44.9$	& $52.0$	\\
PointCNN\,\cite{Li18NIPS}	& $65.4$	& $75.6$	& $\mathbf{94.8}$	& $\mathbf{97.3}$	& $75.8$	& $63.3$	& $51.7$	& $58.4$	& $57.2$	& $71.6$	& $69.1$	& $39.1$	& $61.2$	& $52.2$	& $58.6$	\\
KPConv\,\cite{thomas2019ICCV} & $\mathbf{70.6}$	& $79.1$	& $93.6$	& $92.4$	& $\mathbf{83.1}$	& $\mathbf{63.9}$	& $54.3$	& $66.1$	& $\mathbf{76.6}$	& $57.8$	& $64.0$	& $\mathbf{69.3}$	& $\mathbf{74.9}$	& $61.3$	& $60.3$ \\
\hline \hline
SPGraph\,\cite{Landrieu17CVPR}	& $62.1$	& $73.0$	& $89.9$	& $95.1$	& $76.4$	& $62.8$	& $47.1$	& $55.3$	& $68.4$	& $73.5$	& $69.2$	& $63.2$	& $45.9$	& $8.7$	& $52.9$	\\
HPEIN\,\cite{hpein_iccv19} & $67.8$ & $76.3$ & - & - & - & - & - & - & - & - & - & - & - & - & - \\
SPH3D-GCN*\,\cite{lei2019spherical} & $68.9$ & $77.9$ & $93.3$ & $96.2$ & $81.9$ & $58.6$ & $\mathbf{55.9}$ & $55.9$ & $71.7$ & $72.1$ & $\mathbf{82.4}$ & $48.5$ & $64.5$ & $54.8$ & $60.4$ \\
DCM Net (\textbf{Ours}) & $69.7$ & $\mathbf{80.7}$ & $93.7$ & $96.6$ & $81.2$ & $44.6$ & $44.9$ & $\mathbf{73.0}$ & $73.8$ & $71.4$ & $74.3$ & $63.3$ & $63.9$  & $\mathbf{63.0}$ & $\mathbf{61.9}$ \\
\bottomrule
\end{tabular}
\end{center}
\vspace{-15px}
\caption{\textbf{Semantic segmentation IoU scores on S3DIS $\mathbf{k}$-fold.} We furthermore provide mean class accuracy scores. Among all approaches, we perform second best only outperformed by KPConv~\cite{thomas2019ICCV}. Among graph convolutional approaches, we report state-of-the-art with a gap of $0.8\%$ to the concurrent work SPH3D-GCN~\cite{lei2019spherical}.}
\label{tab:TableS3DISkfold} 
\vspace{5px}
\end{table*}
\definecolor{ceiling}{rgb}{0., 1., 0.}
\definecolor{floor}{rgb}{0., 0., 1.}
\definecolor{wall}{rgb}{0., 1., 1.}
\definecolor{beam}{rgb}{1.0000, 1.0000, 0.0000}
\definecolor{column}{rgb}{1.0000, 0.0000, 1.0000}
\definecolor{window}{rgb}{0.3922, 0.3922, 1.0000}
\definecolor{door}{rgb}{0.7843, 0.7843, 0.3922}
\definecolor{chair}{rgb}{1.0, 0.0, 0.0}
\definecolor{table}{rgb}{0.6667, 0.4706, 0.7843}
\definecolor{bookshelf}{rgb}{0.0392, 0.7843, 0.3922}
\definecolor{sofa}{rgb}{0.7843, 0.3922, 0.3922}
\definecolor{board}{rgb}{0.7843, 0.7843, 0.7843}
\definecolor{clutter}{rgb}{0.1961, 0.1961, 0.1961}

\begin{figure*}[t]
\centering
\includegraphics[width=0.25\linewidth, trim={0 0 0 0}, clip]{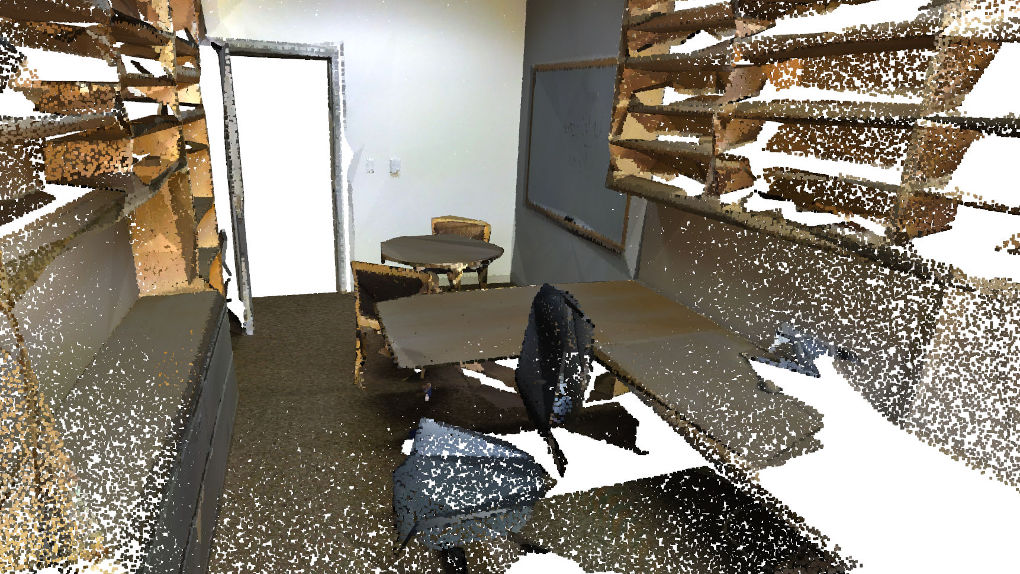}\hfill%
\includegraphics[width=0.25\linewidth, trim={0 0 0 0}, clip]{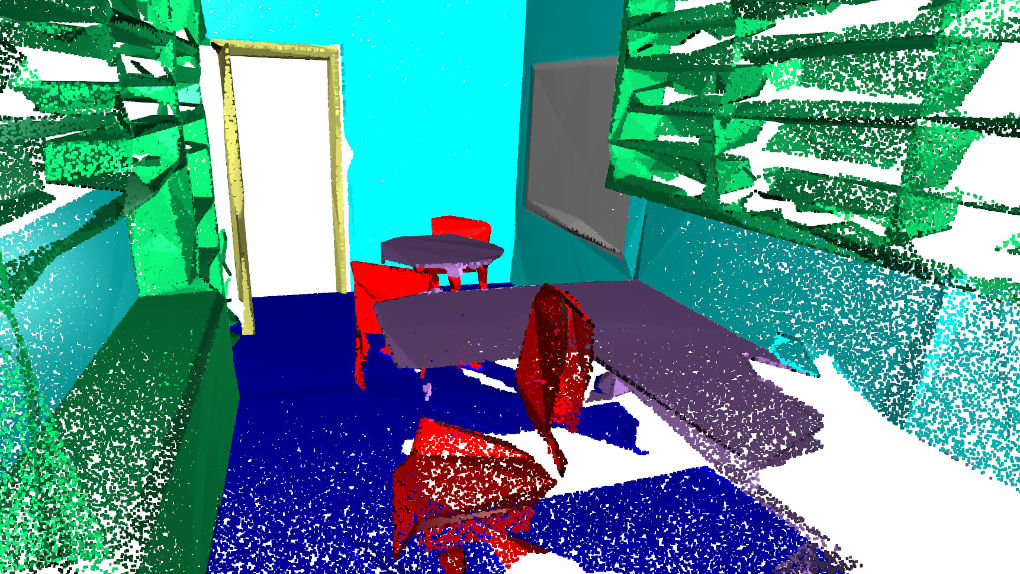}\hfill%
\includegraphics[width=0.25\linewidth, trim={0 0 0 0}, clip]{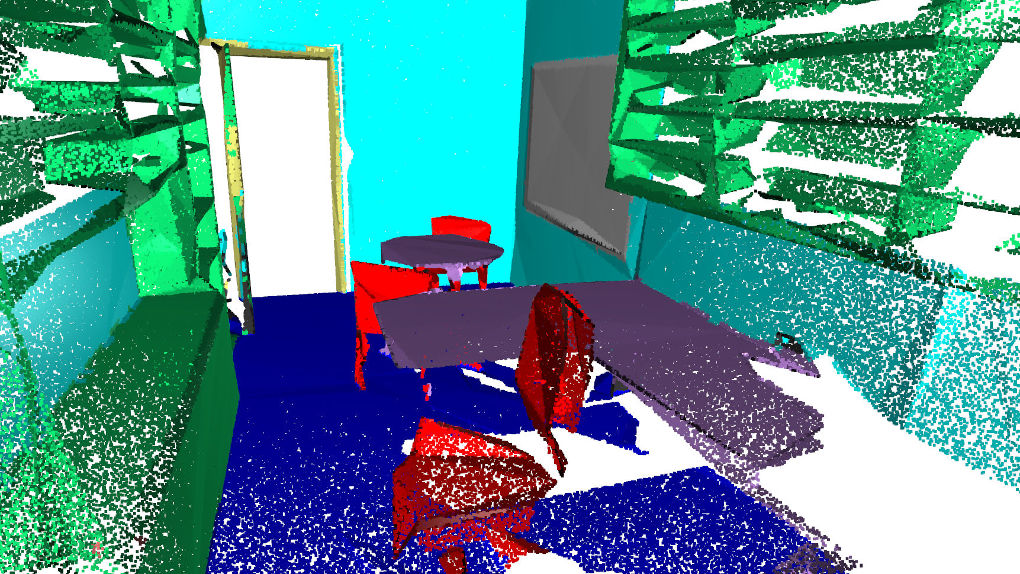}\hfill
\includegraphics[width=0.25\linewidth, trim={0 0 0 0}, clip]{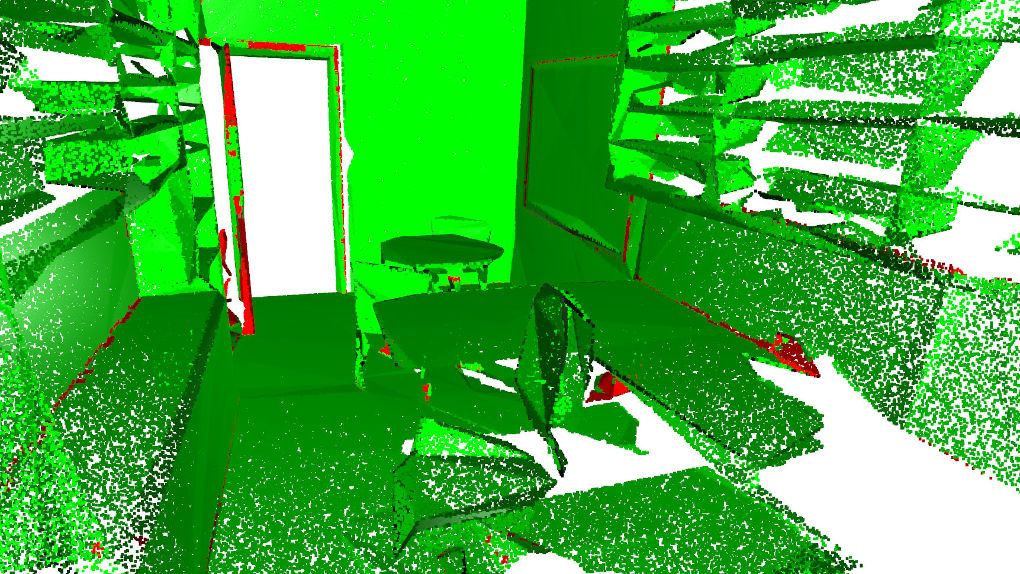}

\includegraphics[width=0.25\linewidth, trim={0 0 0 0}, clip]{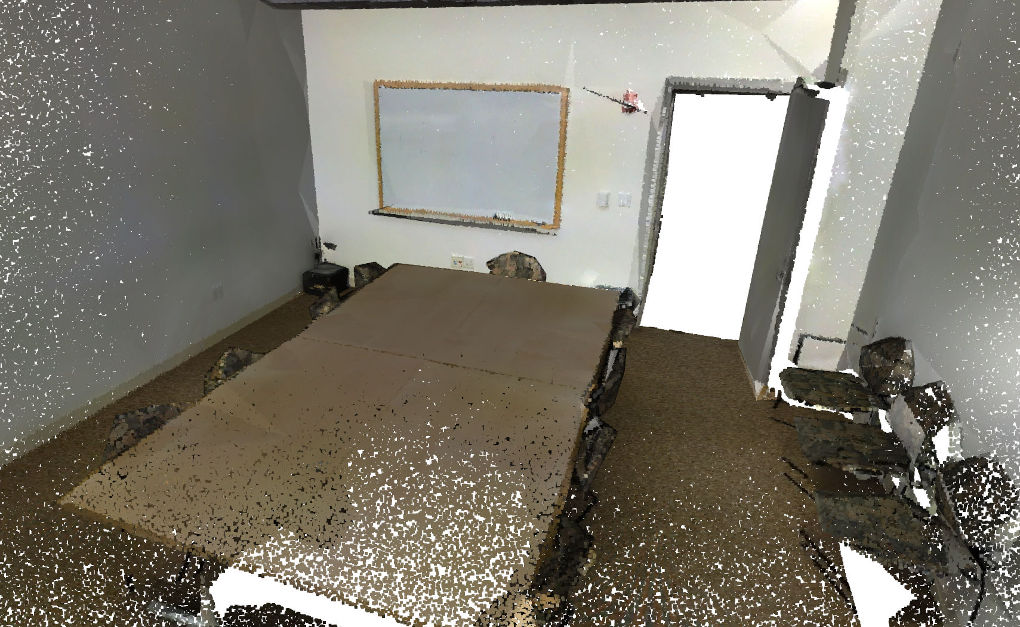}\hfill%
\includegraphics[width=0.25\linewidth, trim={0 0 0 0}, clip]{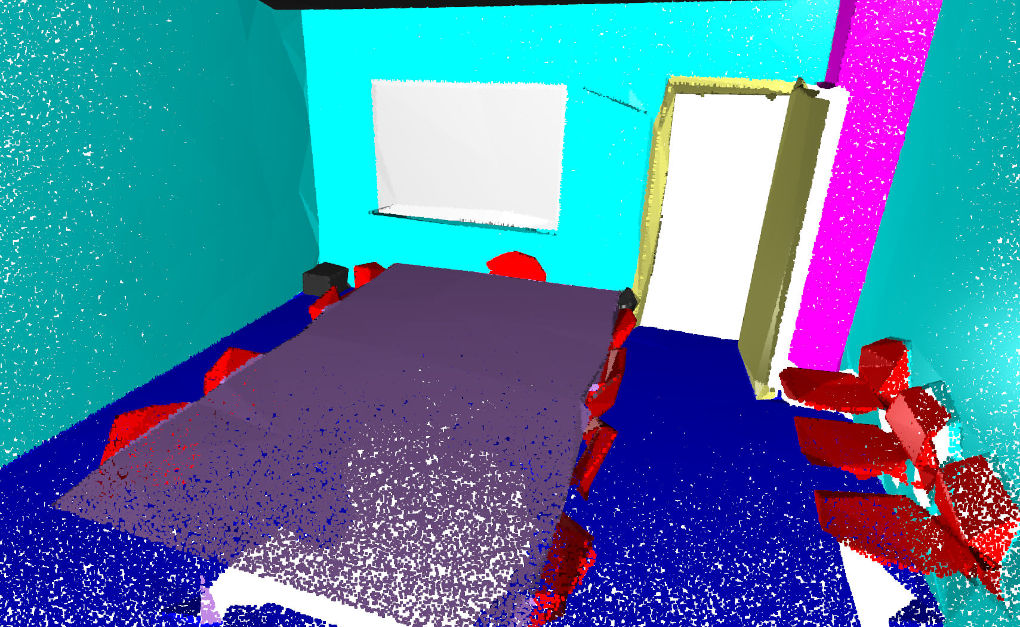}\hfill%
\includegraphics[width=0.25\linewidth, trim={0 0 0 0}, clip]{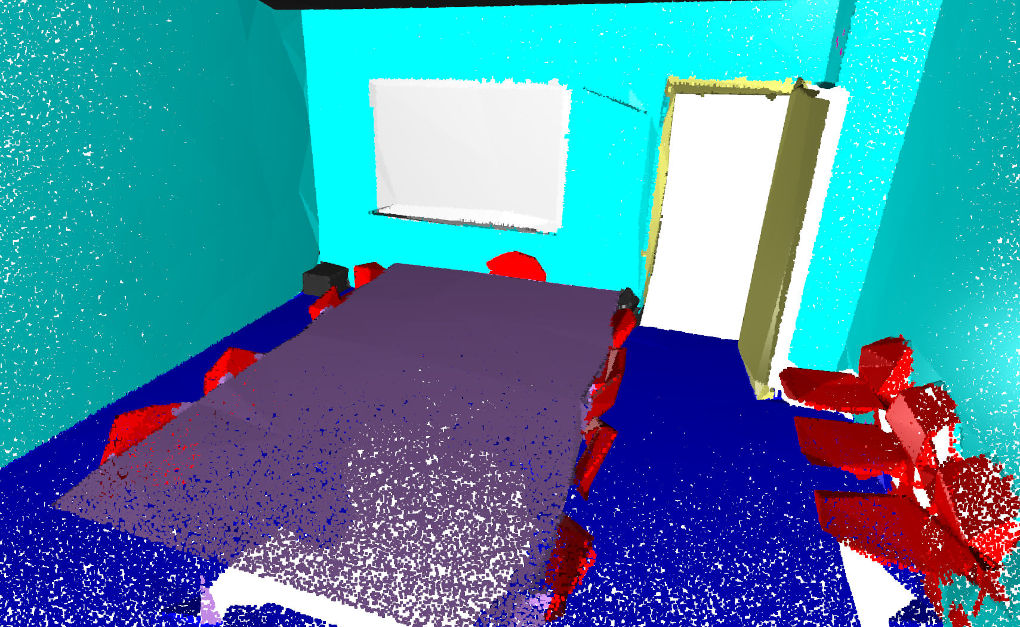}\hfill
\includegraphics[width=0.25\linewidth, trim={0 0 0 0}, clip]{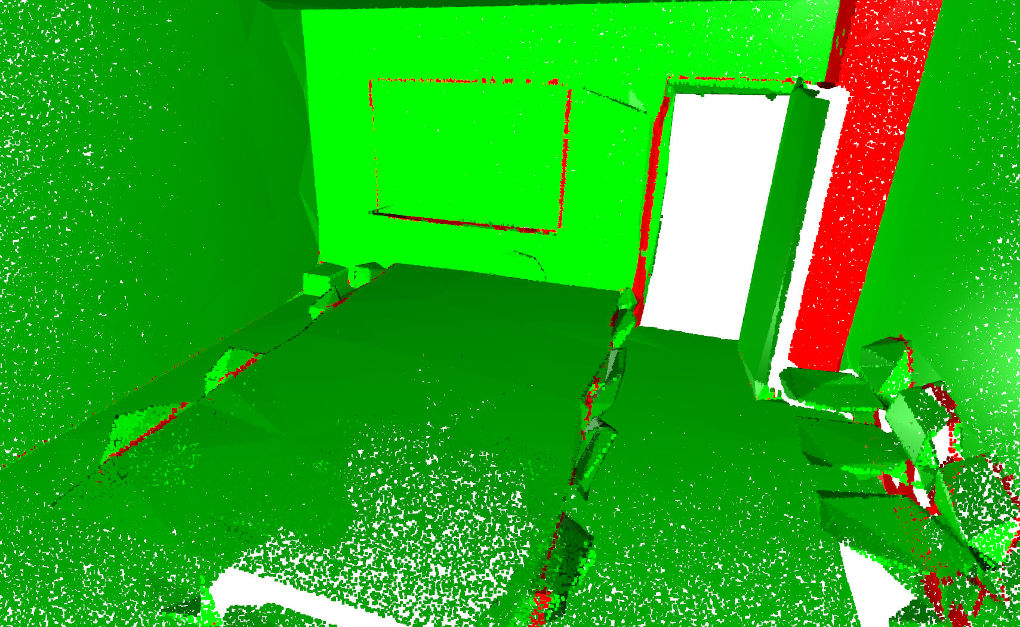}

\includegraphics[width=0.25\linewidth, trim={0 0 0 0}, clip]{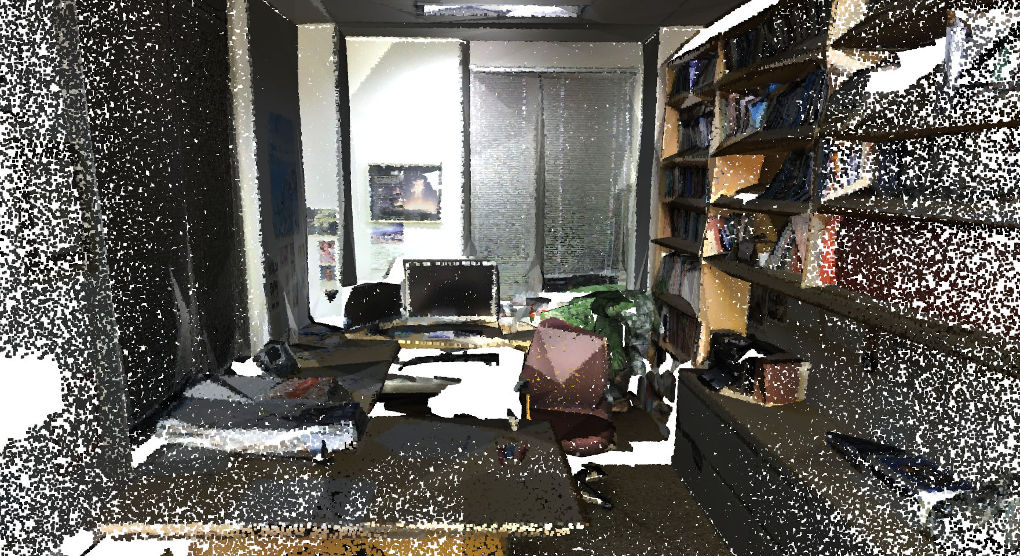}\hfill%
\includegraphics[width=0.25\linewidth, trim={0 0 0 0}, clip]{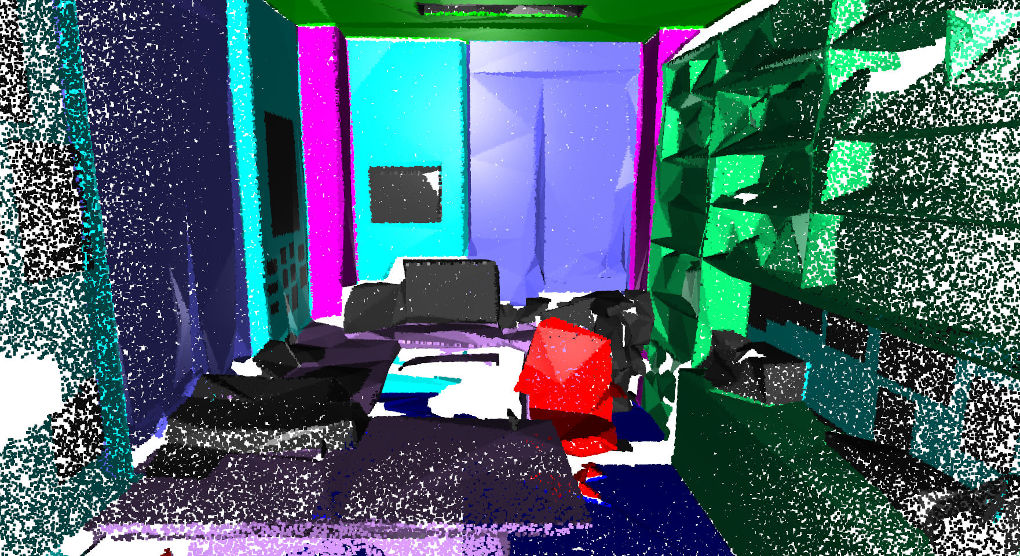}\hfill%
\includegraphics[width=0.25\linewidth, trim={0 0 0 0}, clip]{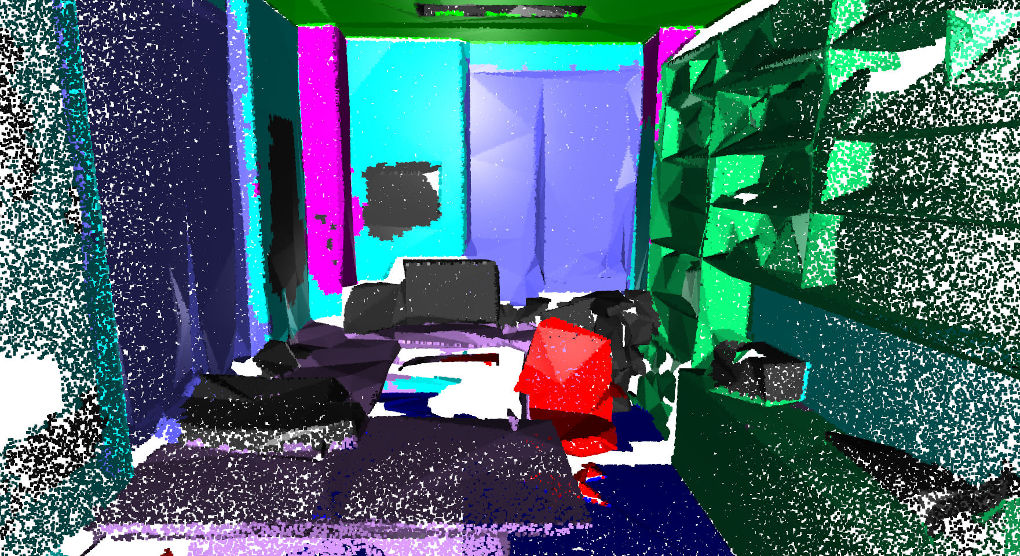}\hfill
\includegraphics[width=0.25\linewidth, trim={0 0 0 0}, clip]{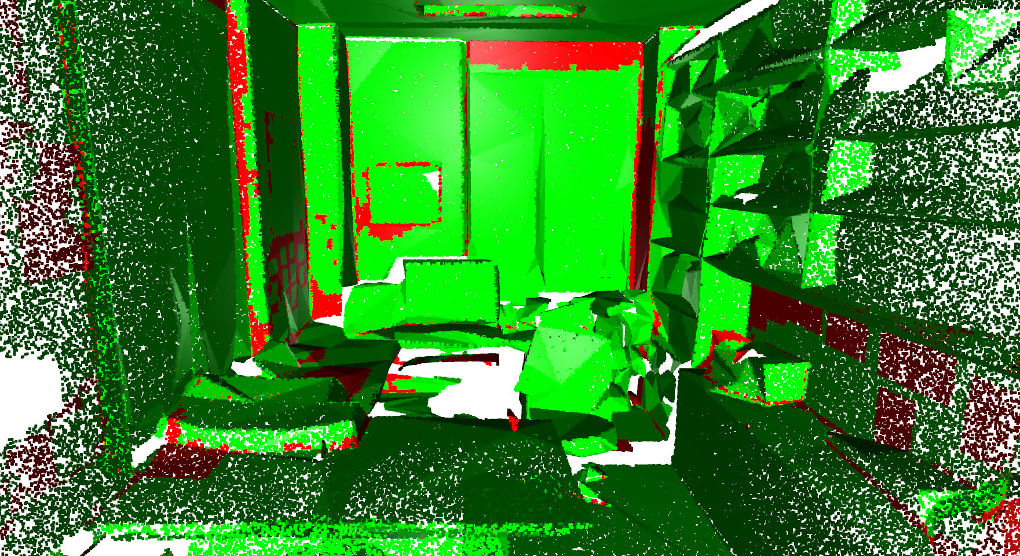}

\vspace{3mm}

\begin{small}
\begin{tabular}{cccc}
\textbf{Input Point Cloud} & \textbf{Ground Truth} & \textbf{Prediction} & \textbf{Error} \\
\hspace{4cm} & \hspace{4cm} & \hspace{4cm} & \hspace{4cm}
\\
\end{tabular}
\\
\vspace{-5pt}
\textcolor{ceiling}{\ColorMapCircle}~ceiling
\textcolor{floor}{\ColorMapCircle}~floor
\textcolor{wall}{\ColorMapCircle}~wall
\textcolor{beam}{\ColorMapCircle}~beam
\textcolor{column}{\ColorMapCircle}~column
\textcolor{window}{\ColorMapCircle}~window
\textcolor{door}{\ColorMapCircle}~door
\textcolor{chair}{\ColorMapCircle}~chair
\textcolor{table}{\ColorMapCircle}~table
\textcolor{bookshelf}{\ColorMapCircle}~bookshelf
\textcolor{sofa}{\ColorMapCircle}~sofa
\textcolor{board}{\ColorMapCircle}~board
\textcolor{clutter}{\ColorMapCircle}~clutter
\end{small}
\\
\caption{\textbf{Results on Stanford Large-Scale 3D Indoor Spaces~\cite{Armeni16CVPR}.}
Our method correctly predicts challenging classes such as \textcolor{board}{\ColorMapCircle}~board, while maintaining clear boundaries for most of the classes.
In the second row, our method confuses the similar classes \textcolor{column}{\ColorMapCircle}~column and \textcolor{wall}{\ColorMapCircle}~wall.
In the last example, it becomes evident that our method tends to produce unclear boundaries for diverse \textcolor{clutter}{\ColorMapCircle}~clutter regions.}
\label{fig:s3dis_quali}
\end{figure*}

\definecolor{unlabeled}{rgb}{0., 0., 0.}
\definecolor{wall}{rgb}{0.68235294, 0.78039216, 0.90980392}
\definecolor{floor}{rgb}{0.59607843, 0.8745098 , 0.54117647}
\definecolor{cabinet}{rgb}{0.12156863, 0.46666667, 0.70588235}
\definecolor{bed}{rgb}{1.        , 0.73333333, 0.47058824}
\definecolor{chair}{rgb}{0.7372549 , 0.74117647, 0.13333333}
\definecolor{sofa}{rgb}{0.54901961, 0.3372549 , 0.29411765}
\definecolor{table}{rgb}{1.        , 0.59607843, 0.58823529}
\definecolor{door}{rgb}{0.83921569, 0.15294118, 0.15686275}
\definecolor{window}{rgb}{0.77254902, 0.69019608, 0.83529412}
\definecolor{bookshelf}{rgb}{0.58039216, 0.40392157, 0.74117647}
\definecolor{picture}{rgb}{0.76862745, 0.61176471, 0.58039216}
\definecolor{counter}{rgb}{0.09019608, 0.74509804, 0.81176471}
\definecolor{desk}{rgb}{0.96862745, 0.71372549, 0.82352941}
\definecolor{curtain}{rgb}{0.85882353, 0.85882353, 0.55294118}
\definecolor{refrigerator}{rgb}{1.        , 0.49803922, 0.05490196}
\definecolor{showercurtain}{rgb}{0.61960784, 0.85490196, 0.89803922}
\definecolor{toilet}{rgb}{0.17254902, 0.62745098, 0.17254902}
\definecolor{sink}{rgb}{0.43921569, 0.50196078, 0.56470588}
\definecolor{bathtub}{rgb}{0.89019608, 0.46666667, 0.76078431}
\definecolor{otherfurn}{rgb}{0.32156863, 0.32941176, 0.63921569}
\definecolor{ceiling}{rgb}{0.090196078, 0.180392157, 0.039215686}

\begin{figure*}[t]
\centering

\includegraphics[width=0.25\linewidth, trim={0 0 0 0}, clip]{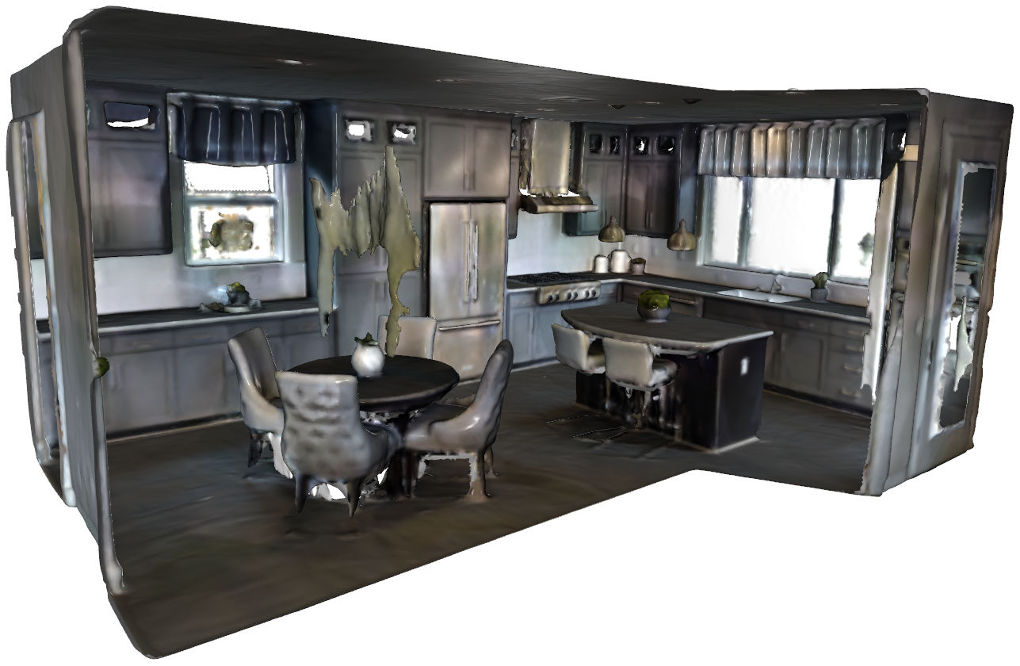}\hfill%
\includegraphics[width=0.25\linewidth, trim={0 0 0 0}, clip]{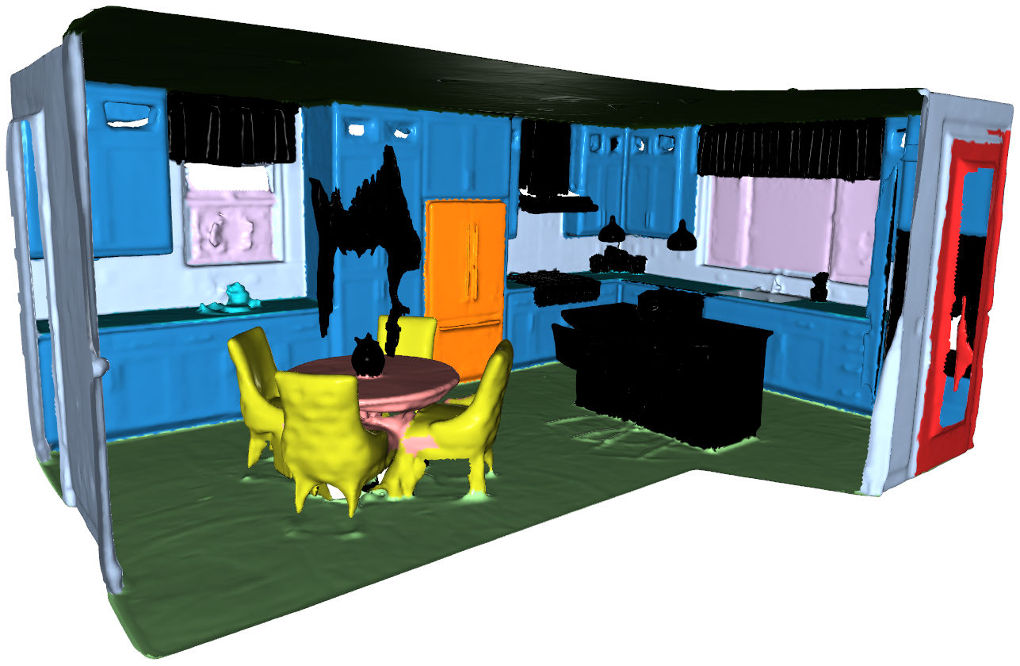}\hfill%
\includegraphics[width=0.25\linewidth, trim={0 0 0 0}, clip]{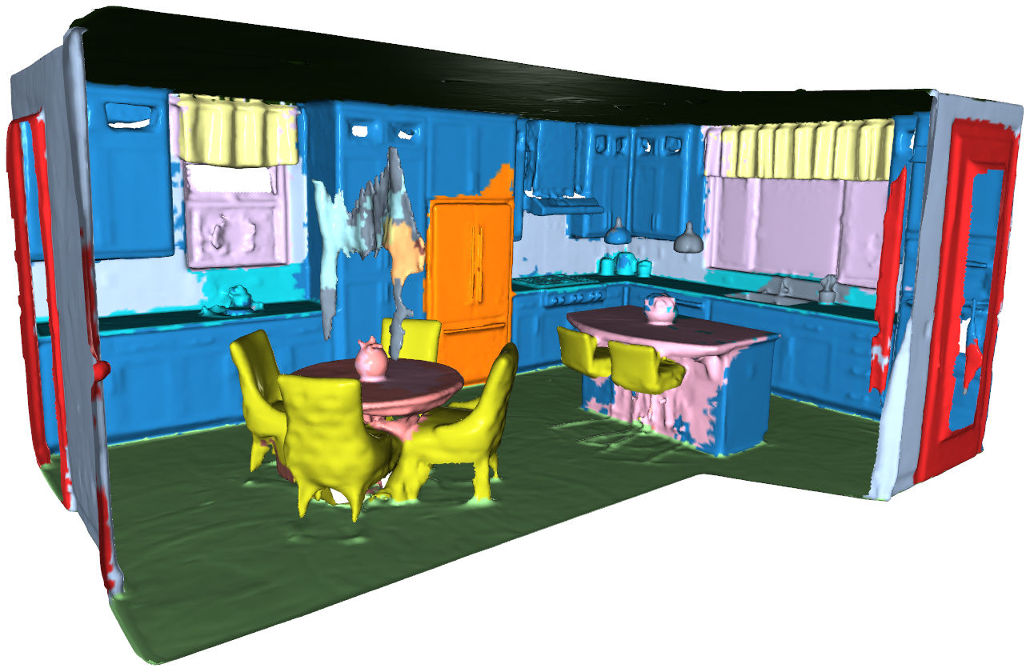}\hfill
\includegraphics[width=0.25\linewidth, trim={0 0 0 0}, clip]{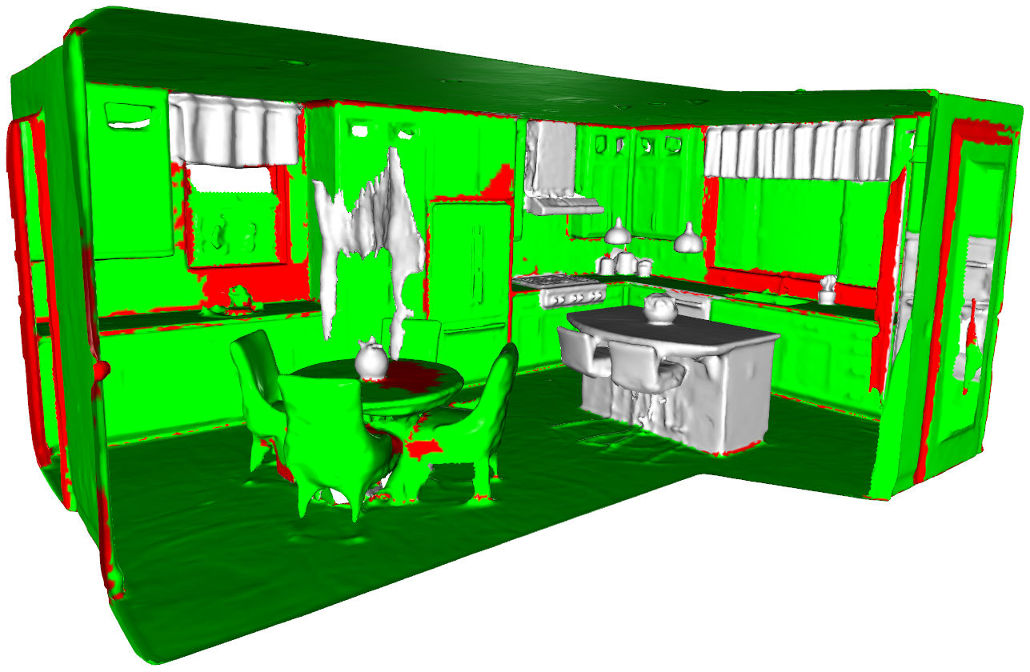}

\includegraphics[width=0.25\linewidth, trim={0 0 0 0}, clip]{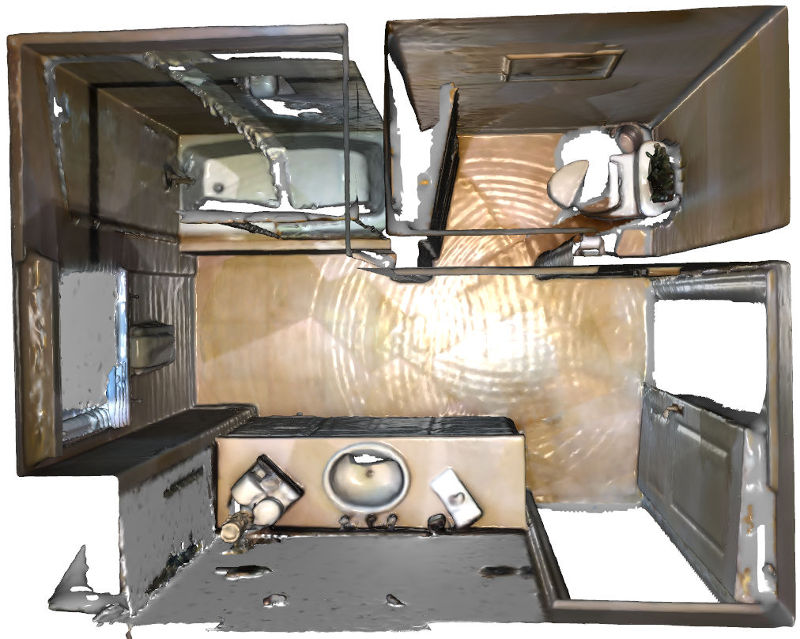}\hfill%
\includegraphics[width=0.25\linewidth, trim={0 0 0 0}, clip]{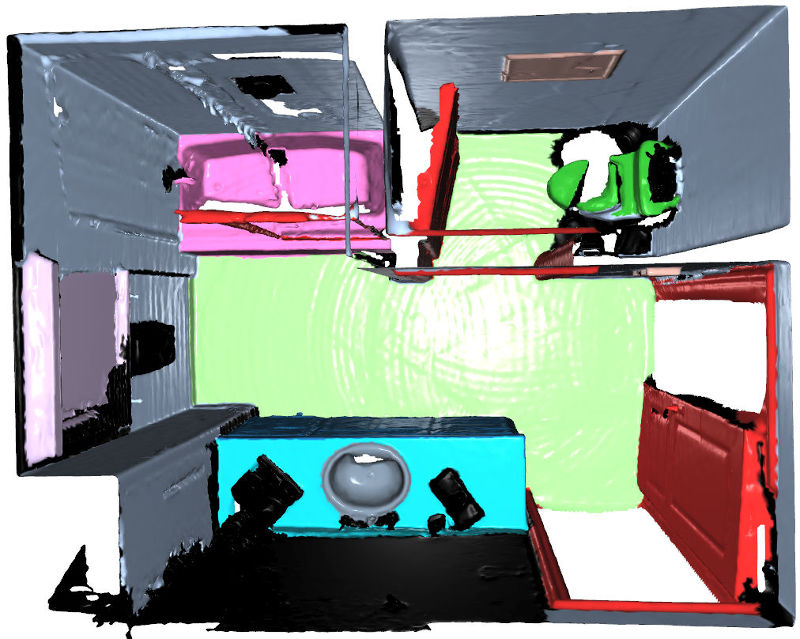}\hfill%
\includegraphics[width=0.25\linewidth, trim={0 0 0 0}, clip]{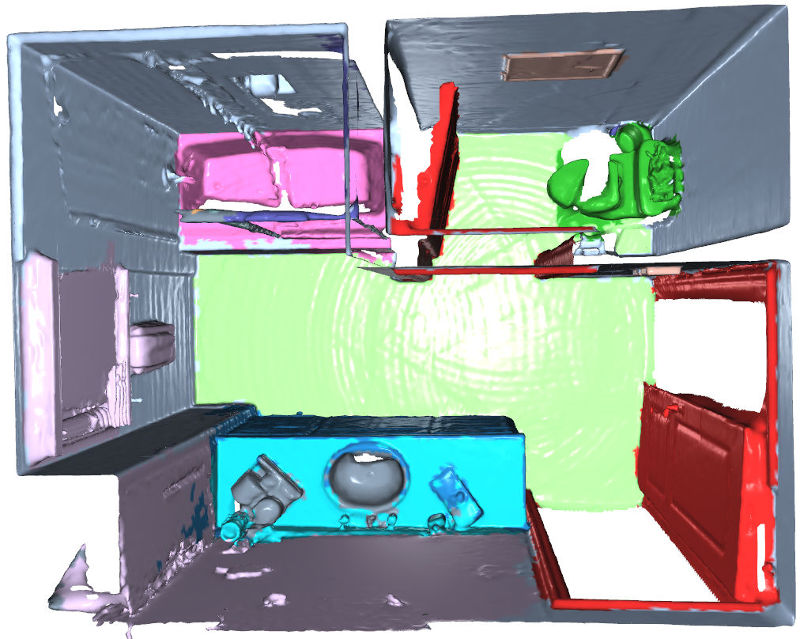}\hfill
\includegraphics[width=0.25\linewidth, trim={0 0 0 0}, clip]{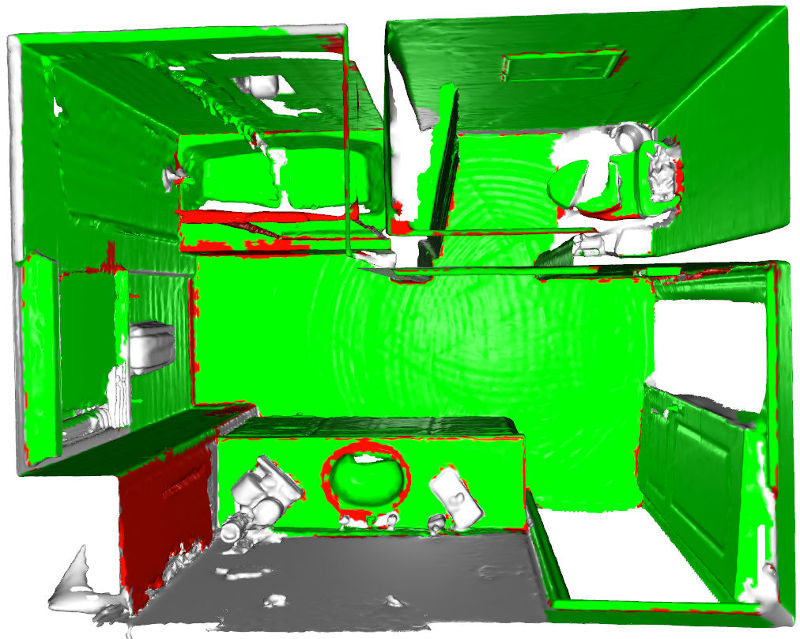}

\includegraphics[width=0.25\linewidth, trim={0 0 0 0}, clip]{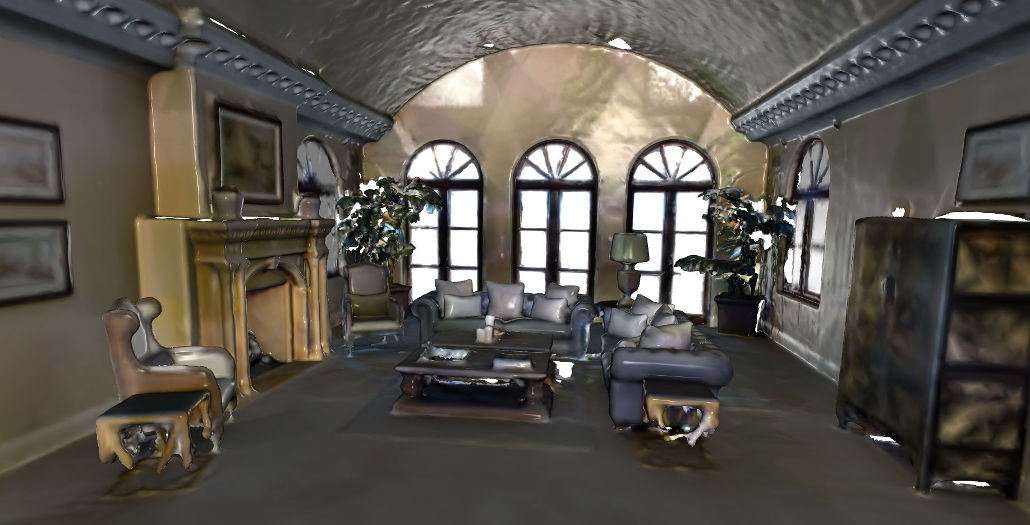}\hfill%
\includegraphics[width=0.25\linewidth, trim={0 0 0 0}, clip]{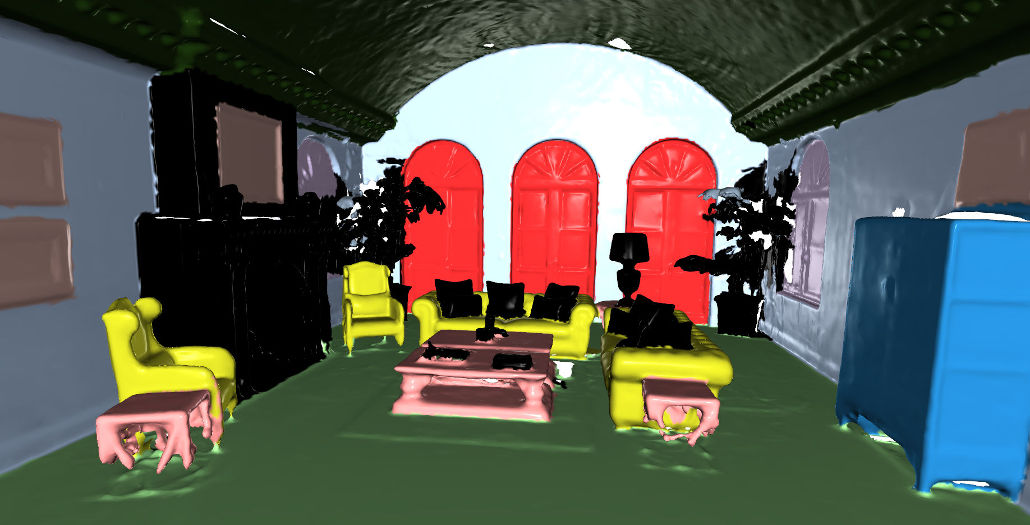}\hfill%
\includegraphics[width=0.25\linewidth, trim={0 0 0 0}, clip]{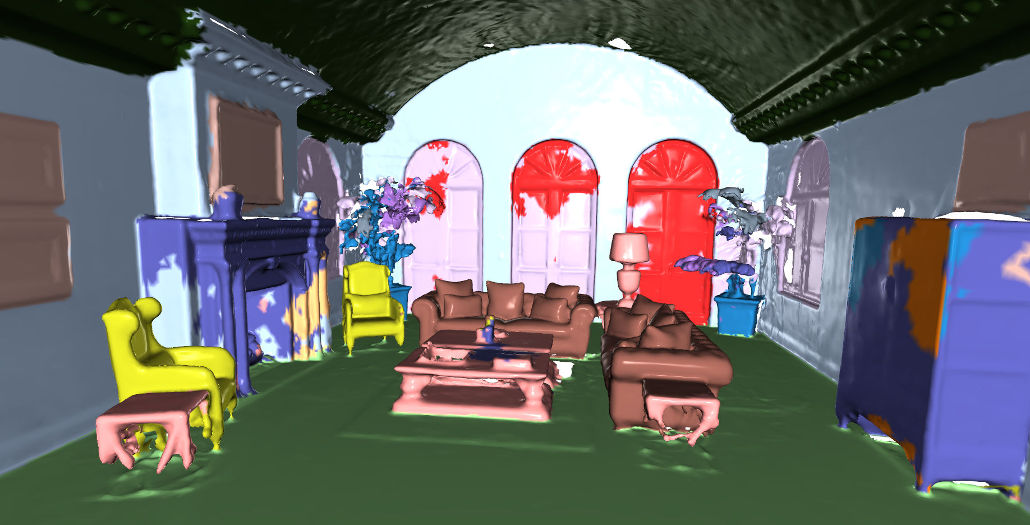}\hfill
\includegraphics[width=0.25\linewidth, trim={0 0 0 0}, clip]{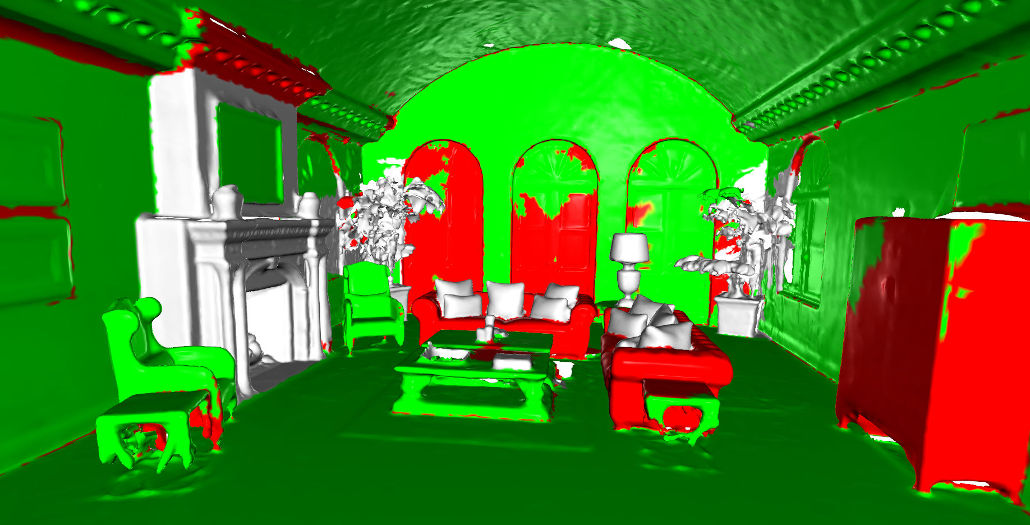}

\vspace{3mm}

\begin{small}
\begin{tabular}{cccc}
\textbf{Input Mesh} & \textbf{Ground Truth} & \textbf{Prediction} & \textbf{Error} \\
\hspace{4cm} & \hspace{4cm} & \hspace{4cm} & \hspace{4cm}
\\
\end{tabular}
\\
\vspace{-5pt}
\textcolor{unlabeled}{\ColorMapCircle}~unlabeled
\textcolor{wall}{\ColorMapCircle}~wall
\textcolor{floor}{\ColorMapCircle}~floor
\textcolor{cabinet}{\ColorMapCircle}~cabinet
\textcolor{chair}{\ColorMapCircle}~chair
\textcolor{sofa}{\ColorMapCircle}~sofa
\textcolor{table}{\ColorMapCircle}~table
\textcolor{door}{\ColorMapCircle}~door
\textcolor{window}{\ColorMapCircle}~window
\textcolor{picture}{\ColorMapCircle}~picture
\textcolor{counter}{\ColorMapCircle}~counter
\textcolor{desk}{\ColorMapCircle}~desk
\textcolor{refrigerator}{\ColorMapCircle}~fridge
\textcolor{toilet}{\ColorMapCircle}~toilet
\textcolor{sink}{\ColorMapCircle}~sink
\textcolor{bathtub}{\ColorMapCircle}~bathtub
\textcolor{otherfurn}{\ColorMapCircle}~otherfurn
\textcolor{ceiling}{\ColorMapCircle}~ceiling
\end{small}
\\
\caption{\textbf{Results on Matterport3D~\cite{Matterport3D}.}
Our method correctly predicts even \textcolor{unlabeled}{\ColorMapCircle}~unlabeled regions.
However, reasonable errors occur, such as confusing \textcolor{window}{\ColorMapCircle}~windows extending down to the floor as \textcolor{door}{\ColorMapCircle}~doors.
In the last row, our algorithm correctly predicts \textcolor{sofa}{\ColorMapCircle}~sofa even though the ground truth is falsely labeled as \textcolor{chair}{\ColorMapCircle}~chair.
}
\label{fig:matterport_quali}
\end{figure*}

\begin{table*}
\centering
\begin{tabular}{rc|cccccl}
\toprule 
\multirow{2}{*}{\textbf{ScanNet [8] Test}} &
\multirow{2}{*}{mIoU} &
\multicolumn{5}{c}{Data Representation} &
\multirow{2}{*}{Features} \\ 
&&Points&Voxel&Mesh&2D&Texture\\
\toprule 
PointNet [39]& - & \checkmark & - & - & - & - & XYZ-RGB\\
PointNet++ [40]				 	 & $33.9$ & \checkmark & - & - &-  &- &XYZ \\
FCPN [11] 		 		 & $44.7$ & \checkmark & \checkmark & - & - & - & XYZ-RGB-N \\
\hline
3DMV [9]			             & $48.3$  & \checkmark & \checkmark & - &\checkmark & - & XYZ-RGB-N \\
JPBNet [6] & $63.4$ & \checkmark & - & - & \checkmark & - & XYZ-RGB-N\\
MVPNet [26] & $64.1$ & \checkmark & - & - & \checkmark & - & XYZ-RGB-N\\
\hline
Tangent Conv [48] & $43.8$ & \checkmark & -  & - & - & - & XYZ-RGB-N \\
SurfaceConvPF [20] & $44.2$&-& - & \checkmark & - &- & XYZ-RGB-N\\
TextureNet [25] & $56.6$& - & - &\checkmark & \checkmark & \checkmark& XYZ-RGB-N\\
\hline
PointCNN [33] &$45.8$ & \checkmark & - & -& - & - & XYZ-RGB-N \\
ParamConv [52] & - & \checkmark & - & - & - & - & XYZ-RGB \\
MCCN [22] & $63.3$ & \checkmark & - & - & - & - & XYZ-RGB-N\\
PointConv [56] & $66.6$ & \checkmark & - & - & - & - & XYZ-RGB-N\\
KPConv [50] & $68.4$ & \checkmark & - & - & - & - & XYZ-RGB\\
\hline
SparseConvNet [17] & $72.5$ & - & \checkmark & - & - & - & XYZ-RGB \\
MinkowskiNet [7] &$\mathbf{73.4}$& - &\checkmark & - & - & - & XYZ-RGB\\
\hline \hline
DeepGCN [31] & - & \checkmark & - & - & - & - & XYZ-RGB-N\\
SPGraph [34] & - & \checkmark & - & - & - & - & XYZ-RGB\\
SPH3D-GCN [30] & $61.0$ & \checkmark & - & - & - & - & XYZ-RGB-N\\
HPEIN [27] & $61.8$ & \checkmark & - & - & - & - & XYZ-RGB-N \\
DCM Net \textbf{(Ours)} & $65.8$ & \checkmark & - & \checkmark & - & - & XYZ-RGB-N \\
\bottomrule
\end{tabular} 
\caption{\textbf{Data representations and input features.}
We show the data representation and input features of each approach.
}
\label{tab:input_features}
\end{table*}
}

\end{document}